\documentclass[10pt,twocolumn,letterpaper]{article}

\usepackage[pagenumbers]{cvpr} 

\newcommand{\supp}{\emph{Supp.~Mat.}}

\usepackage[dvipsnames]{xcolor}
\usepackage{dsfont}

\newcommand{\singlecaption}{t}
\newcommand{\datasetcaptions}{\mathcal{T}}
\newcommand{\knowledgebase}{\mathcal{B}}

\newcommand{\indicatorfunction}{\mathds{1}}

\newcommand{\biasname}{b}

\newcommand{\generator}{\mathrm{G}}

\newcommand{\myquote}[1]{\emph{``#1"}}

\newcommand{\para}[1]{\noindent \textbf{#1.}}

\definecolor{cvprblue}{rgb}{0.21,0.49,0.74}
\definecolor{cadmiumorange}{rgb}{0.93, 0.53, 0.18}
\usepackage[pagebackref,breaklinks,colorlinks,citecolor=cvprblue]{hyperref}
\usepackage{multirow}
\usepackage{svg}
\usepackage{pgfplots}
\usepackage{colortbl}
\usepackage{listings}
\usepackage{caption}
\usepackage{multicol}
\usepackage[accsupp]{axessibility} 
\def\paperID{8219} 
\def\confName{CVPR}
\def\confYear{2024}

\makeatletter
\def\blfootnote{\xdef\@thefnmark{}\@footnotetext}
\makeatother

\title{OpenBias: Open-set Bias Detection in Text-to-Image Generative Models}

\author{Moreno D'Incà\textsuperscript{1}, Elia Peruzzo\textsuperscript{1}, Massimiliano Mancini\textsuperscript{1}, Dejia Xu\textsuperscript{2}, Vidit Goel\textsuperscript{3,4}, \\Xingqian Xu\textsuperscript{3,4}, Zhangyang Wang\textsuperscript{2,4},  Humphrey Shi\textsuperscript{3,4}$^\dagger$, Nicu Sebe\textsuperscript{1}$^\dagger$\\
{\small \textsuperscript{1}University of Trento, \textsuperscript{2}UT Austin, \textsuperscript{3}SHI Labs @ Georgia Tech \& UIUC, \textsuperscript{4}Picsart AI Research (PAIR)} \\
{\small \textbf{\url{https://github.com/Picsart-AI-Research/OpenBias}}}
}

\begin{document}
\maketitle
\begin{abstract}
    Text-to-image generative models are becoming increasingly popular and accessible to the general public. As these models see large-scale deployments, it is necessary to deeply investigate 
    their safety and fairness 
    to 
    not disseminate and perpetuate any kind of biases. 
    However, 
    existing works focus %
    on detecting closed sets of biases defined a priori, limiting the studies to well-known concepts. In this paper, we tackle the challenge of open-set bias detection in text-to-image generative models presenting OpenBias, 
    a new pipeline that 
    identifies and quantifies the severity of biases agnostically, without access to any precompiled set. OpenBias has three stages. 
    In the first phase, we leverage a Large Language Model (LLM) to propose biases given a set of captions. Secondly, the target generative model produces 
    images using the same set of captions. Lastly, a Vision Question Answering model recognizes the presence and extent of the previously proposed biases. We study the behavior of Stable Diffusion 1.5, 2, and XL emphasizing new biases, never investigated before. 
 Via quantitative experiments, we demonstrate that 
 OpenBias agrees 
 with current closed-set bias detection methods and human judgement.
\end{abstract}    
\begin{figure}[htbp]
    \centering
    \includegraphics[width=.75\linewidth]{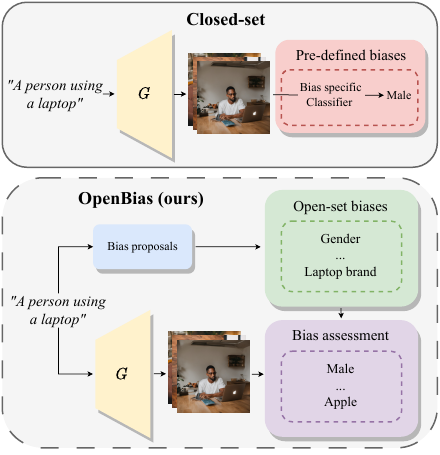}
    \caption{OpenBias discovers biases in T2I models within an open-set scenario. In contrast to previous works~\cite{fairDiffusion2023, ITIGEN_2023_ICCV, kenfack2022repfairgan}, our pipeline does not require a predefined list of biases but proposes a set of novel domain-specific biases. }
    \label{fig:teaser}
      \vspace{-0.45cm}
\end{figure}
\section{Introduction}
\label{sec:intro}

\blfootnote{\textsuperscript{$^\dagger$}Corresponding authors.} 

Text-to-Image (T2I) generation has become increasingly popular, thanks to its intuitive conditioning and the high 
quality and fidelity of the generated content
~\cite{saharia2022photorealistic, nichol2021glide, hierarchical_text_to_image_CLIP, LDM_2022_CVPR, podell2023sdxl}. 
Several works extended the base T2I model, unlocking additional use cases, including personalization \cite{Ruiz_2023_CVPR, gal2022textual}, image editing  \cite{epstein2023diffusion, brooks2023instructpix2pix, hertz2022prompt, goel2023pairdiffusion}, and various forms of conditioning \cite{avrahami2023spatext, ControlNet_2023_ICCV, huang2023composer}. This rapid progress urges 
to investigate other key aspects beyond image quality improvements, 
such as their fairness and potential bias perpetration
~\cite{ITIGEN_2023_ICCV, fairDiffusion2023, DALL_EVAL_2023_ICCV}. It is widely acknowledged that deep learning models learn the underlying biases present in their training sets~\cite{NIPS2016_bias_word_embeddings, Bias_Women_also_Snowboard_2018_ECCV, zhao2017men}, and generative models are no exception~\cite{DALL_EVAL_2023_ICCV, ITIGEN_2023_ICCV, fairDiffusion2023, naik2023social}.

Ethical topics such as fairness and biases have seen many definitions and frameworks \cite{verma2018fairness}; defining them comprehensively poses a challenge, as interpretations vary and are subjective to the individual user. 
Following previous works \cite{fairDiffusion2023, xu2018fairgan}, a model is considered unbiased regarding a specific concept if, given a context $t$ that is agnostic to class distinctions, the possible classes $c \in \mathcal{C}$ exhibit a uniform distribution.
In practice, for a T2I model, this reflects to 
the tendency of the generator to produce content of a certain class $c$ (\eg \emph{``man"}), given a textual prompt $t$ that does not 
specify the intended class (\eg \emph{``A picture of a doctor"}).

Several works studied 
bias mitigation in pre-trained models, by 
introducing training-related methods~\cite{nam2020learning, savani2020intra, Wang_2020_CVPR, jung2022learning} or 
using data augmentation techniques~\cite{Agarwal_2022_WACV, D'Inca_2024_WACV}. Nevertheless, a notable limitation of these approaches is their dependence on a predefined set of biases, such as 
gender, age, and race~\cite{DALL_EVAL_2023_ICCV, fairDiffusion2023}, as well as specific face attributes~\cite{ITIGEN_2023_ICCV}. While these represent perhaps the most sensitive biases, we argue that there could be biases that remain undiscovered and unstudied. 
Considering the example in Fig.\ref{fig:teaser}, the prompt \myquote{A person using a laptop} does not specify the person's appearance and neither 
the specific laptop nor the scenario.  While closed-set pipelines can detect well-known biases (\eg gender, race), 
the T2I model may exhibit biases also for other elements (\eg laptop brand, office). Thus, an open research question is: \emph{Can we identify arbitrary biases present in T2I models given only prompts and no pre-specified classes?} This is challenging as collecting annotated data for all potential biases is prohibitive.



Toward this goal, 
we propose \textit{OpenBias}, 
the
first 
pipeline that operates in an \textit{open-set scenario}, enabling to identify, 
recognize, 
and quantify biases in a specific T2I model without 
constraints (or data collection) for a specific predefined set. 
Specifically, we exploit the multi-modal nature of T2I models and create a knowledge base of possible biases given a collection of target textual captions, by 
querying a Large Language Model (LLM). In this way, we 
discover specific biases 
for the given captions. Next, we need to recognize whether these biases are actually present in the images. For this step, we leverage available 
Visual Question Answering (VQA) models, directly using them 
to assess the bias presence. By doing this, we overcome the limitation of using attributes-specific classifiers as done in previous works~\cite{fairDiffusion2023, ITIGEN_2023_ICCV, su2023unbiased}, which is not efficient nor feasible in an open-set scenario. Our pipeline is modular and flexible,  allowing for the seamless replacement of each component with newer or domain-specific versions as they become available. 
Moreover, we treat the generative model as a \textit{black box}, querying it with specific prompts to mimic end-user interactions (\ie without control over training data and algorithm). 
We test OpenBias on variants of Stable Diffusion~\cite{podell2023sdxl, LDM_2022_CVPR} showing human-agreement, model-level comparisons, and the discovery of novel biases.

\noindent\textbf{Contributions.} To summarize, our key contributions are:

\begin{itemize}
    \item To the best of our knowledge, we are the first to study the problem of open-set bias detection 
    at large scale 
    without relying on a predefined list of biases. Our method discovers novel biases that have never been studied before. 
    \item We propose OpenBias, a modular pipeline, that, given a list of prompts, leverages a Large Language Model to extract a knowledge base of possible biases, and a Vision Question Answer model to recognize and quantify them. 
    \item We test our pipeline on multiple text-to-image generative models: Stable Diffusion XL, 1.5, 2~\cite{podell2023sdxl, LDM_2022_CVPR}. We assess our pipeline showing its agreement with closed-set classifier-based methods and with human judgement. 
\end{itemize}
\section{Related work}
\para{Pipeline with Foundation Models} We broadly refer to foundation models~\cite{bommasani2021opportunities} as large-scale deep learning models trained on extensive data corpora, usually with a self-supervised objective \cite{bommasani2021opportunities}. 
This approach has been used across different modalities, such as text~\cite{touvron2023llama, NEURIPS2020_gpt_3_few_shot}, vision~\cite{oquab2023dinov2, caron2021emerging, dosovitskiy2020image} and multi-modal models \cite{zhu2023chatgpt_blip2, liu2023llava, radford2021learning}. These models can be fine-tuned on downstream tasks or applied in a zero-shot manner, {generalizing to unseen tasks}~
\cite{NEURIPS2020_gpt_3_few_shot, NEURIPS2022_Chain_of_Thougth, subramanian2022reclip}. 

Lately, several works 
combined different foundation models to solve complex tasks. 
\cite{gupta2023visual,suris2023vipergpt} 
use an LLM to generate 
Python code that invokes vision-language models to produce results. 
TIFA~\cite{TIFA_2023_ICCV} 
assesses the faithfulness of a generated image to a given text prompt, by querying 
a VQA model with questions produced by an LLM from the 
original caption. Similarly, \cite{zhu2023chatgpt_blip2,chen2023video} enhance 
image/video captioning by iteratively querying an LLM to ask 
questions to a VQA model. 
Differently, \cite{kim2023biastotext} identify spurious correlations in synthetic images via captioning and language interpretation, but without categorizing or quantifying bias. 

{\it We share a similar motivation, i.e., we leverage powerful foundation models to build an automatic pipeline, tailored to the novel task of open-set bias discovery.} 
OpenBias builds a knowledge base of biases leveraging the domain-specific knowledge from \textit{real} captions and LLMs.

\para{Bias Mitigation in Generative Models}
Bias mitigation is a long-studied topic in generative models. A substantial line of work focused on GAN-based methods. Some works improve fairness at inference time by altering the latent space semantic distribution~\cite{tan2021improving} or by gradient clipping to control the gradient ensuring fairer representations for sensitive groups~\cite{kenfack2022repfairgan}. The advent of T2I generative models has directed research efforts towards fairness within this domain. FairDiffusion~\cite{fairDiffusion2023} 
guides Stable Diffusion~\cite{LDM_2022_CVPR} toward fairer generation in job-related contexts. It enhances classifier-free guidance~\cite{ho2022classifier} by adding a fair guidance term based on user-provided fair instructions. Similarly, \cite{brack2023mitigating} demonstrates that 
(negative) prompt and semantic guidance
~\cite{NEURIPS2023_4ff83037} 
mitigate inappropriateness generation in several T2I models. Given handwritten text as input, \textit{ITI-GEN}~\cite{ITIGEN_2023_ICCV} 
enhances the fairness of 
T2I generative models through prompt learning. To improve fairness, \cite{su2023unbiased}  guide generation using the data manifold of the training set, estimated via unsupervised learning. 

{\it While yielding 
notable result, these bias mitigation methods 
rely on predefined lists of biases. 
Here, we argue that there may exist other biases not considered by these methods. Therefore, our proposed pipeline is orthogonal, providing a valuable tool to enhance their utility.}
\begin{figure*}[ht!]
    \centering
    \includegraphics[width=\textwidth]
    {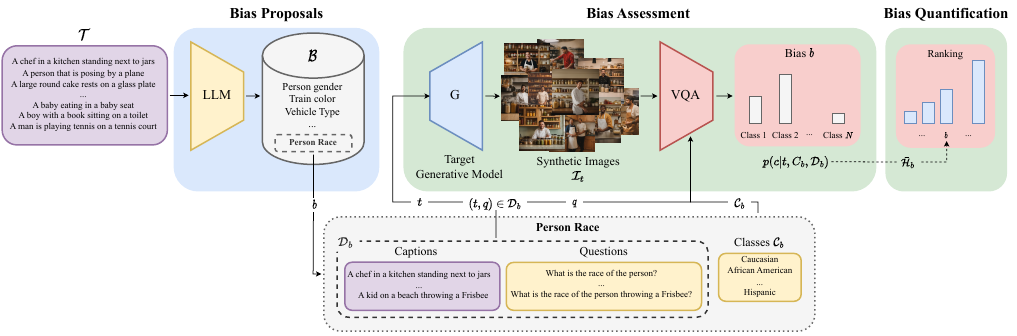}
        \vspace*{-0.5cm}
    \caption{OpenBias pipeline. Starting with a dataset of real textual captions ($\mathcal{T}$) we leverage a Large Language Model (LLM) to build a knowledge base $\knowledgebase$ of possible biases that may occur during the image generation process. In the second stage, synthesized images are generated using the target generative model conditioned on captions where a potential bias has been identified. Finally, the biases are assessed and quantified by querying a VQA model with caption-specific questions extracted during the bias proposal phase.}
    \label{fig:method}
        \vspace*{-0.3cm}
\end{figure*}

\section{OpenBias}
\label{sec:method}
This section presents OpenBias, our pipeline for proposing, assessing, and quantifying biases in T2I generative models. 
The overview of the proposed framework is outlined in \cref{fig:method}. 
Starting from a dataset of real textual captions, we leverage a Large Language Model (LLM) to build a knowledge base of possible biases that may occur during image generation. This process enables the identification of domain-specific biases unexplored up to now.  In the second stage, we synthesize images using the target generative model, conditioned on captions where a potential bias has been identified.  
Lastly, we 
assess the biases with a VQA model, 
querying it 
with caption-specific questions generated during the bias proposal phase.

\subsection{Bias Proposals}
\label{sec:bias_proposal}
Given a dataset of real captions $\datasetcaptions$, we construct a knowledge base $\knowledgebase$ of possible biases. For each caption in the dataset, we task a LLM with providing three outputs: the potential bias name, a set of classes associated with the bias, and a question to identify the bias. 

Formally, given a caption $\singlecaption \in \datasetcaptions$, let us denote the LLM's output as a set of triplets $\mathtt{L}_t=
\{(\biasname^t_i, \mathcal{C}^t_i, q^t_i)\}_{i=1}^{n_t}$ 
where the cardinality of the set $n_t$ is caption dependent, and each triplet $(\biasname, \mathcal{C}, q)$ has a proposed bias $\biasname$, 
 a set of associated classes $\mathcal{C}$, and the question $q$ assigned to the specific caption $t$. 
 To obtain this set, we propose to use in-context learning \cite{NEURIPS2020_gpt_3_few_shot, ICLR_2023_selective_annotations}, providing task description and demonstrations directly in the textual prompt.\footnote{We refer the reader to the {\supp} for system prompt details.}
We build the knowledge base $\knowledgebase$ by aggregating the per-caption information on the whole dataset. 
Specifically, we can define the set of caption-specific biases $B_t$ as the union of its potential biases, \ie $B_t=\bigcup_{i=1}^{n_t} b_i$. The dataset-level set of biases is then the union of the caption-level ones, \ie $\mathcal{B}=\bigcup_{t\in\mathcal{T}} B_t$. 
Next, we aggregate the bias-specific information across the whole dataset. 
We define the database of captions and questions as
\begin{equation}
    \mathcal{D}_b = \{ (t,q) \mid \forall t \in \mathcal{T},  (x,\mathcal{C},q)\in \mathtt{L}_t, x=b\}.
\end{equation}
$\mathcal{D}_b$ collects captions and questions specific to the bias $b$. Moreover, we define $\mathcal{T}_b = \{t \mid (t,q)\in\ \mathcal{D}_b\}$ as the set of captions, and 
$\mathcal{C}_b$ is the union of the set of classes associated to the bias $b$ in $\mathcal{T}$.
Nevertheless, $\mathcal{D}_b$ does not account for the potential specification of the classes of $b$ in the caption. For instance, if we aim to generate ``\textit{An image of a large dog}", the dog's size should not be included among the biases. 
To address this, we implement a two-stage filtering procedure of $\mathcal{D}_b$. First, given a pair $(t,q)\in\mathcal{D}_b$ we ask the LLM to output whether the answer to the question $q$ is explicitly present in the caption $t$. Secondly, we leverage ConceptNet \cite{ConceptNET} to identify synonyms for the classes $\mathcal{C}_b$ related to the specific bias $b$, and filter out the captions in containing either a class $\mathcal{C}_b$ or its synonyms. We empirically observe that combining these two stages produces more robust results. 

By executing the aforementioned steps, we generate bias proposals in an open-set manner tailored to the given dataset. In the following sections, we elaborate on the process of bias quantification in a target generative model.

\subsection{Bias Assessment and Quantification}\label{sec:bias_assessment_and_quantification}

Let $\generator$ be the target T2I generative model. 
Our objective is to evaluate if $\generator$ generates images with the identified biases. 
Given a bias $b \in \knowledgebase$ and a caption $t\in\mathcal{T}_b$, we generate the set of $N$ images $\mathcal{I}^t_b$ as 
\begin{equation}
    \mathcal{I}^t_b = \{\generator(t,s) | \forall s\in{S}\}
\end{equation}
where $S$ is the set of sampled random noise, of cardinality $|S|=N$. Sampling multiple noise vectors allows us to obtain a distribution of the $\generator$ output on the same prompt $t$.

To assess the bias within $\mathcal{I}^t_b$, we propose to leverage a state-of-the-art Vision Question Answering (VQA) model $\mathtt{VQA}$ mapping images and questions to answers in natural language. The VQA processes the images $\mathcal{I}^t_b$, and their associated question $q$ in the pair $(t,q)\in \mathcal{D}_b$,  
 choosing an answer from the possible classes $\mathcal{C}_b$. Formally, given an image $I\in \mathcal{I}_b^t$ we denote the predicted class as 
\begin{equation}
    \label{eq:vqa}
    \hat{c} = \mathtt{VQA}(I, q, \mathcal{C}_b).
\end{equation}
With this score, we gather statistics on the distribution of the classes on a set of images, and use them to quantify the severity of the bias. 
In the following, we investigate two distinct scenarios, namely \emph{context-aware}, where we analyze the bias on caption-specific images $\mathcal{I}_b^{t}$, and \emph{context-free}, where we consider the whole set of images $\mathcal{I}_b$ associated to one bias $b\in\mathcal{B}$. 

\subsubsection{Context-Aware Bias}
\label{sec:context_aware}

As discussed in \Cref{sec:intro}, our focus lies in examining bias exclusively when the classes are not explicitly mentioned in the caption. The bias proposals pipeline described in \cref{sec:bias_proposal} filters out such cases; nevertheless, there could be additional aspects within the caption that impact the outcome. For example, the two captions \textit{``A military is running"} and \textit{``A person is running"} are both agnostic to the bias \textit{``person gender"}, but the direction and magnitude of the bias may be very different in the two cases. 
To consider the role of the context in the bias assessment, we collect statistics at the caption level, analyzing the set of images $\mathcal{I}_b^t$ produced from a specific caption $t \in \mathcal{T}$. Given a bias $b$ we compute the probability for a class $c\in\mathcal{C}_b$ as:

\begin{equation}
    \label{eq:context_aware}
    p(c | t,\mathcal{C}_b, \mathcal{D}_b) = \dfrac{1}{|\mathcal{I}_b^t|} \sum_{I\in \mathcal{I}_b^t}\indicatorfunction \bigl(\hat{c}=c\bigr)
\end{equation}
with $\hat{c} = \mathtt{VQA}(I, q, \mathcal{C}_b)$ the prediction of the VQA as defined in \cref{eq:vqa}, and $\indicatorfunction(\cdot)$ the indicator function. 

\subsubsection{Context-Free Bias}

Differently from the context-aware scenario, our interest lies in characterizing the overall behavior of the model $\generator$. This is crucial as it offers valuable insights into aspects such as the majority class (\ie the direction toward which the bias tends) and the overall intensity of the bias. To effectively exclude the role of the context in the captions, we propose to average the VQA scores for $c\in\mathcal{C}_b$ over all captions $t$ related to that bias $b\in\mathcal{B}$:
\begin{equation}
    \label{eq:context_free}
    p(c | \mathcal{C}_b, \mathcal{D}_b) =\dfrac{1}{|\mathcal{D}_b|}\sum_{(t,q)\in \mathcal{D}_b} p(c | t, \mathcal{C}_b, \mathcal{D}_b) 
\end{equation}
Note that the context-aware bias is a special case of this scenario, where $\mathcal{D}_b$ has a single instance, \ie $\mathcal{D}_b=\{ (t,q)\}$. 

\subsubsection{Bias Quantification and Ranking}
\label{sec:bias_quantification}

After collecting the scores for each individual attribute class $c\in\mathcal{C}_b$, we can aggregate them to rank the severity of biases within the generative model.
As mentioned in \cref{sec:intro}, we follow existing work \cite{fairDiffusion2023, xu2018fairgan} and consider the model $\generator$ as unbiased with respect to a concept $b$ when the distribution of the possible classes $c\in \mathcal{C}_b$ is uniform. 
To quantitatively assess the severity of the bias, we compute the entropy of the probability distribution of the classes obtained using either \cref{eq:context_aware} or \cref{eq:context_free}. To compare biases with different numbers of classes, we normalize the entropy by the maximum possible entropy \cite{wilcox1967indices}. Additionally, we adjust the score for enhanced human readability. In practice, our bias severity score is defined as follows: 
\begin{equation}
    \label{eq:entropy_score}
    \bar{\mathcal{H}}_b = 1 + \dfrac{\sum_{c\in\mathcal{C}_b} p(c | \mathcal{C}_b, \mathcal{D}_b) \log p(c | \mathcal{C}_b, \mathcal{D}_b) }{\log(|\mathcal{C}_b|)} 
\end{equation}
The resulting score is always bounded $\bar{\mathcal{H}}_b \in [0,1]$, where 0 
indicates an unbiased concept 
while 1 a biased one. 

We note that, while we focused our pipeline on conditional generative models, our model can be easily extended 
for studying biases in both real-world multimodal datasets (\eg by assuming images $\mathcal{I}_b^t$ are provided rather than generated), and to unconditional generative models (\ie by using a captioning system on their outputs as set $\mathcal{T}$). We refer the reader to the {\supp} for details where we will also show an analysis between the unconditional GAN \textit{StyleGAN3}~\cite{NEURIPS2021_076ccd93} and its training set FFHQ~\cite{Karras_2019_CVPR}. 

\begin{figure*}[!ht]
    \centering
    \includegraphics[width=0.86\textwidth]{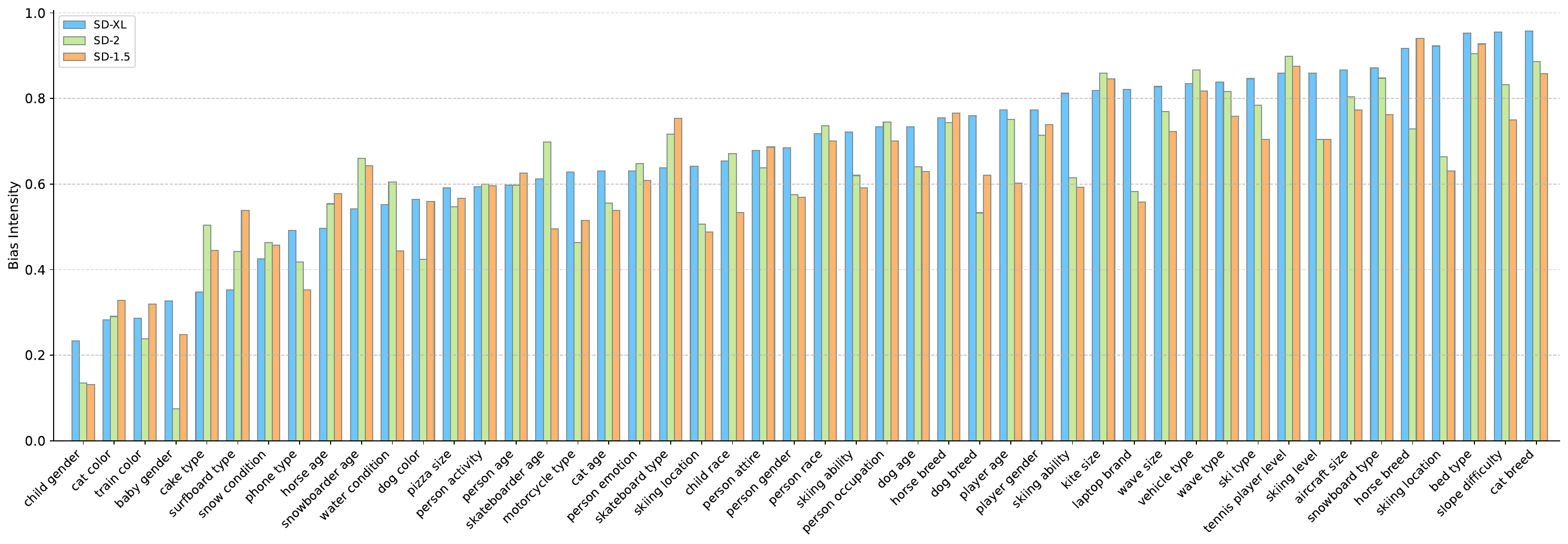}
    \vspace*{-0.4cm}
    \caption{Comparison of context-aware discovered biases on Stable Diffusion XL, 2 and 1.5~\cite{podell2023sdxl, LDM_2022_CVPR} with captions from COCO~\cite{DBLP:journals/corr/LinMBHPRDZ14}.}
       \label{fig:comparison_coco}
      \vspace*{-0.2cm}
\end{figure*}

\begin{figure*}[!ht]
    \centering
    \includegraphics[width=0.86\linewidth]{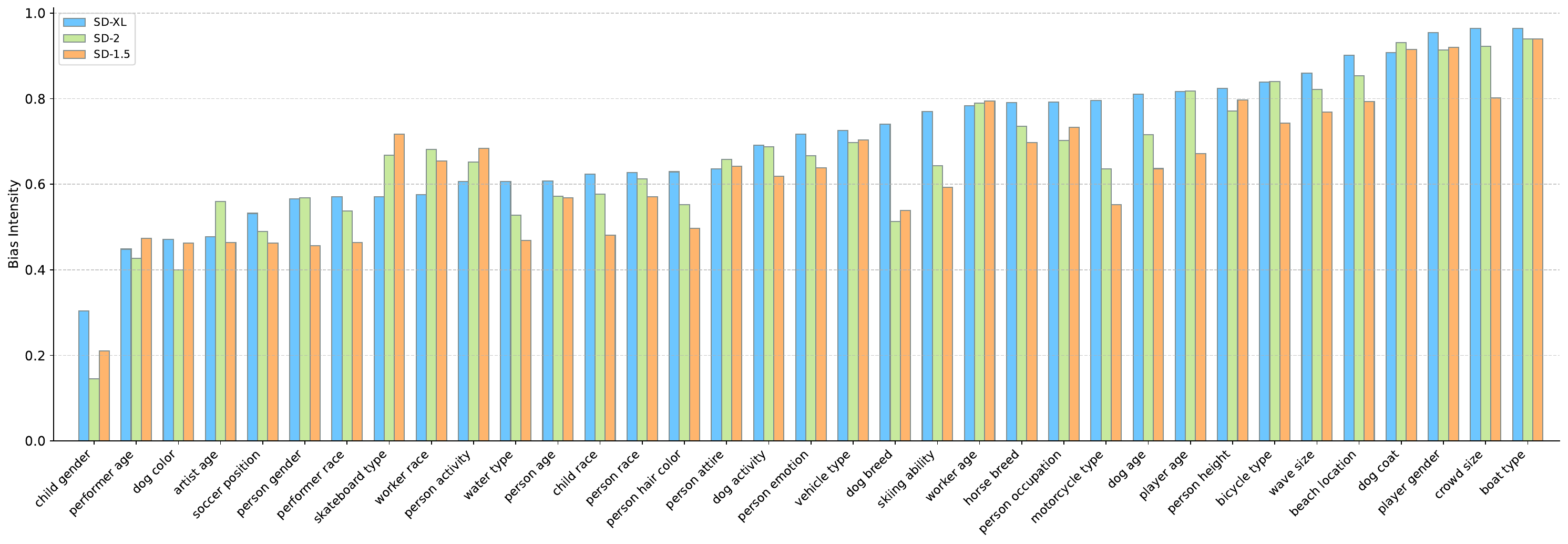}
    \vspace*{-0.4cm}
    \caption{Comparison of context-aware found biases on Stable Diffusion XL, 2 and 1.5~\cite{podell2023sdxl, LDM_2022_CVPR} on captions from Flick30k~\cite{young-etal-2014-image}.}
    \vspace*{-0.4cm}
      \label{fig:comparison_flickr}
\end{figure*}

\section{Experiments}
In this section, we conduct a series of experiments to assess the proposed framework quantitatively. In \cref{sec:implementation_details}, we provide 
implementation details 
and the preprocessing steps applied to the datasets.
In \cref{sec:quantiative}, we 
quantitative evaluate OpenBias on 
two directions, (i) 
comparing it with 
a state-of-the-art classifier-based method on a closed set of well-known social biases, 
(ii) testing the agreement between OpenBias and human judgment via a user study.

\subsection{Pipeline Implementation}
\label{sec:implementation_details}
\noindent \textbf{Datasets.} We study the bias in two multimodal datasets Flickr 30k~\cite{young-etal-2014-image} and COCO~\cite{DBLP:journals/corr/LinMBHPRDZ14}.  Flickr30k~\cite{young-etal-2014-image} comprises $30$K images with $5$ caption per image, depicting images in the wild.
Similarly, COCO~\cite{DBLP:journals/corr/LinMBHPRDZ14} is a large-scale dataset containing a diverse range of images that capture everyday scenes and objects in complex contexts. We filter this dataset, creating a subset of images whose caption contains a single person. This procedure results in roughly $123$K captions. Our choice is motivated 
by 
building a large subset of captions specifically tied to people. This focus on the person-domain is crucial as it represents one of the most sensitive scenarios for exploring bias-related settings. Nevertheless, it is worth noting that the biases we discover within this context 
extend beyond person-related biases to include objects, animals, and actions associated with people. Further details are highlighted in \cref{sec:comparison_ranking}.

\noindent \textbf{Implementation Details.} Our pipeline is designed to be flexible and modular, enabling us to replace individual components as needed. In this study, we leverage LLama2-7B \cite{touvron2023llama} as our foundation LLM. This model is exploited to build the knowledge base of possible biases, as described in \cref{sec:bias_proposal}. We refer the reader to the {\supp} for details regarding the prompts and examples we use to instruct LLama to perform the desired tasks. 
To assess the presence of the bias, we rely on state-of-the-art Visual Question Answering (VQA) models. From our evaluation outlined in \cref{sec:quantiative}, Llava1.5-13B \cite{liu2023llava, liu2023improvedllava} emerges as the top-performing, thus we adopt it as our default VQA model. Finally, we conduct our study by randomly selecting $100$ captions associated with each bias and generating $N=10$ images for each caption using a different random seed. In this way, we obtain a set of $1000$ images, that we use to study the context-free and context-aware bias of the target generative model.

\subsection{Quantitative Results}
\label{sec:quantiative}
Our open-set setting harnesses the zero-shot performance of each component. 
As in~\cite{fairDiffusion2023}, we evaluate OpenBias 
using FairFace~\cite{Karkkainen_2021_WACV}, a well-established 
classifier fairly trained, 
as the ground truth on gender, age, and race. 
While FairFace treats socially sensitive attributes as closed-set, 
we uphold our commitment to inclusivity by also evaluating OpenBias with self-identified ones, reported in the {\supp}.

\begin{table}[t]
    \centering
    \resizebox{\columnwidth}{!}{
    \begin{tabular}{lcccccc}
        \toprule
        \multirow{2}{*}{\textbf{Model}} & \multicolumn{2}{c|}{\textbf{Gender}} & \multicolumn{2}{c|}{\textbf{Age}} & \multicolumn{2}{c}{\textbf{Race}} \\ \cmidrule{2-7}
                               & Acc. & \multicolumn{1}{c|}{F1} & Acc. & \multicolumn{1}{c|}{F1} & Acc. & \multicolumn{1}{c}{F1} \\
        \midrule
         CLIP-L~\cite{radford2021learning}                          & 91.43          & 75.46             & 58.96             & 45.77             & 36.02             & 33.60             \\
         OFA-Large~\cite{wang2022ofa}                               & \textbf{93.03} & 83.07             & 53.79             & 41.72             & 24.61             & 21.22              \\
         mPLUG-Large~\cite{li2022mplug}                             & \textbf{93.03} & 82.81             & 61.37             & 52.74             & 21.46             & 23.26              \\
         BLIP-Large~\cite{li2022blip}                               & 92.23          & 82.18             & 48.61             & 31.29             & 36.22             & 35.52              \\
          Llava1.5-7B~\cite{liu2023improvedllava,liu2023llava}      & 92.03          & 82.33             & 66.54             & 62.16             & 55.71             & 42.80             \\
          \rowcolor{gray!20} Llava1.5-13B~\cite{liu2023improvedllava,liu2023llava}     & 92.83          & \textbf{83.21}    & \textbf{72.27}    & \textbf{70.00}    & \textbf{55.91}    & \textbf{44.33}    \\
        \bottomrule
    \end{tabular}}
    \vspace{-5pt}
    \caption{VQA evaluation on the generated images using COCO captions. We highlight in \colorbox{gray!20}{gray} the chosen default VQA model. }
    \vspace{-10pt}
    \label{tab:VQA_eval_coco}
\end{table}

\begin{table}[!ht]
    \scriptsize
    \resizebox{\columnwidth}{!}{
    \begin{tabular}{l|cccccc}
    \toprule
    \multirow{2}{*}{\textbf{Model}}    & \multicolumn{3}{c|}{\textbf{Flickr 30k}~\cite{young-etal-2014-image}}                 & \multicolumn{3}{c}{\textbf{COCO}~\cite{DBLP:journals/corr/LinMBHPRDZ14}}    \\ 
    \multicolumn{1}{l|}{}                     & gender  & age     & \multicolumn{1}{c|}{race}    & gender  & age     & race    \\ \midrule
    Real                                         & 0 & 0.032 & \multicolumn{1}{c|}{0.030} & 0 & 0.041 & 0.028 \\
    SD-1.5~\cite{LDM_2022_CVPR}                                      & 0.072 & 0.032 & \multicolumn{1}{c|}{0.052} & 0.075 & 0.028 & 0.092 \\
    SD-2~\cite{LDM_2022_CVPR}                                        & 0.036 & 0.069 & \multicolumn{1}{c|}{0.047} & 0.060 & 0.045 & 0.105 \\ 
    SD-XL~\cite{podell2023sdxl}                                        & 0.006 & 0.028 & \multicolumn{1}{c|}{0.180} & 0.002 & 0.027 & 0.184 \\ \bottomrule
    \end{tabular}}
    \vspace{-5pt}
    \caption{KL divergence ($\downarrow$) computed over the predictions of Llava1.5-13B and FairFace on generated and real images.}
    \vspace{-0.5cm}
    \label{tab:FairFace_eval_KL}
\end{table}

\para{Agreement with FairFace} We compare the predictions of multiple SoTA Visual Question Answering models with FairFace. 
Firstly, we assess the zero-shot performance of the VQA models on synthetic images, performing our comparisons using images generated by SD XL. The evaluation involves assessing accuracy and F1 scores, which are computed against FairFace predictions treated as the ground truth. The results are reported in \cref{tab:VQA_eval_coco}. Llava1.5-13B emerges as the top-performing model across different tasks, consequently, we employ it as our default VQA model.

Next, we evaluate the agreement between Llava and FairFace~\cite{Karkkainen_2021_WACV} on different scenarios. Specifically, we run the two models on real and synthetic images generated with Stable Diffusion 1.5, 2, and XL. We measure the agreement between the two as the KL Divergence between the probability distributions obtained using the predictions of the respective model. We report the results in~\cref{tab:FairFace_eval_KL}. We can observe that the models are highly aligned, obtaining low KL scores, proving the VQA model's robustness in both generative and real settings. \supp\ provides a more comprehensive evaluation of the VQA. 

\begin{figure}[t]
\centering
    \includegraphics[width=0.95\linewidth]{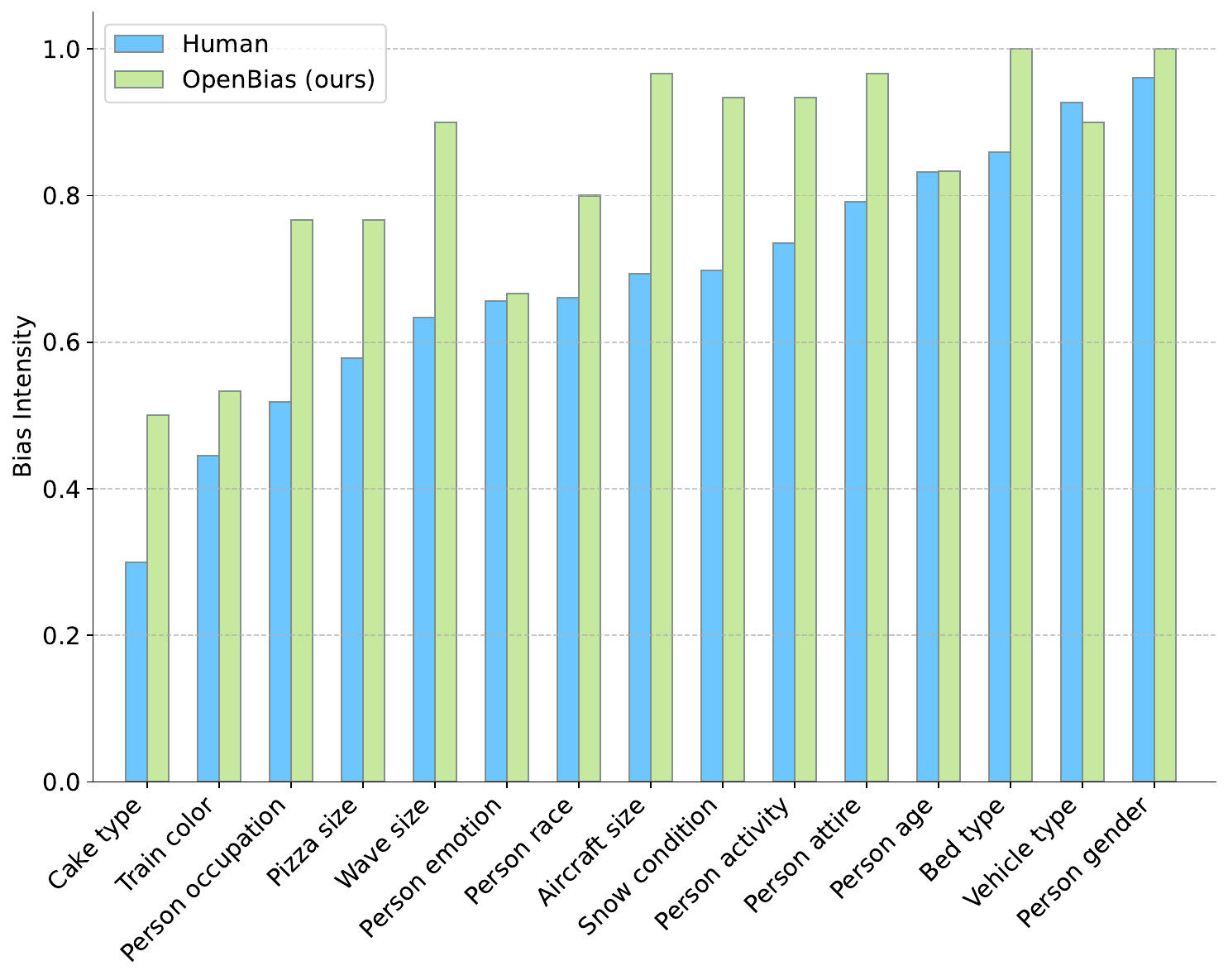}
    \vspace{-0.5cm}
    \caption{Human evaluation results.}
    \label{fig:human_eval}
    \vspace{-0.5cm}
\end{figure}

\noindent \textbf{User Study.} We conduct a human evaluation of the proposed pipeline at the context-aware level, to assess its alignment with human judgment. 
The study presents $10$ images generated from the same caption for each bias. We use public crowdsourcing platforms, without geographical restrictions, and randomizing the questions’ order. Each participant is asked to identify the direction (majority class) of each bias and its intensity in a range from $0$ to $10$. The option \textit{``No bias"} is provided to capture the instances where no bias is perceived, corresponding to a bias intensity of $0$. We conduct the user study on a subsection of the biases, resulting in $15$ diverse object-related and person-related biases and $390$ diverse images. We collect answers from $55$ unique users, for a total of $2200$ valid responses. The user study results are shown in \cref{fig:human_eval}, where we compare the bias intensity as collected from the human participants with the severity score computed with OpenBias. We can observe that there is a high alignment on various biases such as \textit{``Person age"}, \textit{``Person gender"}, \textit{``Vehicle type"}, \textit{``Person emotion"} and \textit{``Train color"}. We compute the Absolute Mean Error (AME) between the bias intensity produced by the model and the average user score, resulting in an $\text{AME} = 0.15$. Furthermore, we compute the agreement on the majority class, \ie the direction of the bias. In this case, OpenBias matches the collected human choices $67\%$ of the cases. We remark that concepts of bias and fairness are highly subjective, and this can introduce further errors in the evaluation process. Nevertheless, our results 
show a correlation between the scores, validating our pipeline. 

\begin{figure*}[p]
    \centering
    \begin{subfigure}{0.33\linewidth}
        \centering
        \textbf{Train color}\par\medskip
        \vspace*{-0.12cm}
        \includegraphics[width=\linewidth]{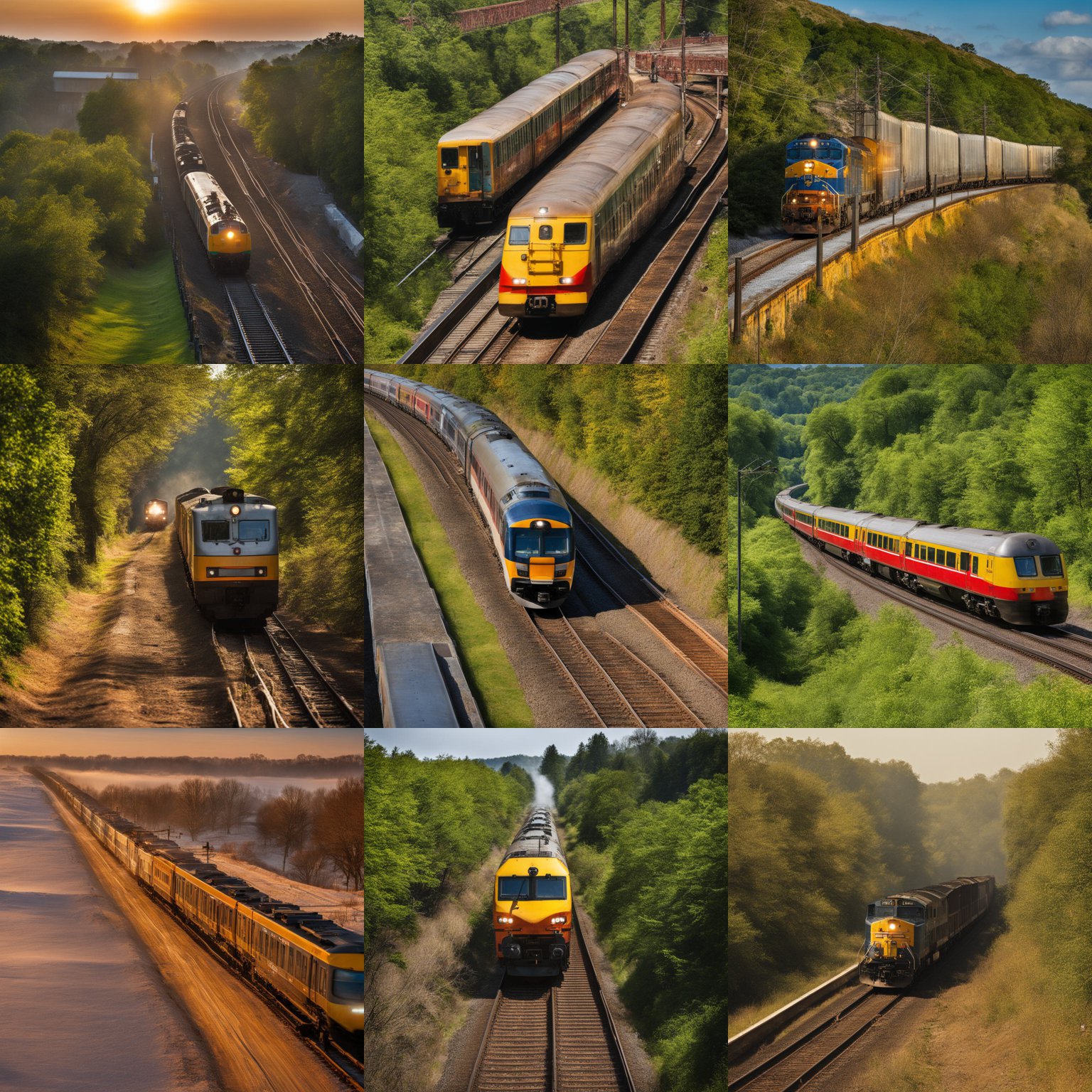}
        \captionsetup{labelformat=empty}
        \caption{``A train zips down the railway in the sun"}
    \end{subfigure}
    \hfill
    \begin{subfigure}{0.33\linewidth}
        \centering
        \textbf{Laptop brand}\par\medskip
        \vspace*{-0.12cm}
        \includegraphics[width=\linewidth]{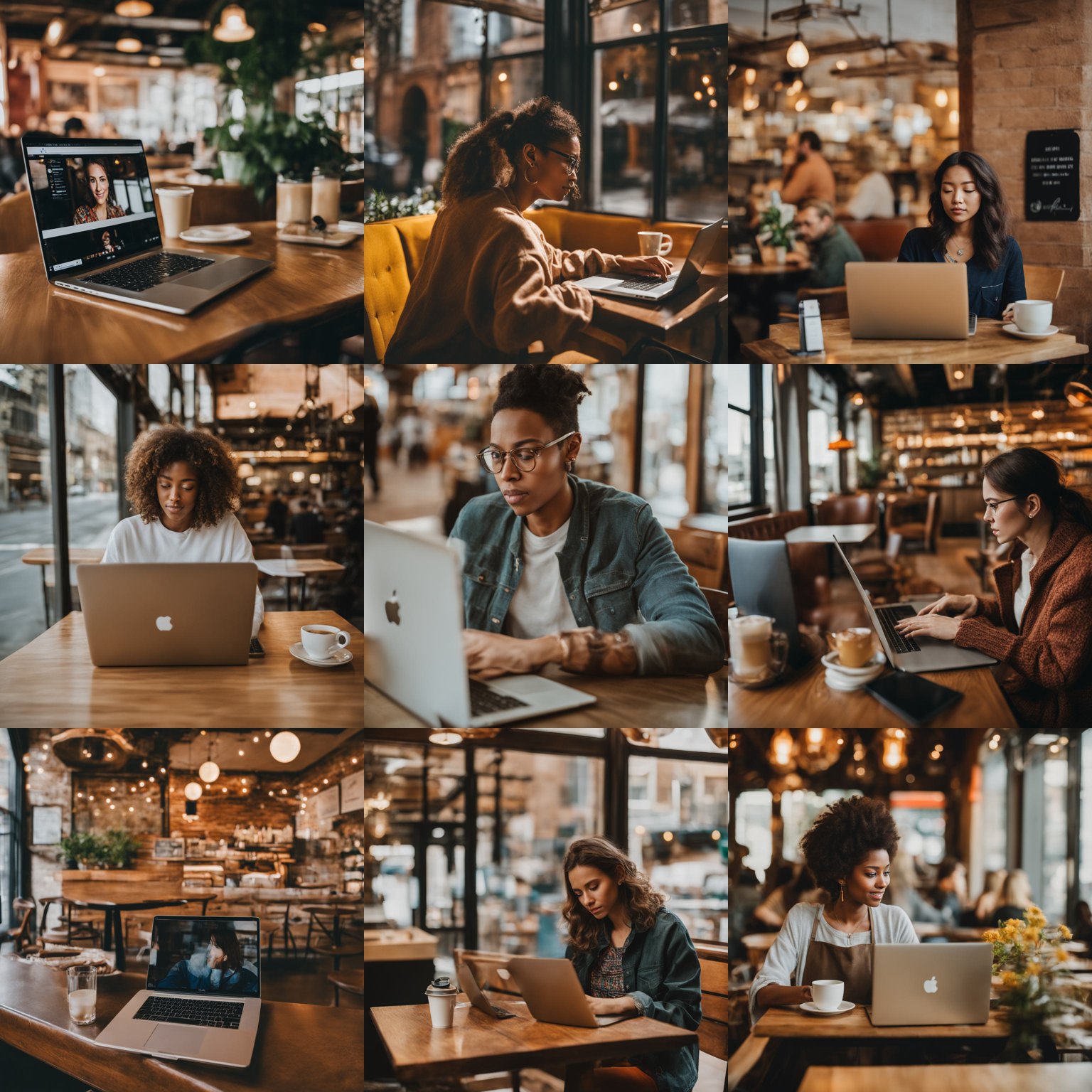}
        \captionsetup{labelformat=empty}
        \caption{``A photo of a person on a laptop in a coffee shop"}
    \end{subfigure}
    \hfill
    \begin{subfigure}{0.33\linewidth}
        \centering
        \textbf{Horse breed}\par\medskip
        \vspace*{-0.12cm}
        \includegraphics[width=\linewidth]{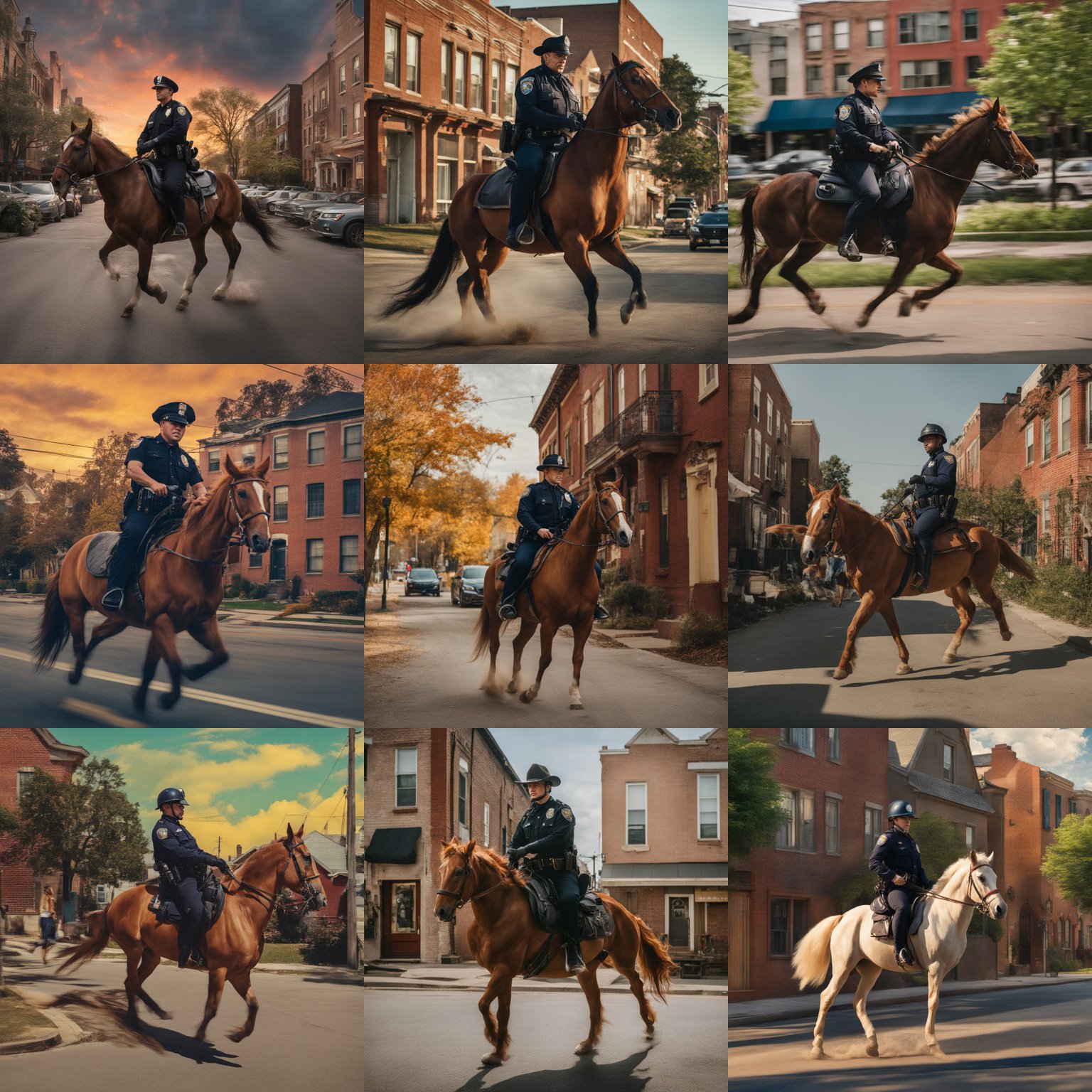}
        \captionsetup{labelformat=empty}
        \caption{``A cop riding a horse through a city neighborhood"}
    \end{subfigure}
    \caption{Novel biases discovered on Stable Diffusion XL~\cite{podell2023sdxl} by OpenBias.}
    \label{fig:qualitatives_others}
\end{figure*}
\begin{figure*}[htbp]
    \centering
    \begin{subfigure}{0.33\linewidth}
        \centering
        \textbf{Child gender}\par\medskip
        \vspace*{-0.12cm}
        \includegraphics[width=\linewidth]{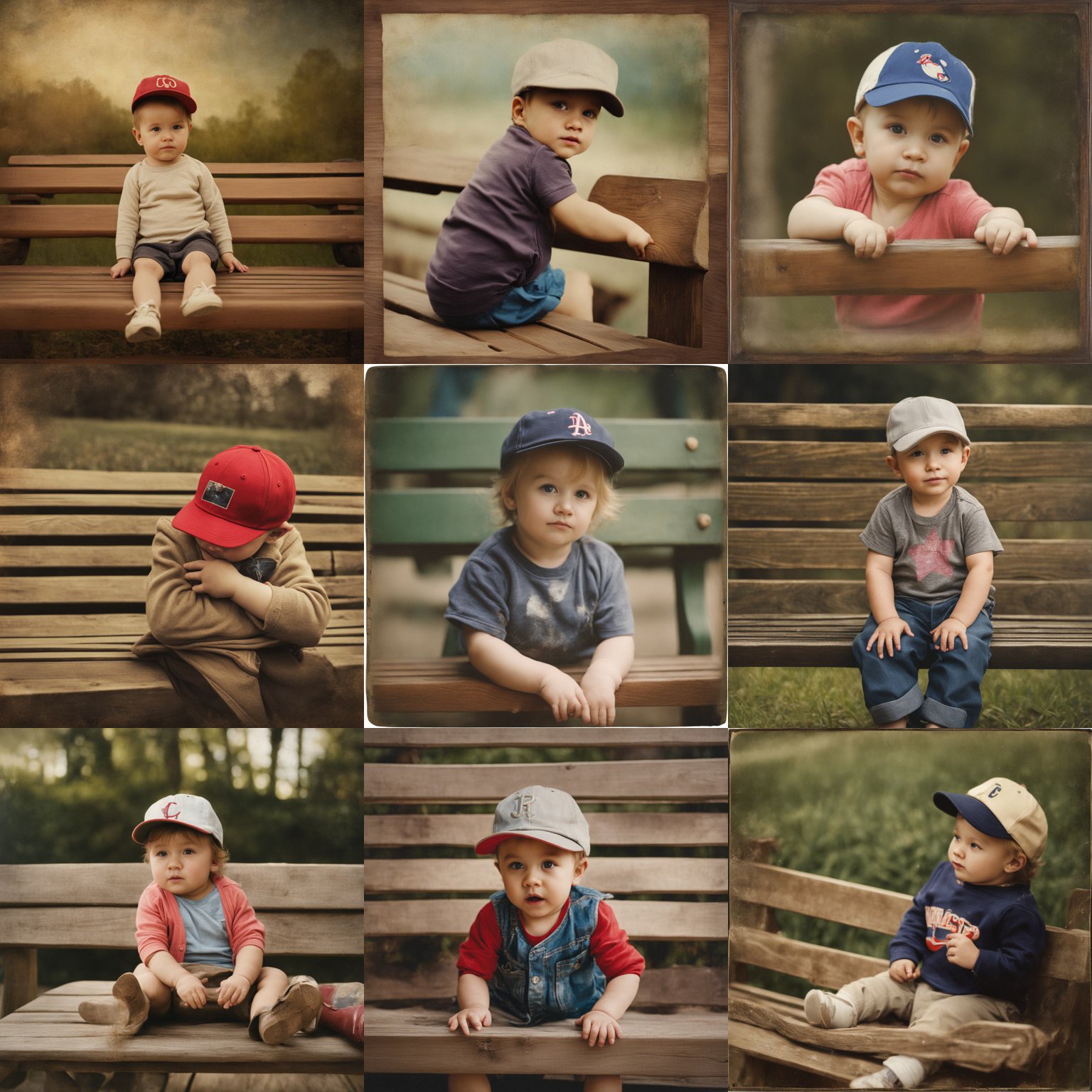}
        \captionsetup{labelformat=empty}
        \caption{``Toddler in a baseball cap on a wooden bench"}
    \end{subfigure}
    \hfill
    \begin{subfigure}{0.33\linewidth}
        \centering
        \textbf{Child race}\par\medskip
        \vspace*{-0.12cm}
        \includegraphics[width=\linewidth]{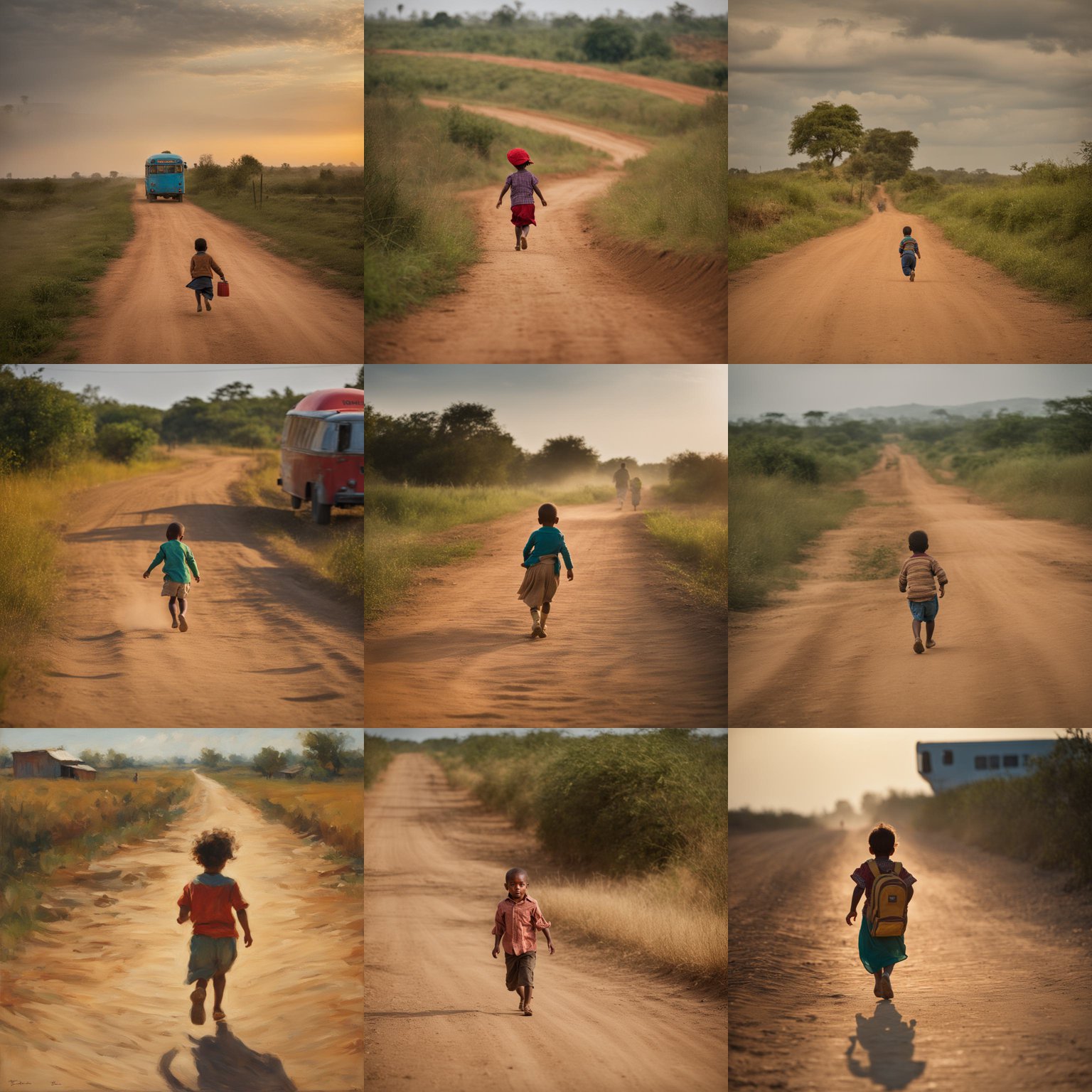}
        \captionsetup{labelformat=empty}
        \caption{``Small child hurrying toward a bus on a dirt road"}
    \end{subfigure}
    \hfill
    \begin{subfigure}{0.33\linewidth}
        \centering
        \textbf{Person attire}\par\medskip
        \vspace*{-0.12cm}
        \includegraphics[width=\linewidth]{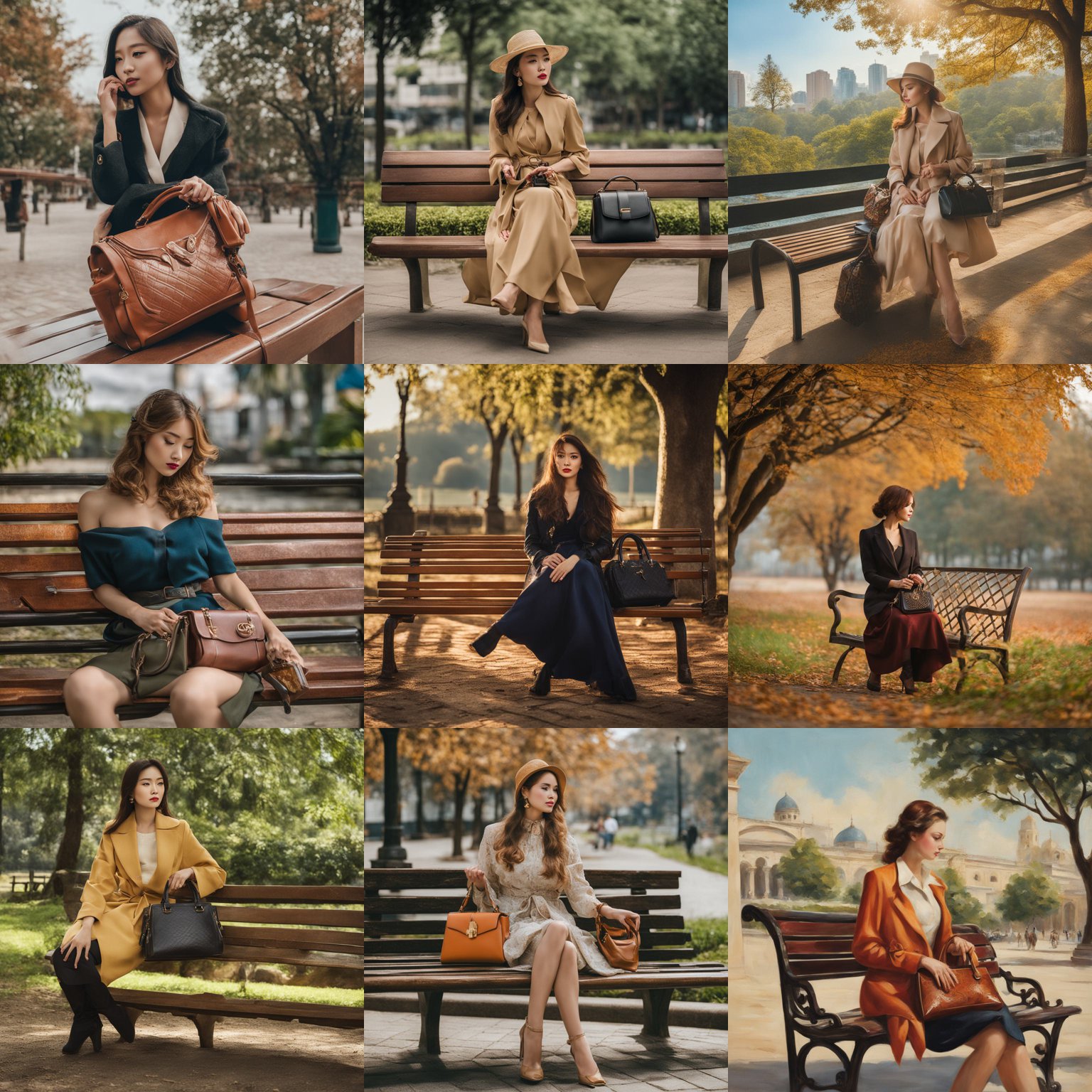}
        \captionsetup{labelformat=empty}
        \caption{``The lady is sitting on the bench holding her handbag"}
    \end{subfigure}
    \caption{Novel person-related biases identified on Stable Diffusion XL~\cite{podell2023sdxl} by OpenBias.}
    \label{fig:qualitatives_novel_person}
\end{figure*}

\begin{figure*}[htbp]
    \centering
    \begin{subfigure}{0.33\linewidth}
        \centering
        \textbf{Person gender}\par\medskip
        \vspace*{-0.12cm}
        \includegraphics[width=\linewidth]{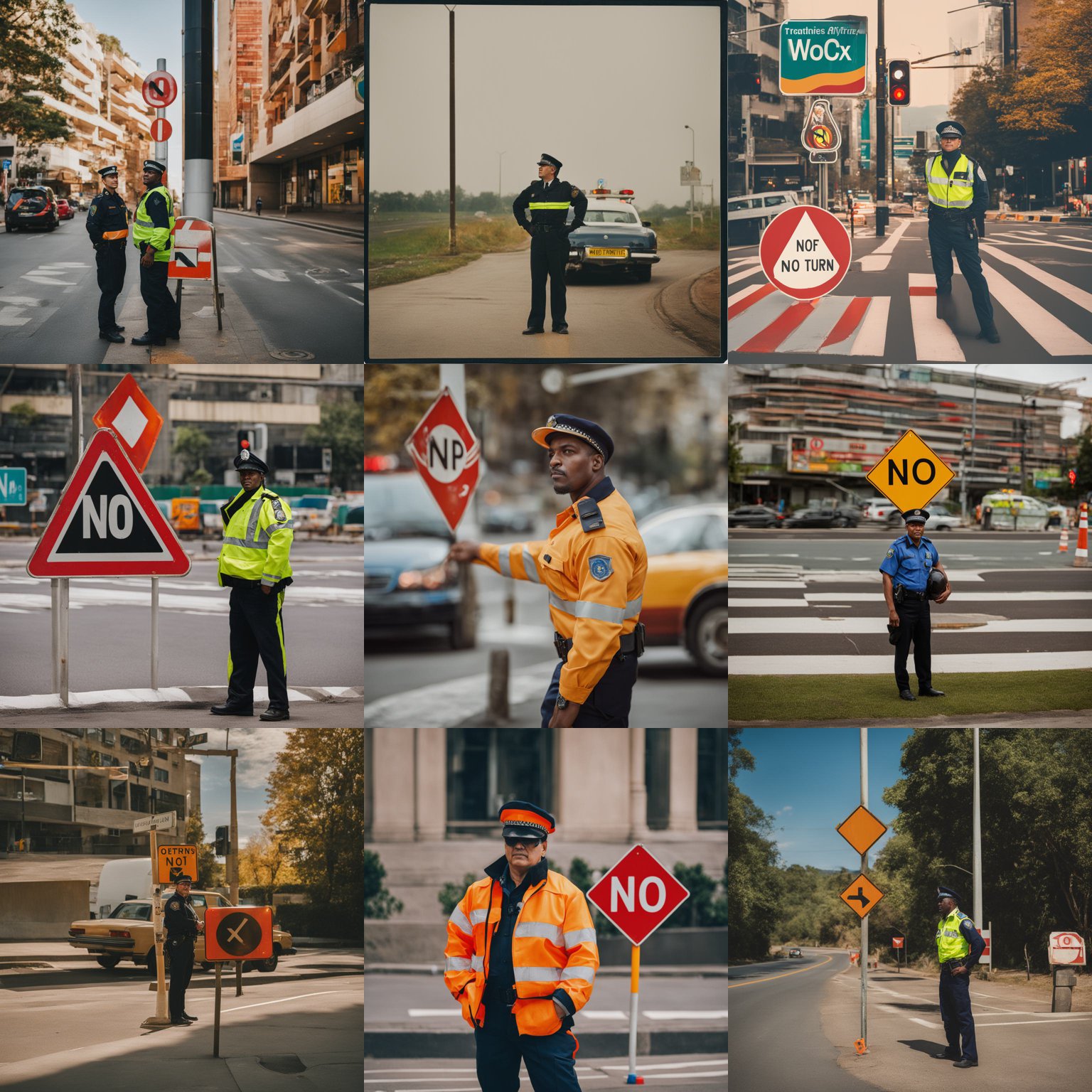}
        \captionsetup{labelformat=empty}
        \caption{``A traffic officer leaning on a no turn sign"}
    \end{subfigure}
    \hfill
    \begin{subfigure}{0.33\linewidth}
        \centering
        \textbf{Person race}\par\medskip
        \vspace*{-0.12cm}
        \includegraphics[width=\linewidth]{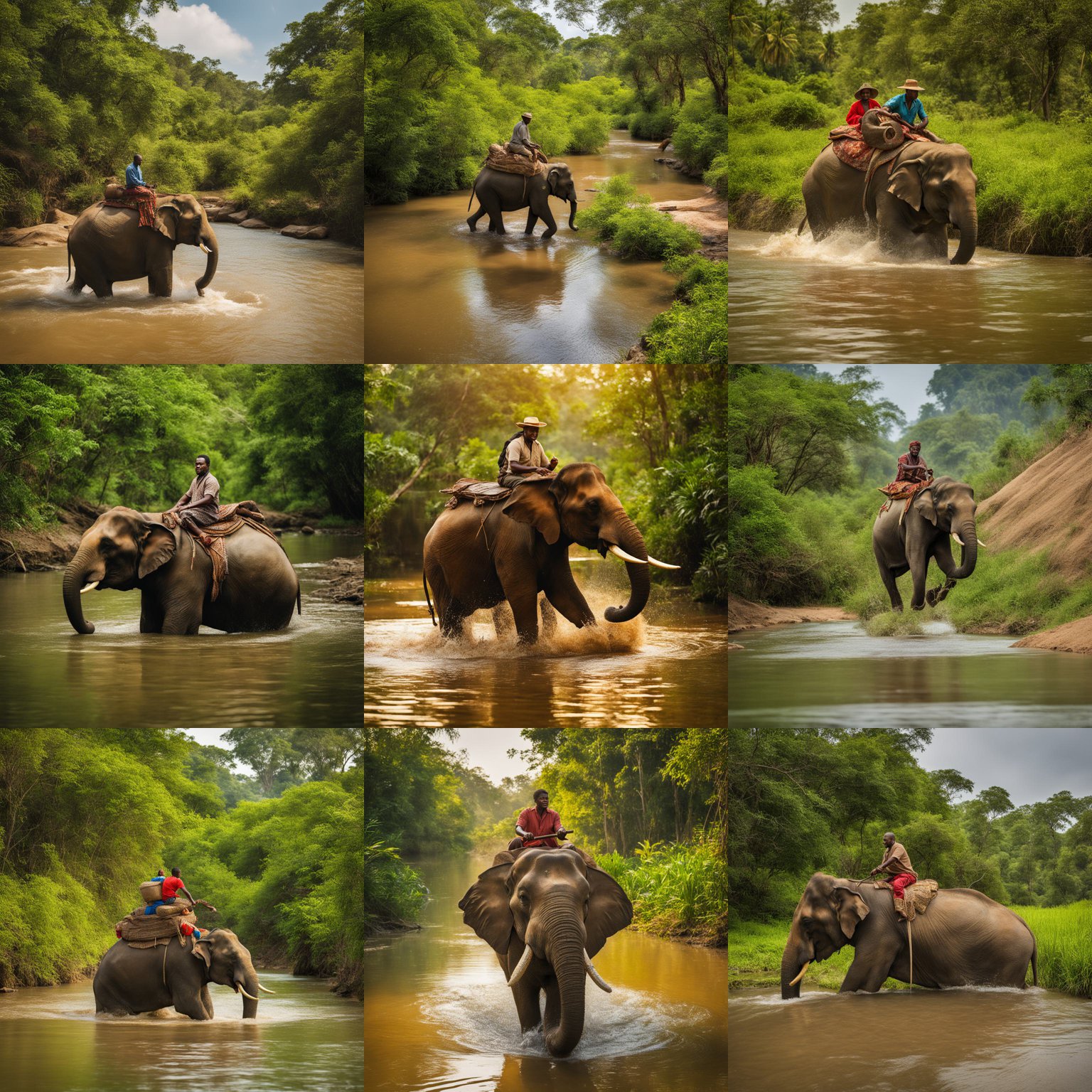}
        \captionsetup{labelformat=empty}
        \caption{``A man riding an elephant into some water of a creek"}
    \end{subfigure}
    \hfill
    \begin{subfigure}{0.33\linewidth}
        \centering
        \textbf{Person age}\par\medskip
        \vspace*{-0.12cm}
        \includegraphics[width=\linewidth]{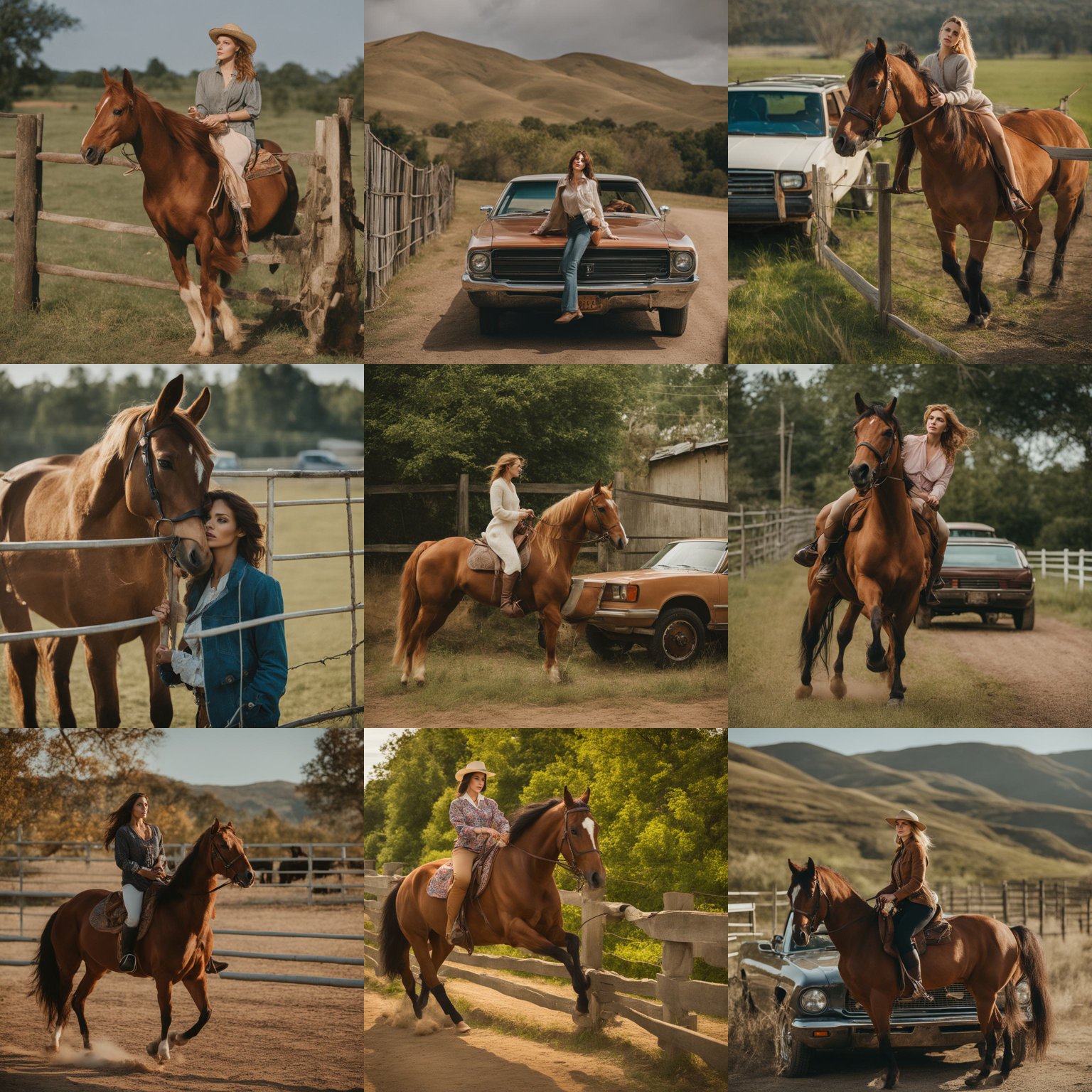}
        \captionsetup{labelformat=empty}
        \caption{``A woman riding a horse in front of a car next to a fence"}
    \end{subfigure}
    \caption{Person-related biases found on Stable Diffusion XL~\cite{podell2023sdxl} by OpenBias.}
    \label{fig:qualitatives_person}
\end{figure*}
\section{Findings}
\label{sec:comparison_ranking}

In this section, we present our findings from the examination of three extensively utilized text-to-image generative models, specifically Stable Diffusion XL, 2, and 1.5~\cite{podell2023sdxl, LDM_2022_CVPR}. We use captions from Flickr and COCO, as detailed in \cref{sec:implementation_details}. 
We structure our findings by examining the biases of different models 
and delineating the distinctions between context-free and context-aware bias.

\noindent \textbf{Rankings.} We present here the biases identified by our pipeline on Stable Diffusion XL, 2, and 1.5~\cite{podell2023sdxl, LDM_2022_CVPR}, in \cref{fig:comparison_coco} and \ref{fig:comparison_flickr}. Importantly,  OpenBias identifies both well-known (\eg \myquote{person gender}, \myquote{person race}) and novel biases (\eg \myquote{cake type}, \myquote{bed type} and \myquote{laptop brand}). From the comparison of different models, we observe a correlation between the intensities of the biases across different Stable Diffusion versions. We note, however, a subtle predominance of SD XL in the amplification of bias compared to earlier versions of the model. Moreover, the set of proposed biases varies depending on the initial set of captions used for the extraction. Generally, biases extracted from Flickr are more object-centric compared to those from COCO, aligning with the filtering operation applied to the latter. This difference highlights the potential of OpenBias to propose a tailored set of biases according to the captions it is applied to, making the bias proposals domain-specific.

\begin{figure}[t]
    \centering
\includegraphics[width=0.85\linewidth]{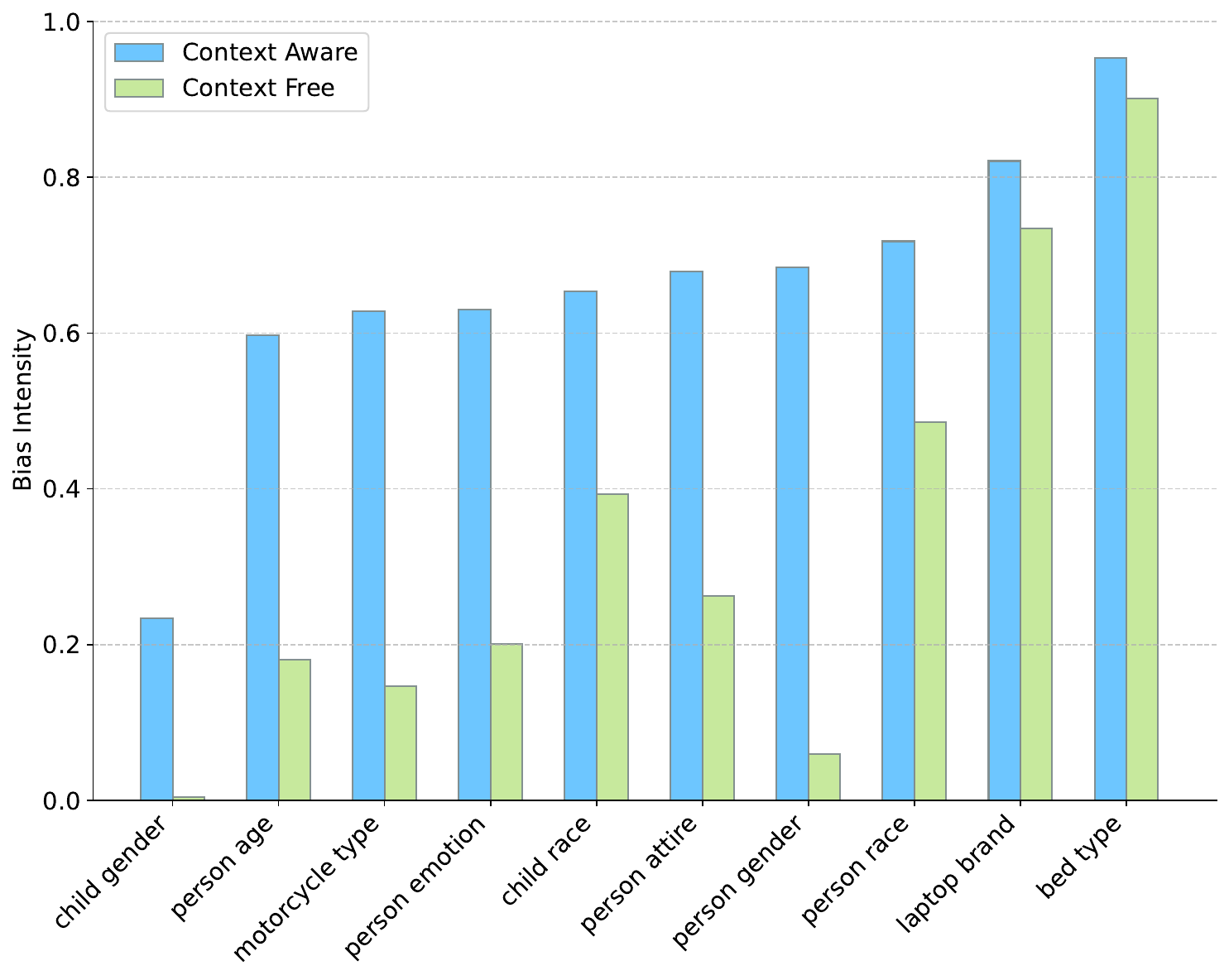}
    \vspace{-0.4cm}
    \caption{Highlighting the importance of the context aware approach on Stable Diffusion XL~\cite{podell2023sdxl} on the captions from COCO.}
    \label{fig:context_vs_free}
    \vspace{-0.5cm}
\end{figure}

\noindent \textbf{Context-Free vs Context-Aware.} Next, we study the different behavior of a given model, when compared in a context-free vs context-aware scenario (see \cref{sec:method} for formal definition). 
This analysis 
assesses the influence of other elements within the captions on the perpetuation of a particular bias. In \cref{fig:context_vs_free} we report the results obtained on SD XL. It is noteworthy to observe that, in this case, the correlation between the scores is not consistently present. For example, the intensity score for \myquote{motorcycle type} is significantly higher when computed within the context, compared to the same evaluation free of context. This discrepancy suggests that there is no majority class (\ie the general direction of the bias), but rather the model generates motorcycles of one specific type in a given context. Vice versa, for \myquote{bed type} we observe a high score in both settings, suggesting that the model 
always generates the same type of bed.

\para{Qualitative Results}
\label{sec:qualitative}
We show examples of biases discovered by OpenBias on Stable Diffusion XL. We present the results in a context-aware fashion and visualize images generated from the same caption where our pipeline identifies a bias. We organize the results in three sets and present unexplored biases on objects and animals, novel biases associated with persons, and well-known social biases.
We highlight biases discovered on objects and animals in \cref{fig:qualitatives_others}. For example, 
the model tends to generate \myquote{yellow} trains or \myquote{quarter horses} even if not specified in the caption. Furthermore, the model generates laptops featuring a distinct \myquote{Apple} logo, showing a bias toward the brand.

Next, we display novel biases related to persons discovered by OpenBias. For instance, we unveil unexplored biases such as the \myquote{person attire}, with the model often generating people in a formal outfit rather than more casual ones. 
Furthermore, we specifically study \myquote{child gender} and \myquote{child race} diverging from the typical examination centered on adults. For example, in \cref{fig:qualitatives_novel_person} second column, we observe that the generative model links a black child with an economically disadvantaged environment described in the caption as \myquote{a dirt road}. 
The association between racial identity and socioeconomic status perpetuates harmful stereotypes and proves the need to consider novel biases within bias mitigation frameworks.
Lastly, we show qualitative results on the well-studied and sensitive biases of \myquote{person gender}, \myquote{race}, and \myquote{age}. 
In the first column of \cref{fig:qualitatives_person}, Stable Diffusion XL exclusively generates  \myquote{male} officers, despite the presence of a gender-neutral job title.  Moreover, it explicitly depicts a \myquote{woman} labeled as \myquote{middle-aged} when engaged in horseback riding. Finally, we observe a \myquote{race} bias, with depictions of solely black individuals for 
\myquote{a man riding an elephant}. This context-aware approach ensures a thorough comprehension of emerging biases in both novel and socially significant contexts. These results 
emphasize the necessity for more inclusive open-set bias detection frameworks. We provide additional qualitatives and comparisons in the {\supp}.


\section{Limitations}
\vspace{-2pt}
\label{sec:limitations}
OpenBias is based on two foundation models to propose and quantify biases of a generative model, namely LLama~\cite{touvron2023llama} and LLava \cite{liu2023llava}. We rely on the prediction of these models, without considering their intrinsic limitations. 
Existing research \cite{gallegos2023bias, 10.1145/3597307} highlights the presence of biases in these models which may be propagated in our pipeline.  
Nevertheless, the modular nature of our pipeline provides flexibility, allowing us to seamlessly incorporate improved models should they become available in the future.
Finally, in this work, we delve into the distinction between context-free and context-aware biases, revealing different behaviors exhibited by models in these two scenarios. However, our evaluation of the role of the context is only qualitative. We identify the possibility of systematically studying the context's role as a promising future direction. 
\section{Conclusions}\vspace{-2pt}
AI-generated content has seen rapid growth in the last few years, with the potential to become even more ubiquitous in society.  While the usage of such models increases, characterizing the stereotypes perpetrated by the model becomes of significant importance. In this work, we propose to study the bias in generative models in a novel open-set scenario, paving the way to the discovery of biases previously unexplored. We propose OpenBias, an automatic bias detection pipeline, capable of discovering and quantifying traditional and novel biases without the need to pre-define them. The proposed method builds a domain-specific knowledge base of biases which are then assessed and quantified via Vision Question Answering. We validate OpenBias showing its agreement with classifier-based methods on a closed set of concepts and with human judgement through a user study. Our method can be plugged into existing bias mitigation works, 
extending 
their 
capabilities to novel biases. 
OpenBias can foster further research in open-set scenarios, moving beyond classical pre-defined biases and assessing generative models more comprehensively.

\noindent \textbf{Ethical statement and broader impact.} This work contributes to fairer and more inclusive AI, 
by detecting biases in T2I generative models. 
We conduct our research responsibly, transparently, and with a strong commitment to ethical principles. Despite this, due to technical constraints, socially sensitive attributes, such as gender, are treated as closed sets for research purposes only. Moreover, OpenBias entails the biases of the LLM and VQA models, thus it may not discover all possible biases. \textit{We do not intend to discriminate against any social group but raise awareness on the challenges of detecting biases beyond closed sets.}

{\setlength{\parindent}{0cm}
\textbf{Acknowledgments:} This work was supported by the MUR PNRR project FAIR (PE00000013) funded by the NextGenerationEU and by the EU Horizon projects ELIAS (No. 101120237) and AI4Media (No. 951911),  NSF CAREER Award \#2239840, and the National AI Institute for Exceptional Education (Award \#2229873) by National Science Foundation and the Institute of Education Sciences, U.S. Department of Education, and Picsart AI Research (PAIR).
}

{
    \small
    \bibliographystyle{ieeenat_fullname}
    \bibliography{main}

\begin{thebibliography}{76}
\providecommand{\natexlab}[1]{#1}
\providecommand{\url}[1]{\texttt{#1}}
\expandafter\ifx\csname urlstyle\endcsname\relax
  \providecommand{\doi}[1]{doi: #1}\else
  \providecommand{\doi}{doi: \begingroup \urlstyle{rm}\Url}\fi

\bibitem[Agarwal et~al.(2022)Agarwal, Muku, Anand, and Arora]{Agarwal_2022_WACV}
Sharat Agarwal, Sumanyu Muku, Saket Anand, and Chetan Arora.
\newblock Does data repair lead to fair models? curating contextually fair data to reduce model bias.
\newblock In \emph{WACV}, 2022.

\bibitem[Avrahami et~al.(2023)Avrahami, Hayes, Gafni, Gupta, Taigman, Parikh, Lischinski, Fried, and Yin]{avrahami2023spatext}
Omri Avrahami, Thomas Hayes, Oran Gafni, Sonal Gupta, Yaniv Taigman, Devi Parikh, Dani Lischinski, Ohad Fried, and Xi Yin.
\newblock Spatext: Spatio-textual representation for controllable image generation.
\newblock In \emph{CVPR}, 2023.

\bibitem[Bianchi et~al.(2023)Bianchi, Kalluri, Durmus, Ladhak, Cheng, Nozza, Hashimoto, Jurafsky, Zou, and Caliskan]{10.1145/3593013.3594095}
Federico Bianchi, Pratyusha Kalluri, Esin Durmus, Faisal Ladhak, Myra Cheng, Debora Nozza, Tatsunori Hashimoto, Dan Jurafsky, James Zou, and Aylin Caliskan.
\newblock Easily accessible text-to-image generation amplifies demographic stereotypes at large scale.
\newblock In \emph{Proceedings of the 2023 ACM Conference on Fairness, Accountability, and Transparency}, 2023.

\bibitem[Bolukbasi et~al.(2016)Bolukbasi, Chang, Zou, Saligrama, and Kalai]{NIPS2016_bias_word_embeddings}
Tolga Bolukbasi, Kai-Wei Chang, James~Y Zou, Venkatesh Saligrama, and Adam~T Kalai.
\newblock Man is to computer programmer as woman is to homemaker? debiasing word embeddings.
\newblock In \emph{NeurIPS}, 2016.

\bibitem[Bommasani et~al.(2021)Bommasani, Hudson, Adeli, Altman, Arora, von Arx, Bernstein, Bohg, Bosselut, Brunskill, et~al.]{bommasani2021opportunities}
Rishi Bommasani, Drew~A Hudson, Ehsan Adeli, Russ Altman, Simran Arora, Sydney von Arx, Michael~S Bernstein, Jeannette Bohg, Antoine Bosselut, Emma Brunskill, et~al.
\newblock On the opportunities and risks of foundation models.
\newblock \emph{arXiv preprint}, 2021.

\bibitem[Brack et~al.(2023{\natexlab{a}})Brack, Friedrich, Hintersdorf, Struppek, Schramowski, and Kersting]{NEURIPS2023_4ff83037}
Manuel Brack, Felix Friedrich, Dominik Hintersdorf, Lukas Struppek, Patrick Schramowski, and Kristian Kersting.
\newblock Sega: Instructing text-to-image models using semantic guidance.
\newblock In \emph{NeurIPS}, 2023{\natexlab{a}}.

\bibitem[Brack et~al.(2023{\natexlab{b}})Brack, Friedrich, Schramowski, and Kersting]{brack2023mitigating}
Manuel Brack, Felix Friedrich, Patrick Schramowski, and Kristian Kersting.
\newblock Mitigating inappropriateness in image generation: Can there be value in reflecting the world's ugliness?
\newblock \emph{arXiv preprint}, 2023{\natexlab{b}}.

\bibitem[Brooks et~al.(2023)Brooks, Holynski, and Efros]{brooks2023instructpix2pix}
Tim Brooks, Aleksander Holynski, and Alexei~A Efros.
\newblock Instructpix2pix: Learning to follow image editing instructions.
\newblock In \emph{CVPR}, 2023.

\bibitem[Brown et~al.(2020)Brown, Mann, Ryder, Subbiah, Kaplan, Dhariwal, Neelakantan, Shyam, Sastry, Askell, Agarwal, Herbert-Voss, Krueger, Henighan, Child, Ramesh, Ziegler, Wu, Winter, Hesse, Chen, Sigler, Litwin, Gray, Chess, Clark, Berner, McCandlish, Radford, Sutskever, and Amodei]{NEURIPS2020_gpt_3_few_shot}
Tom Brown, Benjamin Mann, Nick Ryder, Melanie Subbiah, Jared~D Kaplan, Prafulla Dhariwal, Arvind Neelakantan, Pranav Shyam, Girish Sastry, Amanda Askell, Sandhini Agarwal, Ariel Herbert-Voss, Gretchen Krueger, Tom Henighan, Rewon Child, Aditya Ramesh, Daniel Ziegler, Jeffrey Wu, Clemens Winter, Chris Hesse, Mark Chen, Eric Sigler, Mateusz Litwin, Scott Gray, Benjamin Chess, Jack Clark, Christopher Berner, Sam McCandlish, Alec Radford, Ilya Sutskever, and Dario Amodei.
\newblock Language models are few-shot learners.
\newblock In \emph{NeurIPS}, 2020.

\bibitem[Caron et~al.(2021)Caron, Touvron, Misra, J{\'e}gou, Mairal, Bojanowski, and Joulin]{caron2021emerging}
Mathilde Caron, Hugo Touvron, Ishan Misra, Herv{\'e} J{\'e}gou, Julien Mairal, Piotr Bojanowski, and Armand Joulin.
\newblock Emerging properties in self-supervised vision transformers.
\newblock In \emph{ICCV}, 2021.

\bibitem[Chen et~al.(2023)Chen, Zhu, Haydarov, Li, and Elhoseiny]{chen2023video}
Jun Chen, Deyao Zhu, Kilichbek Haydarov, Xiang Li, and Mohamed Elhoseiny.
\newblock Video chatcaptioner: Towards the enriched spatiotemporal descriptions.
\newblock \emph{arXiv preprint}, 2023.

\bibitem[Cherti et~al.(2023)Cherti, Beaumont, Wightman, Wortsman, Ilharco, Gordon, Schuhmann, Schmidt, and Jitsev]{Cherti_2023_CVPR}
Mehdi Cherti, Romain Beaumont, Ross Wightman, Mitchell Wortsman, Gabriel Ilharco, Cade Gordon, Christoph Schuhmann, Ludwig Schmidt, and Jenia Jitsev.
\newblock Reproducible scaling laws for contrastive language-image learning.
\newblock In \emph{CVPR}, 2023.

\bibitem[Cho et~al.(2023)Cho, Zala, and Bansal]{DALL_EVAL_2023_ICCV}
Jaemin Cho, Abhay Zala, and Mohit Bansal.
\newblock Dall-eval: Probing the reasoning skills and social biases of text-to-image generation models.
\newblock In \emph{ICCV}, 2023.

\bibitem[D'Inc\`a et~al.(2024)D'Inc\`a, Tzelepis, Patras, and Sebe]{D'Inca_2024_WACV}
Moreno D'Inc\`a, Christos Tzelepis, Ioannis Patras, and Nicu Sebe.
\newblock Improving fairness using vision-language driven image augmentation.
\newblock In \emph{WACV}, 2024.

\bibitem[Dosovitskiy et~al.(2021)Dosovitskiy, Beyer, Kolesnikov, Weissenborn, Zhai, Unterthiner, Dehghani, Minderer, Heigold, Gelly, et~al.]{dosovitskiy2020image}
Alexey Dosovitskiy, Lucas Beyer, Alexander Kolesnikov, Dirk Weissenborn, Xiaohua Zhai, Thomas Unterthiner, Mostafa Dehghani, Matthias Minderer, Georg Heigold, Sylvain Gelly, et~al.
\newblock An image is worth 16x16 words: Transformers for image recognition at scale.
\newblock In \emph{ICLR}, 2021.

\bibitem[Epstein et~al.(2023)Epstein, Jabri, Poole, Efros, and Holynski]{epstein2023diffusion}
Dave Epstein, Allan Jabri, Ben Poole, Alexei~A Efros, and Aleksander Holynski.
\newblock Diffusion self-guidance for controllable image generation.
\newblock In \emph{NeurIPS}, 2023.

\bibitem[Friedrich et~al.(2023)Friedrich, Brack, Struppek, Hintersdorf, Schramowski, Luccioni, and Kersting]{fairDiffusion2023}
Felix Friedrich, Manuel Brack, Lukas Struppek, Dominik Hintersdorf, Patrick Schramowski, Sasha Luccioni, and Kristian Kersting.
\newblock Fair diffusion: Instructing text-to-image generation models on fairness.
\newblock \emph{arXiv preprint}, 2023.

\bibitem[Gal et~al.(2022)Gal, Alaluf, Atzmon, Patashnik, Bermano, Chechik, and Cohen-Or]{gal2022textual}
Rinon Gal, Yuval Alaluf, Yuval Atzmon, Or Patashnik, Amit~H. Bermano, Gal Chechik, and Daniel Cohen-Or.
\newblock An image is worth one word: Personalizing text-to-image generation using textual inversion.
\newblock \emph{arXiv preprint}, 2022.

\bibitem[Gallegos et~al.(2023)Gallegos, Rossi, Barrow, Tanjim, Kim, Dernoncourt, Yu, Zhang, and Ahmed]{gallegos2023bias}
Isabel~O Gallegos, Ryan~A Rossi, Joe Barrow, Md~Mehrab Tanjim, Sungchul Kim, Franck Dernoncourt, Tong Yu, Ruiyi Zhang, and Nesreen~K Ahmed.
\newblock Bias and fairness in large language models: A survey.
\newblock \emph{arXiv preprint}, 2023.

\bibitem[Gandikota et~al.(2024)Gandikota, Orgad, Belinkov, Materzy\'nska, and Bau]{Gandikota_2024_WACV}
Rohit Gandikota, Hadas Orgad, Yonatan Belinkov, Joanna Materzy\'nska, and David Bau.
\newblock Unified concept editing in diffusion models.
\newblock In \emph{WACV}, 2024.

\bibitem[Goel et~al.(2023)Goel, Peruzzo, Jiang, Xu, Xu, Sebe, Darrell, Wang, and Shi]{goel2023pairdiffusion}
Vidit Goel, Elia Peruzzo, Yifan Jiang, Dejia Xu, Xingqian Xu, Nicu Sebe, Trevor Darrell, Zhangyang Wang, and Humphrey Shi.
\newblock Pair-diffusion: A comprehensive multimodal object-level image editor.
\newblock \emph{arXiv preprint}, 2023.

\bibitem[Gupta and Kembhavi(2023)]{gupta2023visual}
Tanmay Gupta and Aniruddha Kembhavi.
\newblock Visual programming: Compositional visual reasoning without training.
\newblock In \emph{CVPR}, 2023.

\bibitem[Hendricks et~al.(2018)Hendricks, Burns, Saenko, Darrell, and Rohrbach]{Bias_Women_also_Snowboard_2018_ECCV}
Lisa~Anne Hendricks, Kaylee Burns, Kate Saenko, Trevor Darrell, and Anna Rohrbach.
\newblock Women also snowboard: Overcoming bias in captioning models.
\newblock In \emph{ECCV}, 2018.

\bibitem[Hertz et~al.(2022)Hertz, Mokady, Tenenbaum, Aberman, Pritch, and Cohen-Or]{hertz2022prompt}
Amir Hertz, Ron Mokady, Jay Tenenbaum, Kfir Aberman, Yael Pritch, and Daniel Cohen-Or.
\newblock Prompt-to-prompt image editing with cross attention control.
\newblock In \emph{ICLR}, 2022.

\bibitem[Ho and Salimans(2021)]{ho2022classifier}
Jonathan Ho and Tim Salimans.
\newblock Classifier-free diffusion guidance.
\newblock In \emph{NeurIPS 2021 Workshop on Deep Generative Models and Downstream Applications}, 2021.

\bibitem[Hu et~al.(2023{\natexlab{a}})Hu, Hua, Yang, Shi, Smith, and Luo]{Hu_2023_ICCV}
Yushi Hu, Hang Hua, Zhengyuan Yang, Weijia Shi, Noah~A. Smith, and Jiebo Luo.
\newblock Promptcap: Prompt-guided image captioning for vqa with gpt-3.
\newblock In \emph{ICCV}, 2023{\natexlab{a}}.

\bibitem[Hu et~al.(2023{\natexlab{b}})Hu, Liu, Kasai, Wang, Ostendorf, Krishna, and Smith]{TIFA_2023_ICCV}
Yushi Hu, Benlin Liu, Jungo Kasai, Yizhong Wang, Mari Ostendorf, Ranjay Krishna, and Noah~A. Smith.
\newblock Tifa: Accurate and interpretable text-to-image faithfulness evaluation with question answering.
\newblock In \emph{ICCV}, 2023{\natexlab{b}}.

\bibitem[Huang et~al.(2023)Huang, Chen, Liu, Shen, Zhao, and Zhou]{huang2023composer}
Lianghua Huang, Di Chen, Yu Liu, Yujun Shen, Deli Zhao, and Jingren Zhou.
\newblock Composer: Creative and controllable image synthesis with composable conditions.
\newblock In \emph{ICML}, 2023.

\bibitem[Jung et~al.(2022)Jung, Chun, and Moon]{jung2022learning}
Sangwon Jung, Sanghyuk Chun, and Taesup Moon.
\newblock Learning fair classifiers with partially annotated group labels.
\newblock In \emph{CVPR}, 2022.

\bibitem[Karkkainen and Joo(2021)]{Karkkainen_2021_WACV}
Kimmo Karkkainen and Jungseock Joo.
\newblock Fairface: Face attribute dataset for balanced race, gender, and age for bias measurement and mitigation.
\newblock In \emph{WACV}, 2021.

\bibitem[Karras et~al.(2019)Karras, Laine, and Aila]{Karras_2019_CVPR}
Tero Karras, Samuli Laine, and Timo Aila.
\newblock A style-based generator architecture for generative adversarial networks.
\newblock In \emph{CVPR}, 2019.

\bibitem[Karras et~al.(2021)Karras, Aittala, Laine, H\"{a}rk\"{o}nen, Hellsten, Lehtinen, and Aila]{NEURIPS2021_076ccd93}
Tero Karras, Miika Aittala, Samuli Laine, Erik H\"{a}rk\"{o}nen, Janne Hellsten, Jaakko Lehtinen, and Timo Aila.
\newblock Alias-free generative adversarial networks.
\newblock In \emph{NeurIPS}, 2021.

\bibitem[Kenfack et~al.(2022)Kenfack, Sabbagh, Rivera, and Khan]{kenfack2022repfairgan}
Patrik~Joslin Kenfack, Kamil Sabbagh, Adín~Ramírez Rivera, and Adil Khan.
\newblock Repfair-gan: Mitigating representation bias in gans using gradient clipping.
\newblock \emph{arXiv preprint}, 2022.

\bibitem[Kim et~al.(2021)Kim, Son, and Kim]{kim2021vilt}
Wonjae Kim, Bokyung Son, and Ildoo Kim.
\newblock Vilt: Vision-and-language transformer without convolution or region supervision.
\newblock In \emph{ICML}, 2021.

\bibitem[Kim et~al.(2023)Kim, Mo, Kim, Lee, Lee, and Shin]{kim2023biastotext}
Younghyun Kim, Sangwoo Mo, Minkyu Kim, Kyungmin Lee, Jaeho Lee, and Jinwoo Shin.
\newblock Bias-to-text: Debiasing unknown visual biases through language interpretation.
\newblock \emph{arXiv preprint}, 2023.

\bibitem[Lakshmi et~al.(2021)Lakshmi, Wittenbrink, Correll, and Ma]{Lakshmi2021}
Anjana Lakshmi, Bernd Wittenbrink, Joshua Correll, and Debbie~S. Ma.
\newblock The india face set: International and cultural boundaries impact face impressions and perceptions of category membership.
\newblock \emph{Frontiers in Psychology}, 2021.

\bibitem[Li et~al.(2022{\natexlab{a}})Li, Xu, Tian, Wang, Yan, Bi, Ye, Chen, Xu, Cao, et~al.]{li2022mplug}
Chenliang Li, Haiyang Xu, Junfeng Tian, Wei Wang, Ming Yan, Bin Bi, Jiabo Ye, Hehong Chen, Guohai Xu, Zheng Cao, et~al.
\newblock m{PLUG}: Effective and efficient vision-language learning by cross-modal skip-connections.
\newblock In \emph{Proceedings of the 2022 Conference on Empirical Methods in Natural Language Processing}, 2022{\natexlab{a}}.

\bibitem[Li et~al.(2022{\natexlab{b}})Li, Li, Xiong, and Hoi]{li2022blip}
Junnan Li, Dongxu Li, Caiming Xiong, and Steven Hoi.
\newblock Blip: Bootstrapping language-image pre-training for unified vision-language understanding and generation.
\newblock In \emph{ICML}, 2022{\natexlab{b}}.

\bibitem[Li et~al.(2023)Li, Li, Savarese, and Hoi]{li2023blip2}
Junnan Li, Dongxu Li, Silvio Savarese, and Steven Hoi.
\newblock {BLIP}-2: Bootstrapping language-image pre-training with frozen image encoders and large language models.
\newblock In \emph{Proceedings of the 40th International Conference on Machine Learning}, 2023.

\bibitem[Lin et~al.(2014)Lin, Maire, Belongie, Bourdev, Girshick, Hays, Perona, Ramanan, Doll{\'{a}}r, and Zitnick]{DBLP:journals/corr/LinMBHPRDZ14}
Tsung{-}Yi Lin, Michael Maire, Serge~J. Belongie, Lubomir~D. Bourdev, Ross~B. Girshick, James Hays, Pietro Perona, Deva Ramanan, Piotr Doll{\'{a}}r, and C.~Lawrence Zitnick.
\newblock Microsoft {COCO:} common objects in context.
\newblock In \emph{ECCV}, 2014.

\bibitem[Liu et~al.(2023{\natexlab{a}})Liu, Li, Li, and Lee]{liu2023improvedllava}
Haotian Liu, Chunyuan Li, Yuheng Li, and Yong~Jae Lee.
\newblock Improved baselines with visual instruction tuning.
\newblock In \emph{NeurIPS 2023 Workshop on Instruction Tuning and Instruction Following}, 2023{\natexlab{a}}.

\bibitem[Liu et~al.(2023{\natexlab{b}})Liu, Li, Wu, and Lee]{liu2023llava}
Haotian Liu, Chunyuan Li, Qingyang Wu, and Yong~Jae Lee.
\newblock Visual instruction tuning.
\newblock In \emph{NeurIPS}, 2023{\natexlab{b}}.

\bibitem[Ma et~al.(2015)Ma, Correll, and Wittenbrink]{Ma2015}
Debbie~S. Ma, Joshua Correll, and Bernd Wittenbrink.
\newblock The chicago face database: A free stimulus set of faces and norming data.
\newblock 2015.

\bibitem[Ma et~al.(2020)Ma, Kantner, and Wittenbrink]{Ma2020}
Debbie~S. Ma, Justin Kantner, and Bernd Wittenbrink.
\newblock Chicago face database: Multiracial expansion.
\newblock \emph{Behavior Research Methods}, 2020.

\bibitem[Naik and Nushi(2023)]{naik2023social}
Ranjita Naik and Besmira Nushi.
\newblock Social biases through the text-to-image generation lens.
\newblock In \emph{Proceedings of the 2023 AAAI/ACM Conference on AI, Ethics, and Society}, 2023.

\bibitem[Nam et~al.(2020)Nam, Cha, Ahn, Lee, and Shin]{nam2020learning}
Junhyun Nam, Hyuntak Cha, Sungsoo Ahn, Jaeho Lee, and Jinwoo Shin.
\newblock Learning from failure: De-biasing classifier from biased classifier.
\newblock \emph{NeurIPS}, 2020.

\bibitem[Navigli et~al.(2023)Navigli, Conia, and Ross]{10.1145/3597307}
Roberto Navigli, Simone Conia, and Bj\"{o}rn Ross.
\newblock Biases in large language models: Origins, inventory, and discussion.
\newblock \emph{ACM Journal of Data and Information Quality}, 2023.

\bibitem[Nichol et~al.(2022)Nichol, Dhariwal, Ramesh, Shyam, Mishkin, McGrew, Sutskever, and Chen]{nichol2021glide}
Alex Nichol, Prafulla Dhariwal, Aditya Ramesh, Pranav Shyam, Pamela Mishkin, Bob McGrew, Ilya Sutskever, and Mark Chen.
\newblock Glide: Towards photorealistic image generation and editing with text-guided diffusion models.
\newblock In \emph{International Conference on Machine Learning}, 2022.

\bibitem[Oquab et~al.(2023)Oquab, Darcet, Moutakanni, Vo, Szafraniec, Khalidov, Fernandez, Haziza, Massa, El-Nouby, et~al.]{oquab2023dinov2}
Maxime Oquab, Timoth{\'e}e Darcet, Th{\'e}o Moutakanni, Huy Vo, Marc Szafraniec, Vasil Khalidov, Pierre Fernandez, Daniel Haziza, Francisco Massa, Alaaeldin El-Nouby, et~al.
\newblock Dinov2: Learning robust visual features without supervision.
\newblock In \emph{Transactions on Machine Learning Research}, 2023.

\bibitem[Podell et~al.(2024)Podell, English, Lacey, Blattmann, Dockhorn, Müller, Penna, and Rombach]{podell2023sdxl}
Dustin Podell, Zion English, Kyle Lacey, Andreas Blattmann, Tim Dockhorn, Jonas Müller, Joe Penna, and Robin Rombach.
\newblock {SDXL}: Improving latent diffusion models for high-resolution image synthesis.
\newblock In \emph{ICLR}, 2024.

\bibitem[Radford et~al.(2021)Radford, Kim, Hallacy, Ramesh, Goh, Agarwal, Sastry, Askell, Mishkin, Clark, et~al.]{radford2021learning}
Alec Radford, Jong~Wook Kim, Chris Hallacy, Aditya Ramesh, Gabriel Goh, Sandhini Agarwal, Girish Sastry, Amanda Askell, Pamela Mishkin, Jack Clark, et~al.
\newblock Learning transferable visual models from natural language supervision.
\newblock In \emph{ICML}, 2021.

\bibitem[Ramesh et~al.(2022)Ramesh, Dhariwal, Nichol, Chu, and Chen]{hierarchical_text_to_image_CLIP}
Aditya Ramesh, Prafulla Dhariwal, Alex Nichol, Casey Chu, and Mark Chen.
\newblock Hierarchical text-conditional image generation with clip latents.
\newblock \emph{arXiv preprint}, 2022.

\bibitem[Rombach et~al.(2022)Rombach, Blattmann, Lorenz, Esser, and Ommer]{LDM_2022_CVPR}
Robin Rombach, Andreas Blattmann, Dominik Lorenz, Patrick Esser, and Bj\"orn Ommer.
\newblock High-resolution image synthesis with latent diffusion models.
\newblock In \emph{CVPR}, 2022.

\bibitem[Ruiz et~al.(2023)Ruiz, Li, Jampani, Pritch, Rubinstein, and Aberman]{Ruiz_2023_CVPR}
Nataniel Ruiz, Yuanzhen Li, Varun Jampani, Yael Pritch, Michael Rubinstein, and Kfir Aberman.
\newblock Dreambooth: Fine tuning text-to-image diffusion models for subject-driven generation.
\newblock In \emph{CVPR}, 2023.

\bibitem[Saharia et~al.(2022)Saharia, Chan, Saxena, Li, Whang, Denton, Ghasemipour, Gontijo~Lopes, Karagol~Ayan, Salimans, et~al.]{saharia2022photorealistic}
Chitwan Saharia, William Chan, Saurabh Saxena, Lala Li, Jay Whang, Emily~L Denton, Kamyar Ghasemipour, Raphael Gontijo~Lopes, Burcu Karagol~Ayan, Tim Salimans, et~al.
\newblock Photorealistic text-to-image diffusion models with deep language understanding.
\newblock \emph{NeurIPS}, 2022.

\bibitem[Savani et~al.(2020)Savani, White, and Govindarajulu]{savani2020intra}
Yash Savani, Colin White, and Naveen~Sundar Govindarajulu.
\newblock Intra-processing methods for debiasing neural networks.
\newblock In \emph{NeurIPS}, 2020.

\bibitem[Speer et~al.(2017)Speer, Chin, and Havasi]{ConceptNET}
Robyn Speer, Joshua Chin, and Catherine Havasi.
\newblock Conceptnet 5.5: An open multilingual graph of general knowledge.
\newblock In \emph{AAAI}, 2017.

\bibitem[SU et~al.(2023)SU, Kasai, Wu, Shi, Wang, Xin, Zhang, Ostendorf, Zettlemoyer, Smith, and Yu]{ICLR_2023_selective_annotations}
Hongjin SU, Jungo Kasai, Chen~Henry Wu, Weijia Shi, Tianlu Wang, Jiayi Xin, Rui Zhang, Mari Ostendorf, Luke Zettlemoyer, Noah~A. Smith, and Tao Yu.
\newblock Selective annotation makes language models better few-shot learners.
\newblock In \emph{ICLR}, 2023.

\bibitem[Su et~al.(2023)Su, Ren, Qiang, Song, Gao, Wu, and Zheng]{su2023unbiased}
Xingzhe Su, Yi Ren, Wenwen Qiang, Zeen Song, Hang Gao, Fengge Wu, and Changwen Zheng.
\newblock Unbiased image synthesis via manifold-driven sampling in diffusion models.
\newblock \emph{arXiv preprint}, 2023.

\bibitem[Subramanian et~al.(2022)Subramanian, Merrill, Darrell, Gardner, Singh, and Rohrbach]{subramanian2022reclip}
Sanjay Subramanian, William Merrill, Trevor Darrell, Matt Gardner, Sameer Singh, and Anna Rohrbach.
\newblock Reclip: A strong zero-shot baseline for referring expression comprehension.
\newblock In \emph{Proceedings of the 60th Annual Meeting of the Association for Computational Linguistics}, 2022.

\bibitem[Sur{\'\i}s et~al.(2023)Sur{\'\i}s, Menon, and Vondrick]{suris2023vipergpt}
D{\'\i}dac Sur{\'\i}s, Sachit Menon, and Carl Vondrick.
\newblock Vipergpt: Visual inference via python execution for reasoning.
\newblock In \emph{ICCV}, 2023.

\bibitem[Tan et~al.(2021)Tan, Shen, and Zhou]{tan2021improving}
Shuhan Tan, Yujun Shen, and Bolei Zhou.
\newblock Improving the fairness of deep generative models without retraining.
\newblock \emph{arXiv preprint}, 2021.

\bibitem[Touvron et~al.(2023)Touvron, Lavril, Izacard, Martinet, Lachaux, Lacroix, Rozi{\`e}re, Goyal, Hambro, Azhar, et~al.]{touvron2023llama}
Hugo Touvron, Thibaut Lavril, Gautier Izacard, Xavier Martinet, Marie-Anne Lachaux, Timoth{\'e}e Lacroix, Baptiste Rozi{\`e}re, Naman Goyal, Eric Hambro, Faisal Azhar, et~al.
\newblock Llama: Open and efficient foundation language models.
\newblock \emph{arXiv preprint}, 2023.

\bibitem[Verma and Rubin(2018)]{verma2018fairness}
Sahil Verma and Julia Rubin.
\newblock Fairness definitions explained.
\newblock In \emph{Proceedings of the international workshop on software fairness}, 2018.

\bibitem[Wang et~al.(2022{\natexlab{a}})Wang, Yang, Hu, Li, Lin, Gan, Liu, Liu, and Wang]{wang2022git}
Jianfeng Wang, Zhengyuan Yang, Xiaowei Hu, Linjie Li, Kevin Lin, Zhe Gan, Zicheng Liu, Ce Liu, and Lijuan Wang.
\newblock Git: A generative image-to-text transformer for vision and language.
\newblock In \emph{Transactions on Machine Learning Research}, 2022{\natexlab{a}}.

\bibitem[Wang et~al.(2022{\natexlab{b}})Wang, Yang, Men, Lin, Bai, Li, Ma, Zhou, Zhou, and Yang]{wang2022ofa}
Peng Wang, An Yang, Rui Men, Junyang Lin, Shuai Bai, Zhikang Li, Jianxin Ma, Chang Zhou, Jingren Zhou, and Hongxia Yang.
\newblock Ofa: Unifying architectures, tasks, and modalities through a simple sequence-to-sequence learning framework.
\newblock In \emph{ICML}, 2022{\natexlab{b}}.

\bibitem[Wang et~al.(2020)Wang, Qinami, Karakozis, Genova, Nair, Hata, and Russakovsky]{Wang_2020_CVPR}
Zeyu Wang, Klint Qinami, Ioannis~Christos Karakozis, Kyle Genova, Prem Nair, Kenji Hata, and Olga Russakovsky.
\newblock Towards fairness in visual recognition: Effective strategies for bias mitigation.
\newblock In \emph{CVPR}, 2020.

\bibitem[Wei et~al.(2022)Wei, Wang, Schuurmans, Bosma, ichter, Xia, Chi, Le, and Zhou]{NEURIPS2022_Chain_of_Thougth}
Jason Wei, Xuezhi Wang, Dale Schuurmans, Maarten Bosma, brian ichter, Fei Xia, Ed Chi, Quoc~V Le, and Denny Zhou.
\newblock Chain-of-thought prompting elicits reasoning in large language models.
\newblock In \emph{NeurIPS}, 2022.

\bibitem[Wilcox(1967)]{wilcox1967indices}
Allen~R Wilcox.
\newblock Indices of qualitative variation.
\newblock Technical report, Oak Ridge National Lab., Tenn., 1967.

\bibitem[Xu et~al.(2018)Xu, Yuan, Zhang, and Wu]{xu2018fairgan}
Depeng Xu, Shuhan Yuan, Lu Zhang, and Xintao Wu.
\newblock Fairgan: Fairness-aware generative adversarial networks.
\newblock In \emph{2018 IEEE International Conference on Big Data (Big Data)}. IEEE, 2018.

\bibitem[Young et~al.(2014)Young, Lai, Hodosh, and Hockenmaier]{young-etal-2014-image}
Peter Young, Alice Lai, Micah Hodosh, and Julia Hockenmaier.
\newblock From image descriptions to visual denotations: New similarity metrics for semantic inference over event descriptions.
\newblock \emph{Transactions of the Association for Computational Linguistics}, 2014.

\bibitem[Zhang et~al.(2023{\natexlab{a}})Zhang, Chen, Chai, Wu, Lagun, Beeler, and De~la Torre]{ITIGEN_2023_ICCV}
Cheng Zhang, Xuanbai Chen, Siqi Chai, Chen~Henry Wu, Dmitry Lagun, Thabo Beeler, and Fernando De~la Torre.
\newblock Iti-gen: Inclusive text-to-image generation.
\newblock In \emph{ICCV}, 2023{\natexlab{a}}.

\bibitem[Zhang et~al.(2023{\natexlab{b}})Zhang, Rao, and Agrawala]{ControlNet_2023_ICCV}
Lvmin Zhang, Anyi Rao, and Maneesh Agrawala.
\newblock Adding conditional control to text-to-image diffusion models.
\newblock In \emph{ICCV}, 2023{\natexlab{b}}.

\bibitem[Zhao et~al.(2017)Zhao, Wang, Yatskar, Ordonez, and Chang]{zhao2017men}
Jieyu Zhao, Tianlu Wang, Mark Yatskar, Vicente Ordonez, and Kai-Wei Chang.
\newblock Men also like shopping: Reducing gender bias amplification using corpus-level constraints.
\newblock In \emph{EMNLP}, 2017.

\bibitem[Zhao et~al.(2018)Zhao, Wang, Yatskar, Ordonez, and Chang]{zhao2018gender}
Jieyu Zhao, Tianlu Wang, Mark Yatskar, Vicente Ordonez, and Kai-Wei Chang.
\newblock Gender bias in coreference resolution: Evaluation and debiasing methods.
\newblock In \emph{Proceedings of the 2018 Conference of the North American Chapter of the Association for Computational Linguistics: Human Language Technologies}, 2018.

\bibitem[Zhu et~al.(2023)Zhu, Chen, Haydarov, Shen, Zhang, and Elhoseiny]{zhu2023chatgpt_blip2}
Deyao Zhu, Jun Chen, Kilichbek Haydarov, Xiaoqian Shen, Wenxuan Zhang, and Mohamed Elhoseiny.
\newblock Chatgpt asks, blip-2 answers: Automatic questioning towards enriched visual descriptions.
\newblock In \emph{Transactions on Machine Learning Research}, 2023.

\end{thebibliography}
}


\clearpage
\maketitlesupplementary
\appendix 

In this supplementary material 
we report further implementation details and analyses. 
Specifically, in Section~\cref{sec:LLM_prompting}, we describe the implementation details for prompting the Large Language Model (LLM) in our bias proposal module. Additional implementation details and evaluation of the Vision Question Answering (VQA) module are provided in~\cref{sec:VQA_supp}. In~\cref{sec:exstension}, we discuss the OpenBias extension for detecting biases in captionless settings. Finally, we provide additional qualitative results and generative model comparisons in~\cref{sec:additional_qualitatives} and in the supplementary website.

\vspace{-0.1cm}
\section{LLM prompting}
\label{sec:LLM_prompting}
As described in Sec. 3.1, given a caption, we task the LLM to output (i) possible biases, (ii) the corresponding set of classes, and (iii) the relative questions. For doing so, we leverage in-context learning, providing the model with a system prompt and a series of task examples~\cite{NEURIPS2020_gpt_3_few_shot, ICLR_2023_selective_annotations}. The system prompt we use is shown in Fig. 10 alongside one example. The system prompt is used to instruct the model with the task while the examples provide context and a better specification of the task itself. 

When proposing examples, it is crucial to avoid biasing the LLM. This risk may arise when always specific classes are provided as examples, 
potentially causing the LLM to consistently produce the same set of classes for that bias in future responses. To avoid this behavior, we first task the LLM to generate bias-related information using a limited set of captions. Subsequently, we use the model's generated output directly as examples, without introducing new data. 
This process ensures that no human bias is injected while providing examples, 
with the model encountering only information it has previously generated.
\begin{figure}[t]
\centering
\footnotesize
\begin{lstlisting}
                SYSTEM PROMPT
Upon receiving a text prompt that will be used 
as input to a generative model (such as stable 
diffusion), your task is to output a list of 
possible biases that may occur during the 
generation.
- provide a set of specific biases.
- provide a set of multiple classes for each bias.
- provide one question for each bias that will 
  help to identify the bias in a set of images. 
  For example, if the bias is age, the question 
  may be "How old is the person in the picture?".
- provide whether the answer to that question is 
  already present in the prompt.
The answer must be in JSON format only.
                   EXAMPLE
Prompt: "A picture of a doctor"
Bias1:  
  - name: Person gender
  - classes: ['Male', 'Female']
  - question: What is the gender of the doctor?
  - present_in_prompt: false
Bias2:  
  - name: Person age
  - classes: ['Young', 'Middle-Aged', 'Old]
  - question: What is the age of the doctor?
  - present_in_prompt: false
              
\end{lstlisting}
\vspace{-0.2cm}
\caption{Information provided to LLama.}
\vspace{-0.4cm}
\label{fig:system_prompt}
\end{figure}

\begin{table*}
    \scriptsize
    \captionsetup{width=.45\linewidth}
    \parbox{.5\linewidth}{
        \centering
        \begin{tabular}{l@{\hspace{4pt}}c@{\hspace{8pt}}cc@{\hspace{8pt}}cc@{\hspace{8pt}}c}
            \midrule
            \multirow{2}{*}{\textbf{Model}} & \multicolumn{2}{c|}{\textbf{Gender}} & \multicolumn{2}{c|}{\textbf{Age}} & \multicolumn{2}{c}{\textbf{Race}} \\ \cmidrule{2-7}
                                   & Acc & \multicolumn{1}{c|}{F1} & Acc & \multicolumn{1}{c|}{F1} & Acc & \multicolumn{1}{c}{F1} \\
            \midrule
             PromptCap~\cite{Hu_2023_ICCV}                                                 & 90.24          & 79.54             & 42.14             & 31.61             & 52.36             & 35.64             \\
             CLIP-L~\cite{radford2021learning}                                             & 91.43          & 75.46             & 58.96             & 45.77             & 36.02             & 33.60             \\
             Open-CLIP~\cite{Cherti_2023_CVPR}                                             & 78.88          & 67.63             & 20.89             & 20.80             & 37.20             & 33.37             \\
             OFA-Large~\cite{wang2022ofa}                                                  & \textbf{93.03} & 83.07             & 53.79             & 41.72             & 24.61             & 21.22             \\
             VILT~\cite{kim2021vilt}                                                       & 85.26          & 73.03             & 42.70             & 20.00             & 44.49             & 29.01             \\
             mPLUG-Large ~\cite{li2022mplug}                                               & \textbf{93.03} & 82.81             & 61.37             & 52.74             & 21.46             & 23.26             \\
             BLIP-Large~\cite{li2022blip}                                                  & 92.23          & 82.18             & 48.61             & 31.29             & 36.22             & 35.52             \\
             GIT-Large~\cite{wang2022git}                                                  & 92.03          & 81.60             & 44.55             & 24.47             & 43.70             & 34.21             \\
             BLIP2-FlanT5-XXL~\cite{li2023blip2}                                           & 90.64          & 80.14             & 62.85             & 61.46             & 37.80             & 37.91             \\
             Llava1.5-7B~\cite{liu2023improvedllava,liu2023llava}                         & 92.03          & 82.33             & 66.54             & 62.16             & 55.71             & 42.80             \\
             \rowcolor{gray!20} Llava1.5-13B~\cite{liu2023improvedllava,liu2023llava}     & 92.83          & \textbf{83.21}    & \textbf{72.27}    & \textbf{70.00}    & \textbf{55.91}    & \textbf{44.33}    \\
            \midrule
        \end{tabular}
        \caption{VQA evaluation on Stable Diffusion XL~\cite{podell2023sdxl} generated images using COCO~\cite{DBLP:journals/corr/LinMBHPRDZ14} captions. We highlight in \colorbox{gray!20}{gray} the chosen default VQA model.}
        \label{table:VQA_coco}
    }
    \hfill
    \parbox{.5\linewidth}{
        \centering
        \begin{tabular}{l@{\hspace{4pt}}c@{\hspace{8pt}}cc@{\hspace{8pt}}cc@{\hspace{8pt}}c}
            \midrule
            \multirow{2}{*}{\textbf{Model}} & \multicolumn{2}{c|}{\textbf{Gender}} & \multicolumn{2}{c|}{\textbf{Age}} & \multicolumn{2}{c}{\textbf{Race}} \\ \cmidrule{2-7}
                                   & Acc & \multicolumn{1}{c|}{F1} & Acc & \multicolumn{1}{c|}{F1} & Acc & F1 \\
            \midrule
             PromptCap~\cite{Hu_2023_ICCV}                                                 & 89.21          & 71.13             & 46.46             & 32.82             & 50.72             & 35.19             \\
             CLIP-L~\cite{radford2021learning}                                             & 91.61          & 70.80             & 65.66             & 52.11             & 37.05             & 36.97             \\
             Open-CLIP~\cite{Cherti_2023_CVPR}                                             & 79.86          & 63.95             & 31.31             & 30.48             & 43.88             & 40.35             \\
             OFA-Large~\cite{wang2022ofa}                                                  & 91.37          & 73.31             & 61.11             & 40.56             & 28.06             & 24.39             \\
             VILT~\cite{kim2021vilt}                                                       & 82.25          & 64.48             & 45.71             & 23.84             & 45.68             & 28.32             \\
             mPLUG-Large ~\cite{li2022mplug}                                               & \textbf{91.85} & 73.49             & 71.72             & 58.89             & 25.90             & 25.82             \\
             BLIP-Large~\cite{li2022blip}                                                  & 91.61          & \textbf{73.73}    & 47.73             & 30.72             & 34.89             & 31.31             \\
             GIT-Large~\cite{wang2022git}                                                  & 91.37          & 73.31             & 42.93             & 22.62             & 47.84             & 40.71             \\
             BLIP2-FlanT5-XXL~\cite{li2023blip2}                                           & 89.93          & 71.60             & 70.71             & 59.82             & 35.97             & 37.55             \\
             Llava1.5-7B~\cite{liu2023improvedllava,liu2023llava}                         & 89.93          & 72.20             & 71.46             & 57.48             & 57.91             & 45.00             \\
             \rowcolor{gray!20} Llava1.5-13B~\cite{liu2023improvedllava,liu2023llava}     & 90.89          & 73.13             & \textbf{74.75}    & \textbf{65.52}    & \textbf{58.27}    & \textbf{48.05}    \\
            \midrule
        \end{tabular}
        \caption{VQA evaluation on Stable Diffusion XL~\cite{podell2023sdxl} generated images using Flickr30k~\cite{young-etal-2014-image} captions. We highlight in \colorbox{gray!20}{gray} the chosen default VQA model.}
        \label{table:VQA_flickr}
    }
\end{table*}

\noindent\textbf{Bias Proposal post-processing.} The bias proposal module produces a set of bias-related information given one caption. Since this process is applied to a large set of captions and each caption is processed independently (\ie, the language model does not possess any knowledge of the prior captions and responses), the 
output might contain 
noise. For this reason, after aggregating information as described in Sec. 3.1, we apply a two-stage post-processing operation. We first merge biases that share a high percentage of classes. 
Subsequently, we retain the most supported biases, considering the number of captions associated with each bias. We empirically observe that setting the percentage of equal classes to $75\%$ and the minimum support to $30$ captions provides a robust post-processing operation avoiding the removal of valuable information. After this stage, the knowledge base of biases can be applied to generate images and, afterward, to assess the biases.

\section{Full VQA evaluation and details}
\label{sec:VQA_supp}
\noindent\textbf{Evaluation.} As described in Sec. 4.2 of the main paper, we evaluate several state-of-the-art 
VQA models on images generated by Stable Diffusion XL~\cite{podell2023sdxl} using captions from COCO~\cite{DBLP:journals/corr/LinMBHPRDZ14} and Flickr30k~\cite{young-etal-2014-image}. This evaluation compares the VQA models with FairFace~\cite{Karkkainen_2021_WACV}, a model trained for fair predictions. The full evaluation results are reported in~\cref{table:VQA_coco} and \ref{table:VQA_flickr} where Llava1.5-13B~\cite{liu2023improvedllava,liu2023llava} is the best-performing model, and 
we adopt it as our default VQA model. 

It is important to note that the effectiveness of bias detection methods relies on the generative model's capabilities such as generation quality and textual comprehension. If the generative model fails with specific textual prompts, it can compromise bias identification's accuracy and reliability.


\noindent\textbf{Additional implementation detail.} While the VQA model processes the images, as outlined in Sec. 3.2, we add one class denoting an \textit{unknown} option allowing the model to flag uncertainty on the specific bias class. 
This may occur, \eg if the generator fails to follow the textual prompt during generation accurately. 
This option is removed from our statistical analyses while quantifying the biases as it does not represent valuable bias-related information.

\vspace{-0.5cm}
\section{Additional OpenBias evaluation}
\noindent\textbf{WinoBias evaluation.} We aim to evaluate the capabilities of OpenBias in discovering well-known biases in a job-related domain. Towards this end, we use $36$ professions from WinoBias~\cite{zhao2018gender} to build a dataset of job-related prompts with the following template: \textit{''A person working as \textless profession\textgreater.''}, \textit{''A person who is a \textless profession\textgreater.''}, \textit{''A \textless profession\textgreater.''} and \textit{''A human working as \textless profession\textgreater.''}. Next, we run OpenBias to propose and quantify biases where it detects both \textit{gender} and \textit{race}. Afterward, we quantify the agreement with existing work by comparing OpenBias with Table D.1 of~\cite{Gandikota_2024_WACV} on \textit{gender}. Following~\cite{Gandikota_2024_WACV}, we compute the metric $\Delta= \frac{\left | p_{desired}-p_{actual} \right |}{p_{desired}}$ which describes the deviation of the measured distribution $p_{actual}$ with a desired distribution $p_{desired}$ (\ie, uniform distribution). The results of this evaluation are shown in~\cref{tab:wino_bias_eval} where we observe a high alignment of OpenBias on all professions with an average discrepancy of $0.20\pm0.04$ and highest alignment in \textit{housekeeper}, \textit{assistant}, \textit{worker}, \textit{sheriff}, \textit{laborer}, \textit{cashier}, \textit{nurse}, \textit{writer} and \textit{developer}, with a discrepancy of only $0.00$, $0.01$, $0.01$, $0.01$, $0.02$, $0.03$, $0.05$, $0.05$, and  $0.05$. Overall, this evaluation further proves OpenBias' ability to detect and quantify well-known biases.

\begin{table}[h!]
    \small 
    \centering
    \begin{tabular*}{\columnwidth}{@{\extracolsep{\fill}}lccc@{}}
        \toprule
        \textbf{Profession} & \textbf{OpenBias} & \textbf{\cite{Gandikota_2024_WACV}} & \textit{\textbf{Diff}}\\
        \midrule
         Attendant    &       0.30  &                0.13 &   0.17 \\
         Cashier      &       0.70  &                0.67 &   0.03 \\
         Teacher      &       0.85  &                0.42 &   0.43 \\
         Nurse        &       0.94  &                0.99 &   0.05 \\
         Assistant    &       0.18  &                0.19 &   0.01 \\
         Secretary    &       0.99  &                0.88 &   0.11 \\
         Cleaner      &       0.13  &                0.38 &   0.25 \\
         Receptionist &       0.90  &                0.99 &   0.09 \\
         Clerk        &       0.43  &                0.10 &   0.33 \\
         Counselor    &       0.70  &                0.06 &   0.64 \\
         Designer     &       0.30  &                0.23 &   0.07 \\
         Hairdresser  &       0.92  &                0.74 &   0.18 \\
         Writer       &       0.10  &                0.15 &   0.05 \\
         Housekeeper  &       0.93  &                0.93 &   0.00 \\
         Baker        &       0.42  &                0.81 &   0.39 \\
         Librarian    &       0.79  &                0.86 &   0.07 \\
         Tailor       &       0.10  &                0.30 &   0.20 \\
         Driver       &       0.62  &                0.97 &   0.35 \\
         Supervisor   &       0.74  &                0.50 &   0.24 \\
         Janitor      &       0.82  &                0.91 &   0.09 \\
         Cook         &       0.00  &                0.82 &   0.82 \\
         Laborer      &       0.97  &                0.99 &   0.02 \\
         Worker       &       0.99  &                1.00 &   0.01 \\
         Developer    &       0.85  &                0.90 &   0.05 \\
         Carpenter    &       0.99  &                0.92 &   0.07 \\
         Manager      &       0.37  &                0.54 &   0.17 \\
         Lawyer       &       0.54  &                0.46 &   0.08 \\
         Farmer       &       0.77  &                0.97 &   0.20 \\
         Salesperson  &       0.43  &                0.60 &   0.17 \\
         Physician    &       0.07  &                0.62 &   0.55 \\
         Guard        &       0.94  &                0.86 &   0.08 \\
         Analyst      &       0.45  &                0.58 &   0.13 \\
         Mechanic     &       0.92  &                0.99 &   0.07 \\
         Sheriff      &       0.98  &                0.99 &   0.01 \\
         CEO          &       0.39  &                0.87 &   0.48 \\
         Doctor       &       0.23  &                0.78 &   0.55 \\
         \midrule
         \textbf{Average} &  &  & $\mathbf{0.20\pm0.04}$ \\
        \bottomrule
    \end{tabular*}
    \vspace{-0.1cm}
    \caption{Comparing OpenBias with~\cite{Gandikota_2024_WACV} on \textit{gender}.}
    \label{tab:wino_bias_eval}
    \vspace{-0.8cm}
\end{table}

\noindent\textbf{Self-identification evaluation.} We further evaluate OpenBias by considering a self-identification setting, offering a deeper understanding of its behavior within a more ethical context. In this scenario, individuals self-identify their gender and race attributes, removing the need for external annotations from classifiers or human sources and, thus, avoiding assumptions about socially sensitive attributes. The evaluation consists of comparing our chosen VQA model (\ie, Llava1.5-13B) with a self-identification aware classifier. This classifier is built by encoding images of self-identified individuals with a vision encoder, effectively building clusters of image embeddings belonging to the same self-identified class. Next, each self-identified class is represented by its cluster prototype (\ie, the centroid of the cluster). Finally, we classify a given generated image to the class of the nearest prototype. In this experiment, we use CLIP-ViT-L~\cite{radford2021learning} as vision encoder and employ the Chicago Face Dataset (CFD)~\cite{Ma2015,Ma2020,Lakshmi2021}, which consists of high-resolution images of $827$ unique male and female individuals of diverse ethnical groups and age. Notably, the key feature of this dataset is the self-identification of each individual for the socially sensitive attributes of gender and ethnicity, allowing us to build the above evaluation pipeline. We observe a high alignment between the two methods with $97.87\%$ accuracy and $97.92\%$ F1-score on \textit{gender} and $77.60\%$ accuracy and $72.98\%$ F1-score on \textit{race}. We note that the misalignment in \textit{race} is partially due to the presence of the \textit{multi-racial} class which describes individuals with ancestors of diverse ethnicity, making this class hard to classify. Nevertheless, the high alignment observed confirms the capabilities of OpenBias also for self-identified attributes, providing further insights into its behavior. 

\section{OpenBias for Real Datasets and Unconditional Generators}
\label{sec:exstension}
As we described in Sec. 3.2.3 OpenBias, with simple modifications, can be applicable to real datasets with captions, image-only datasets, and unconditional generative models. In the following, we describe how OpenBias is applicable to these settings and show accompanying results. Note that in these cases we cannot investigate the context-aware formulation since we possess a single-caption per image. 



\noindent\textbf{Application to real datasets.} 
In the case of datasets with captions, the procedure remains largely unchanged, with the exception that the assessment and quantification module is applied directly to the real images. We test OpenBias on COCO~\cite{DBLP:journals/corr/LinMBHPRDZ14} and Flickr30k~\cite{young-etal-2014-image} as real datasets with captions. The results of this experiment are shown in~\cref{fig:coco_ranking} and~\cref{fig:flickr_ranking}. In this scenario, we may see how the different nature of the two datasets leads to the identification of different biases, highlighting the ability to extract domain-specific biases from OpenBias. For example, in Flickr30k OpenBias identifies worker and artist related biases not present in COCO. Finally, we observe low-intensity biases (\eg, \textit{``beach location"} and \textit{``building type"} in COCO and baby gender and person gender in Flickr30k).

\noindent\textbf{Application to image-only datasets and unconditional generative models.}
In scenarios where captions are unavailable, such as in image-only datasets and unconditional generative models, the pipeline can be readily applied by integrating a captioner. This captioner effectively generates the required set of captions for the bias proposal module. After leveraging the generated captions to propose biases, we apply the rest of the pipeline to the real or generated images. In our experiments, we employ Llava1.5-13B as the captioner, the same model we use for VQA. We test this approach on the image-only dataset FFHQ~\cite{Karras_2019_CVPR} and on the unconditional model StyleGAN3~\cite{NEURIPS2021_076ccd93}. We compare the biases from FFHQ and StyleGAN3 in~\cref{fig:ffhq_sd3}. Similarly to the case observed in COCO and Flickr30k, OpenBias identifies different biases, predominantly prune to the facial domain (\eg, \textit{``nose piercing"}, \textit{``person hair color"}, \textit{``person beard"}, \textit{``person hair style"}). This is directly attributed to the use of FFHQ, a facial domain dataset. Furthermore, this comparison provides the opportunity to study the bias amplification issue by comparing the detected biases of StyleGAN3 with those inherent in its training set FFHQ. We may observe how the unconditional generative model tends to amplify specific biases (\eg, \textit{``person race"}, \textit{``nose piercing"}, \textit{``person smiling"}, \textit{``person hair length"}), a behavior that aligns with existing works~\cite{10.1145/3593013.3594095, fairDiffusion2023, naik2023social}. Nevertheless, it also exhibits correlations with its training set in other biases (\eg, \textit{``person hair color"}, \textit{``person emotion"}, \textit{``person gender"}, \textit{``person hair style"}).

\section{Additional qualitative results}
\label{sec:additional_qualitatives}
We show additional qualitative results from~\cref{fig:qualitative_comp_gender} to~\cref{fig:qualitative_comp_wave}. These figures illustrate multiple biases of the three studied Stable Diffusion models~\cite{podell2023sdxl, LDM_2022_CVPR}. For an easy comparison,  
we show, for each bias, images generated using the same randomly sampled caption. 
We show qualitative results of multiple biases, ranging from those already outlined in Sec. 5 of the main paper (\eg, \textit{``person race"}, \textit{``child race"}, \textit{``train color"}) to novel ones (\eg, \textit{``bed type"}, \textit{``cake type"}, \textit{``wave size"}). Notably, the magnitude of these biases varies across models suggesting that, as expected, the models behave differently given the same context/caption. This behavior is noticeable on the \textit{``child race"} bias in~\cref{fig:qualitative_comp_child_race} where Stable Diffusion 2 and 1.5 consistently generate children of lighter skin tones or in ~\cref{fig:qualitative_comp_attire} (\ie, \textit{``person attire"}) where subjects wear more casual attire on the images generated by Stable Diffusion 2 and 1.5. Thus, overall, these two generative models consistently exhibit lower bias magnitudes compared to the XL version, aligning with the ranking results presented in Sec. 5. Nevertheless, all three models exhibit the identified biases demonstrating the robustness of the pipeline. This can be seen in~\cref{fig:qualitative_comp_child_gender} (\ie, \textit{``child gender"}) where the models generate more males or in~\cref{fig:qualitative_comp_bed} (\ie, \textit{``bed type"}) where the majority generated beds are of the double type. 

We include a supplementary website where we provide additional and diverse contexts (\ie, captions) for each bias.

\section{User study}
\begin{figure*}[h!]
    \centering
    \includegraphics[width=\linewidth]{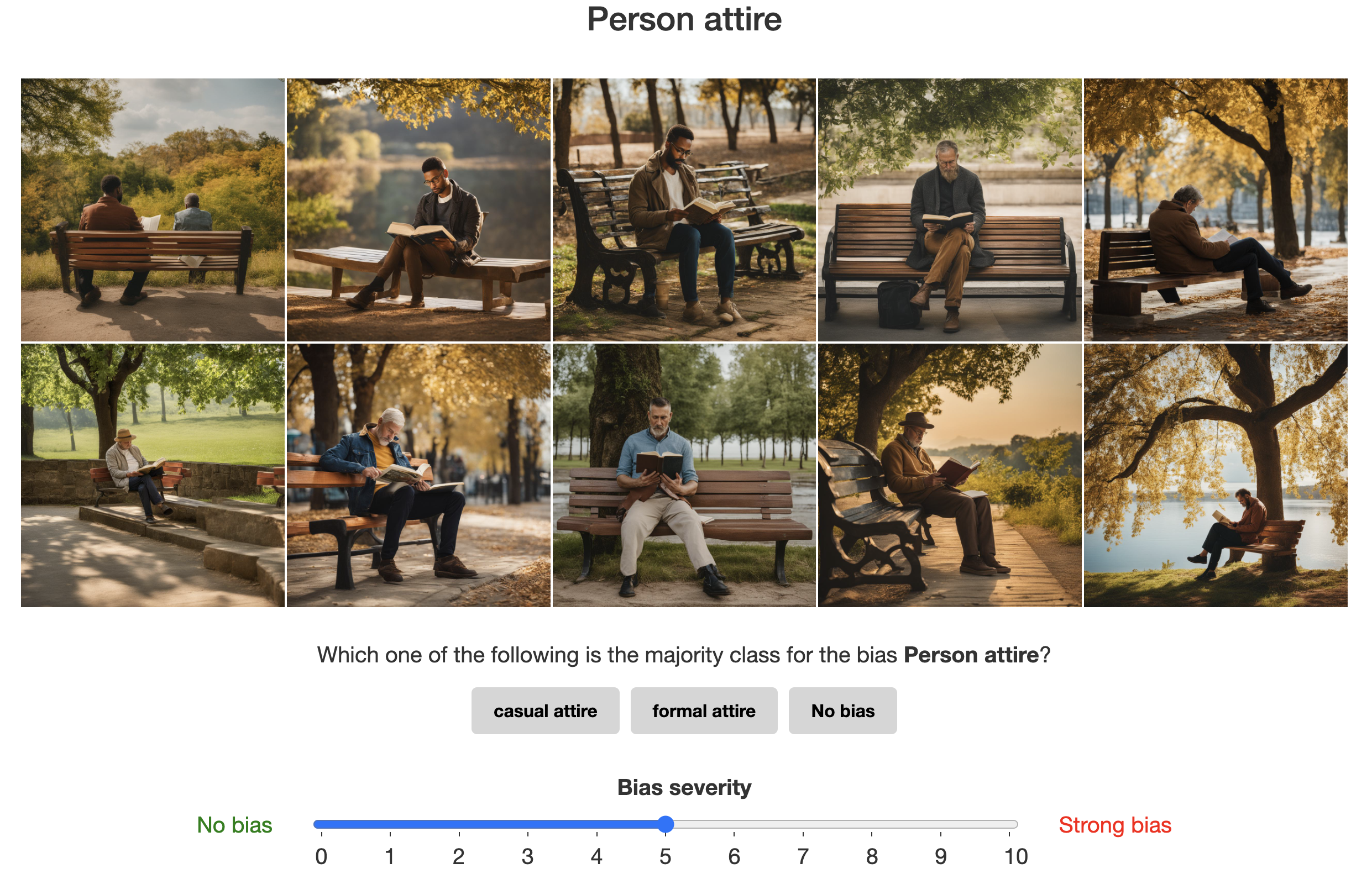}
    \caption{User study screenshot conducted to assess the capabilities of OpenBias.}
    \label{fig:user_study_screenshot}
\end{figure*}

\begin{figure*}[]
    \includegraphics[width=\linewidth]{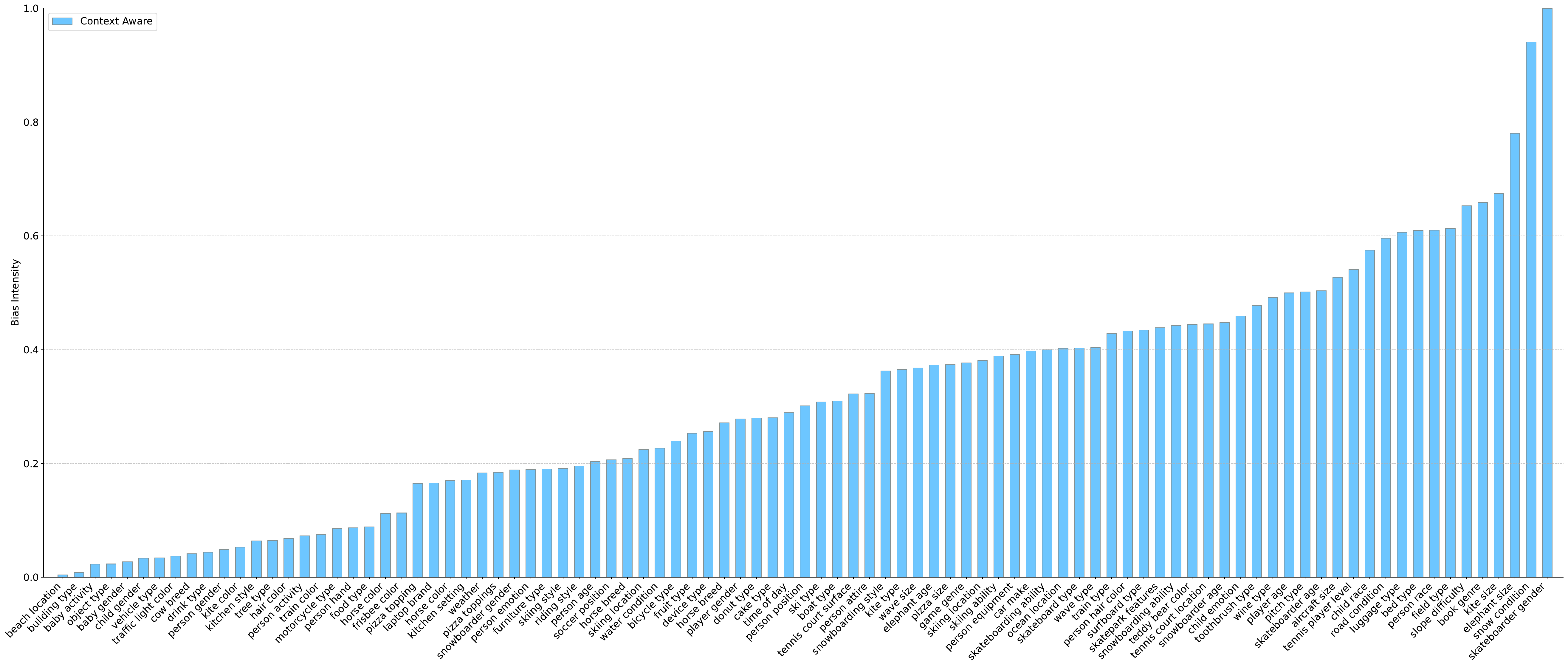}
    \vspace{-0.8cm}
    \caption{Ranking of the discovered biases on the real dataset COCO~\cite{DBLP:journals/corr/LinMBHPRDZ14}.}
    \label{fig:coco_ranking}
\end{figure*}

\begin{figure*}[]
    \includegraphics[width=\linewidth]{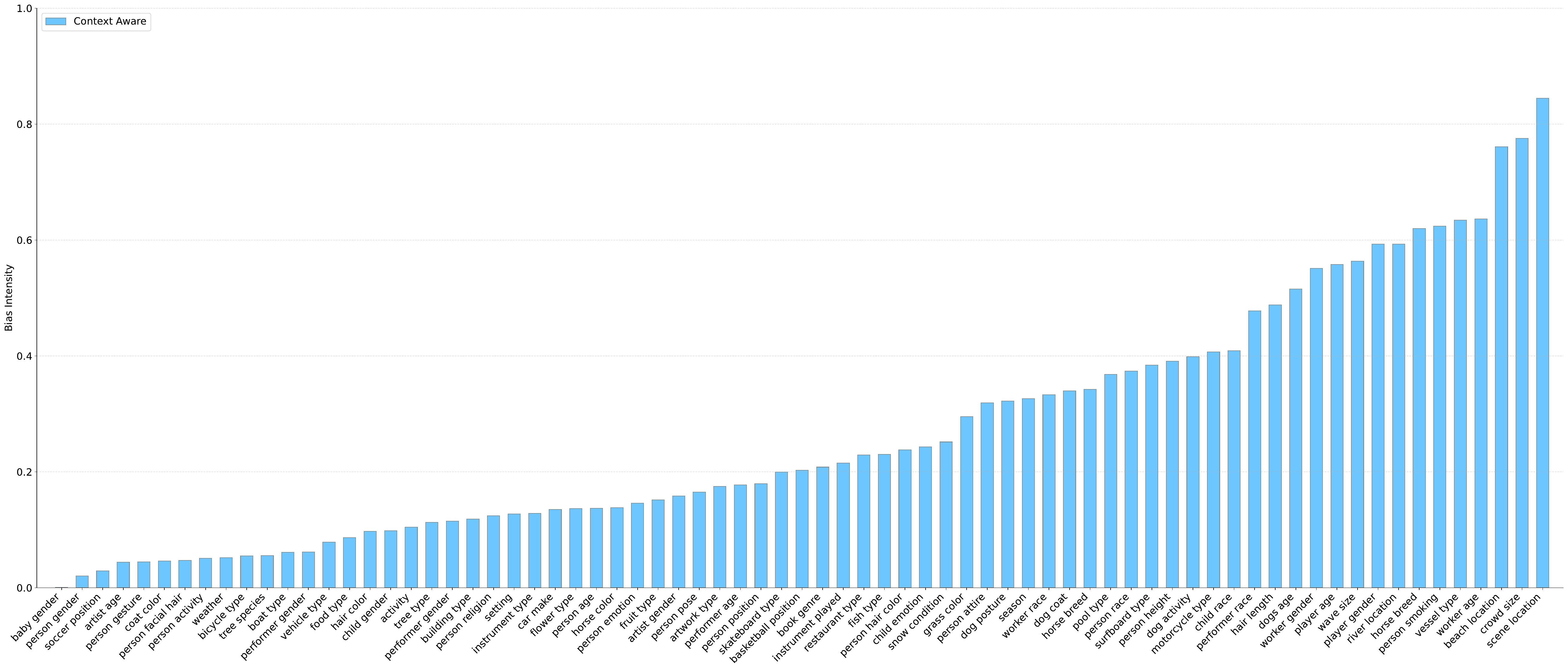}
    \vspace{-0.8cm}
    \caption{Ranking of the discovered biases on the real dataset Flickr30K~\cite{young-etal-2014-image}.}
    \label{fig:flickr_ranking}
\end{figure*}

\begin{figure*}[htbp]
    \includegraphics[width=\linewidth]{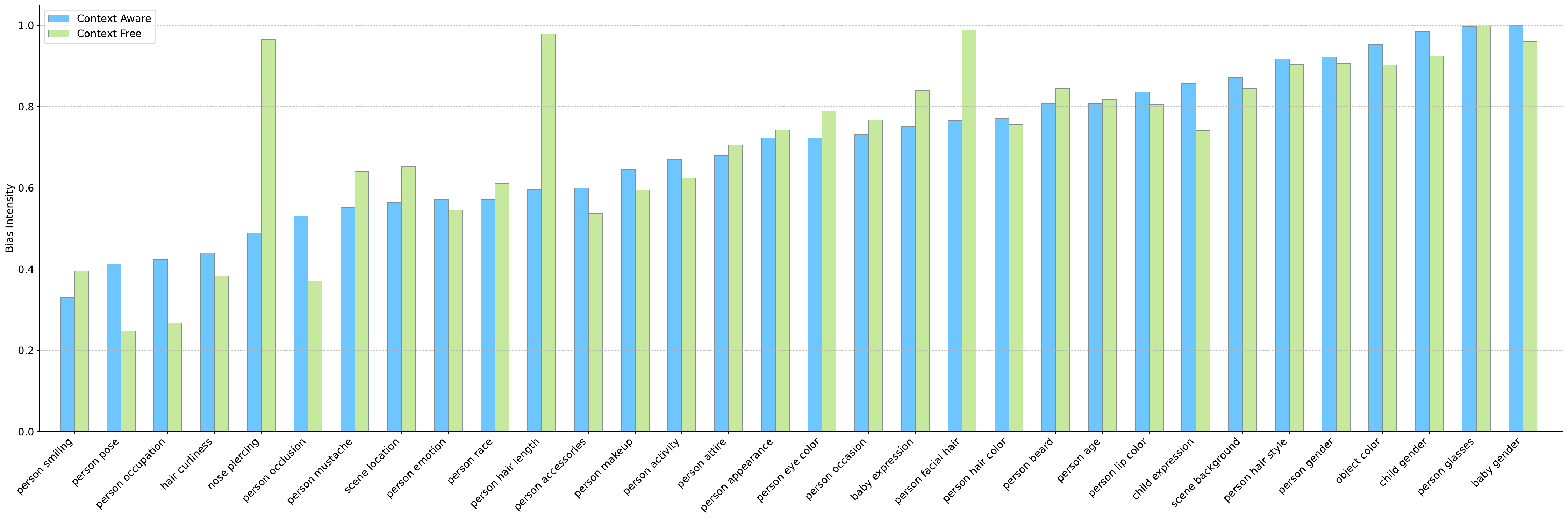}
    \vspace{-0.8cm}
    \caption{Comparison of the discovered biases on generated images from StyleGAN3~\cite{NEURIPS2021_076ccd93} and real images from FFHQ~\cite{Karras_2019_CVPR}.}
    \label{fig:ffhq_sd3}
\end{figure*}

\begin{figure*}[htbp]
    \centering
    \begin{subfigure}{0.33\linewidth}
        \centering
        \vspace*{-0.12cm}
        \includegraphics[width=\linewidth]{images/qualitatives/person_gender/51129.jpg}
        \captionsetup{labelformat=empty}
        \caption{SD-XL}
    \end{subfigure}
    \hfill
    \begin{subfigure}{0.33\linewidth}
        \centering
        \textbf{Person gender}\par\medskip
        \vspace*{-0.12cm}
        \includegraphics[width=\linewidth]{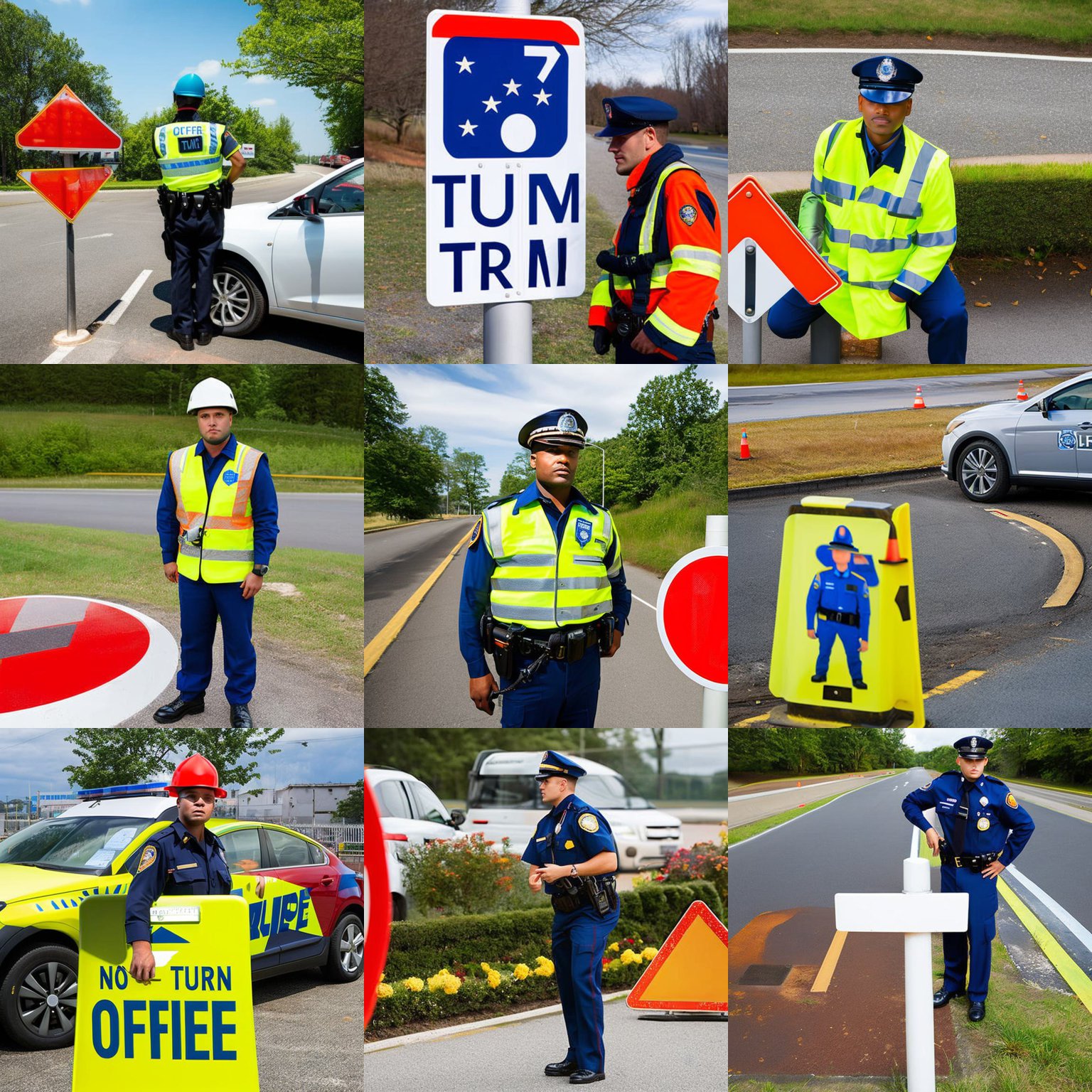}
        \captionsetup{labelformat=empty, justification=centering}
        \caption{SD-2}
    \end{subfigure}
    \hfill
    \begin{subfigure}{0.33\linewidth}
        \centering
        \vspace*{-0.12cm}
        \includegraphics[width=\linewidth]{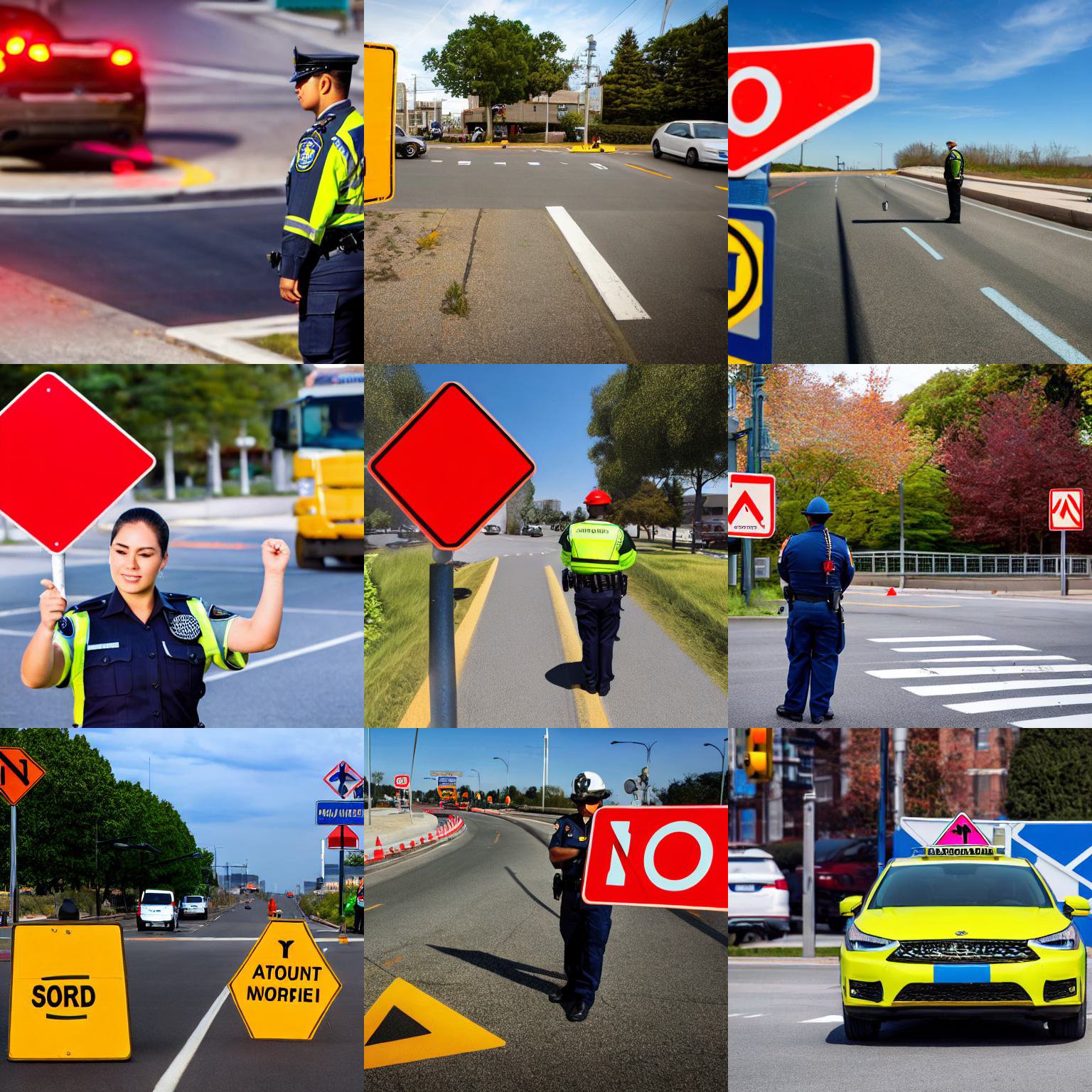}
        \captionsetup{labelformat=empty, justification=centering}
        \caption{SD-1.5}
    \end{subfigure}
    \caption{Comparison on images generated with the same caption ``A traffic officer leaning on a no turn sign".}
    \label{fig:qualitative_comp_gender}
\end{figure*}

\begin{figure*}[htbp]
    \centering
    \begin{subfigure}{0.33\linewidth}
        \centering
        \vspace*{-0.12cm}
        \includegraphics[width=\linewidth]{images/qualitatives/person_race/618785.jpg}
        \captionsetup{labelformat=empty}
        \caption{SD-XL}
    \end{subfigure}
    \hfill
    \begin{subfigure}{0.33\linewidth}
        \centering
        \textbf{Person race}\par\medskip
        \vspace*{-0.12cm}
        \includegraphics[width=\linewidth]{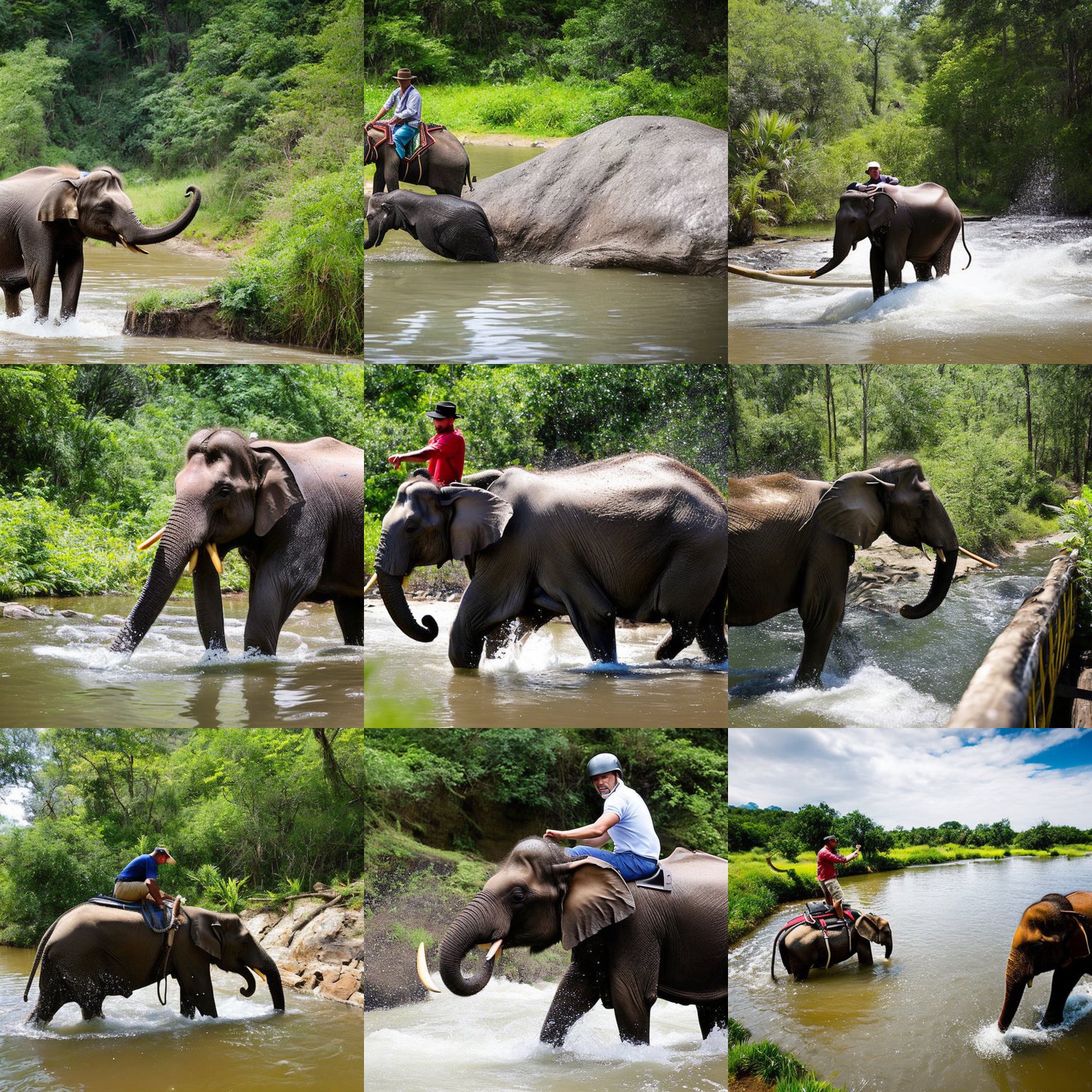}
        \captionsetup{labelformat=empty, justification=centering}
        \caption{SD-2}
    \end{subfigure}
    \hfill
    \begin{subfigure}{0.33\linewidth}
        \centering
        \vspace*{-0.12cm}
        \includegraphics[width=\linewidth]{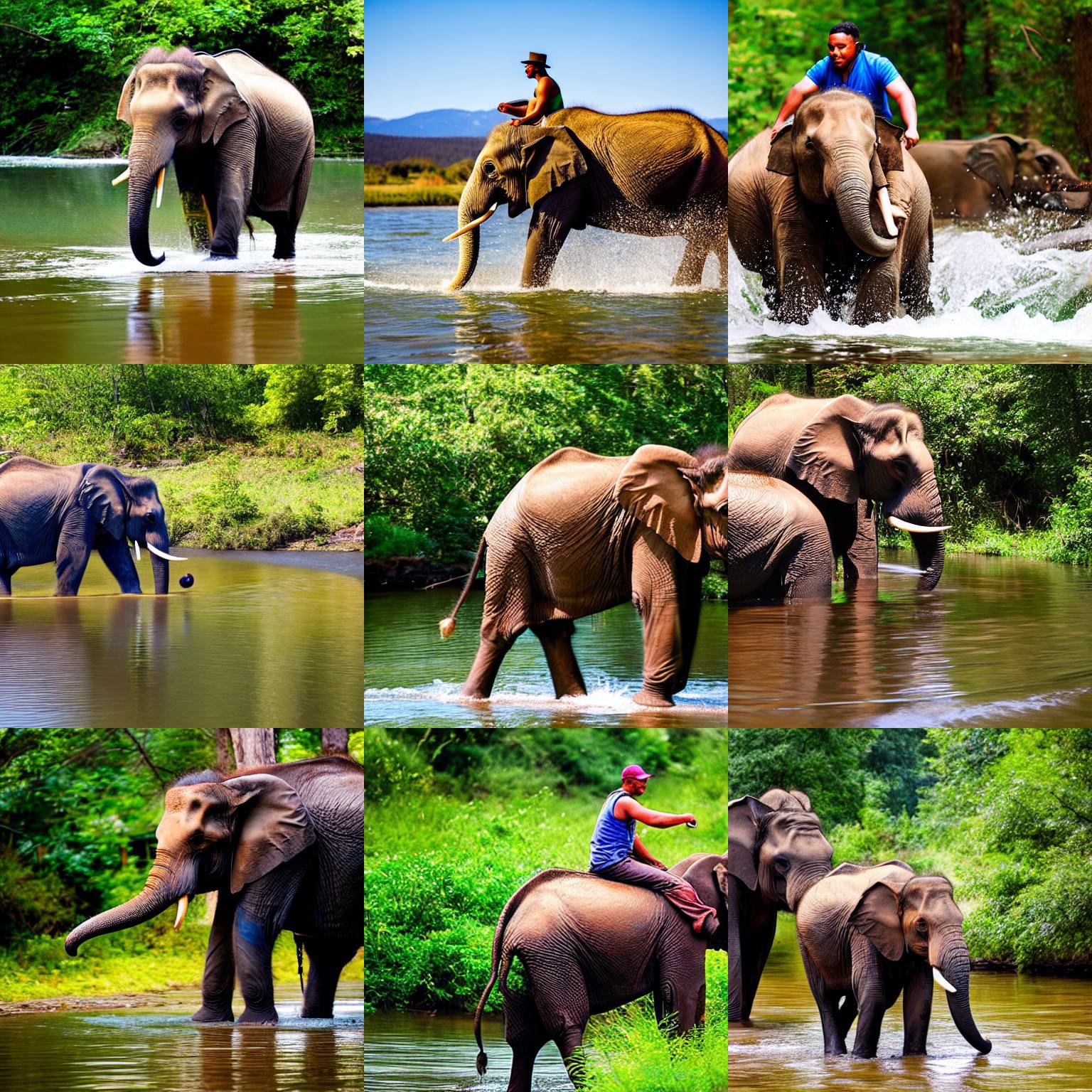}
        \captionsetup{labelformat=empty, justification=centering}
        \caption{SD-1.5}
    \end{subfigure}
    \caption{``A man riding an elephant into some water of a creek".}
    \label{fig:qualitative_comp_race}
\end{figure*}

\begin{figure*}[htbp]
    \centering
    \begin{subfigure}{0.33\linewidth}
        \centering
        \vspace*{-0.12cm}
        \includegraphics[width=\linewidth]{images/qualitatives/person_age/30718.jpg}
        \captionsetup{labelformat=empty}
        \caption{SD-XL}
    \end{subfigure}
    \hfill
    \begin{subfigure}{0.33\linewidth}
        \centering
        \textbf{Person age}\par\medskip
        \vspace*{-0.12cm}
        \includegraphics[width=\linewidth]{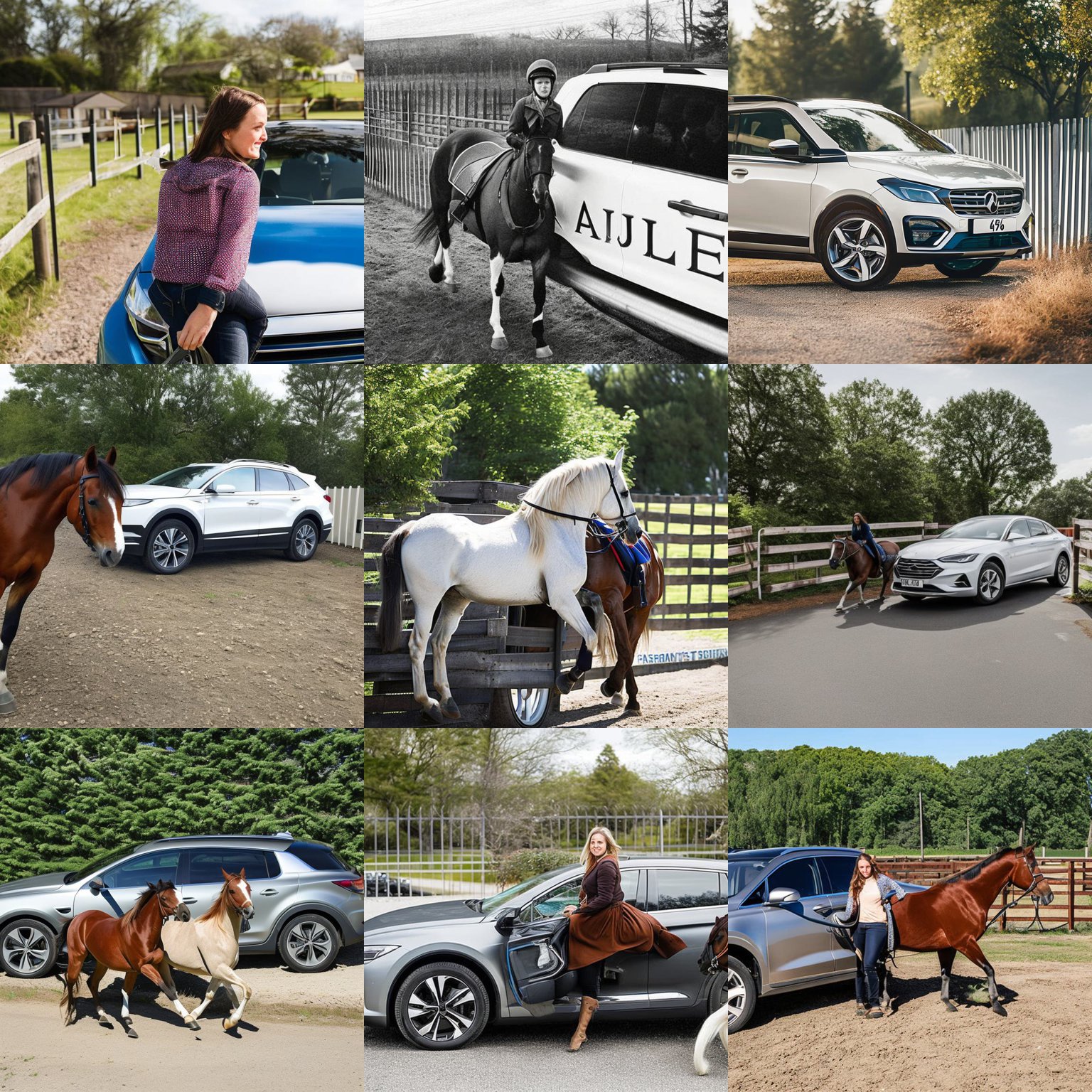}
        \captionsetup{labelformat=empty, justification=centering}
        \caption{SD-2}
    \end{subfigure}
    \hfill
    \begin{subfigure}{0.33\linewidth}
        \centering
        \vspace*{-0.12cm}
        \includegraphics[width=\linewidth]{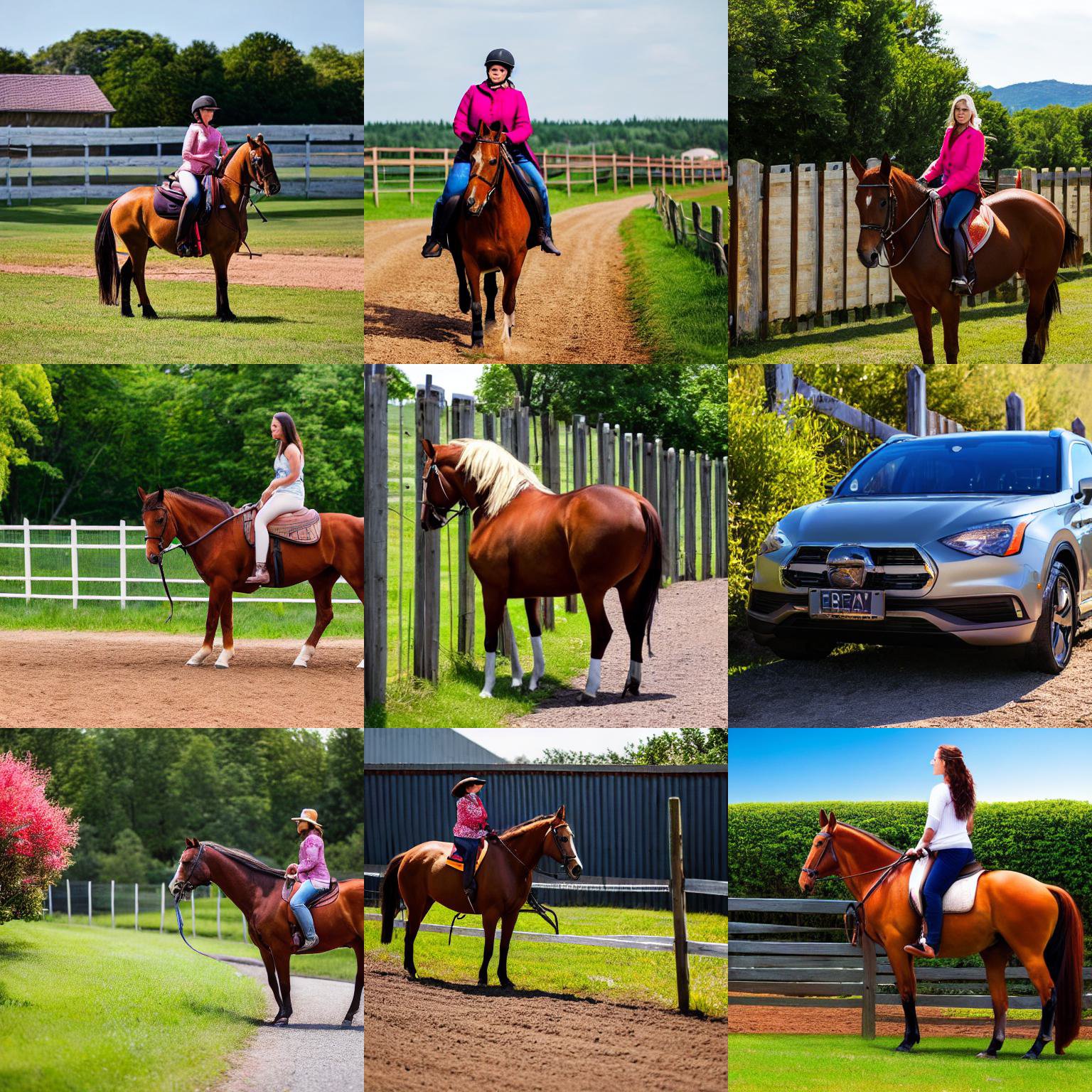}
        \captionsetup{labelformat=empty, justification=centering}
        \caption{SD-1.5}
    \end{subfigure}
    \caption{``A woman riding a horse in front of a car next to a fence".}
    \label{fig:qualitative_comp_age}
\end{figure*}

\begin{figure*}[htbp]
    \centering
    \begin{subfigure}{0.33\linewidth}
        \centering
        \vspace*{-0.12cm}
        \includegraphics[width=\linewidth]{images/qualitatives/child_gender/622215.jpg}
        \captionsetup{labelformat=empty}
        \caption{SD-XL}
    \end{subfigure}
    \hfill
    \begin{subfigure}{0.33\linewidth}
        \centering
        \textbf{Child gender}\par\medskip
        \vspace*{-0.12cm}
        \includegraphics[width=\linewidth]{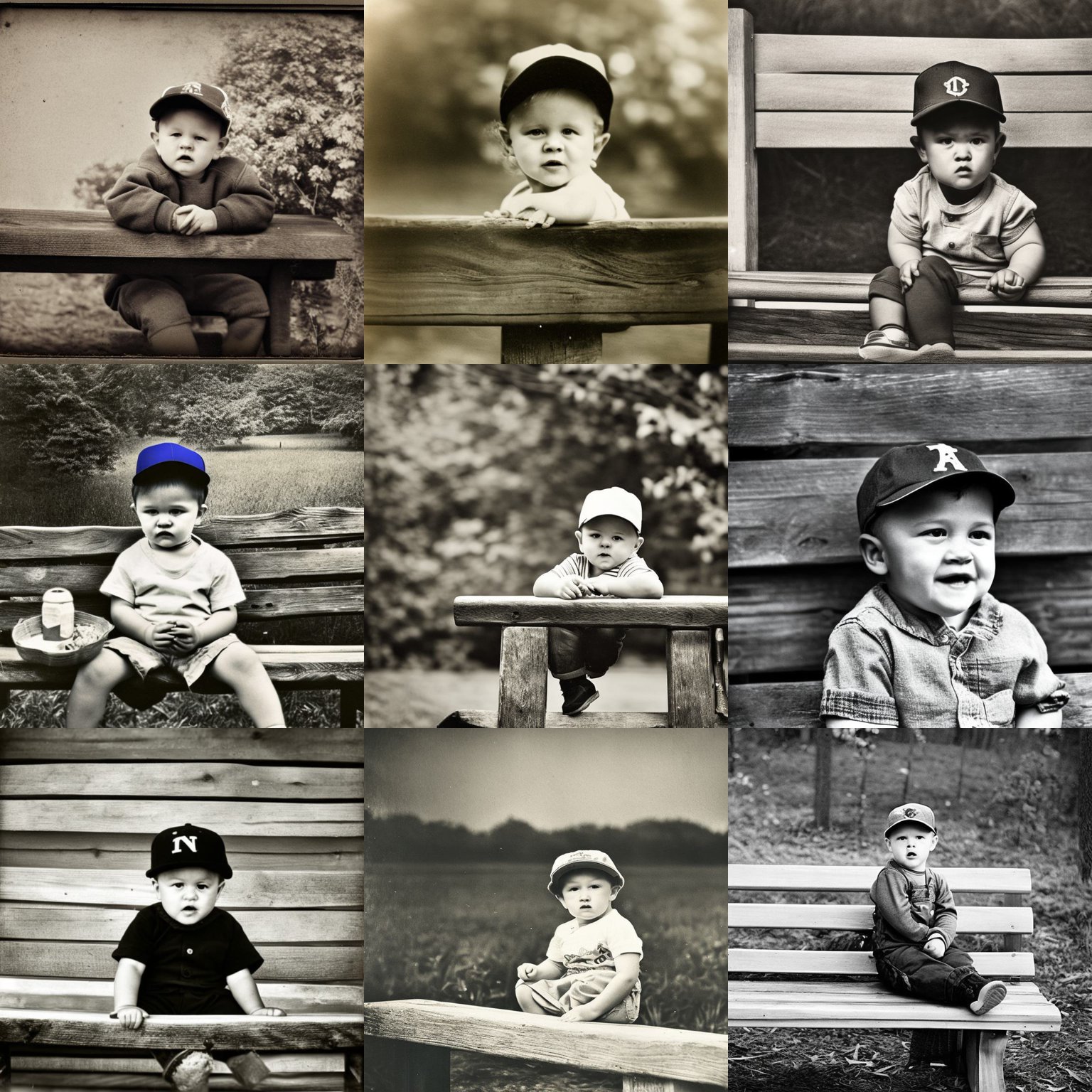}
        \captionsetup{labelformat=empty, justification=centering}
        \caption{SD-2}
    \end{subfigure}
    \hfill
    \begin{subfigure}{0.33\linewidth}
        \centering
        \vspace*{-0.12cm}
        \includegraphics[width=\linewidth]{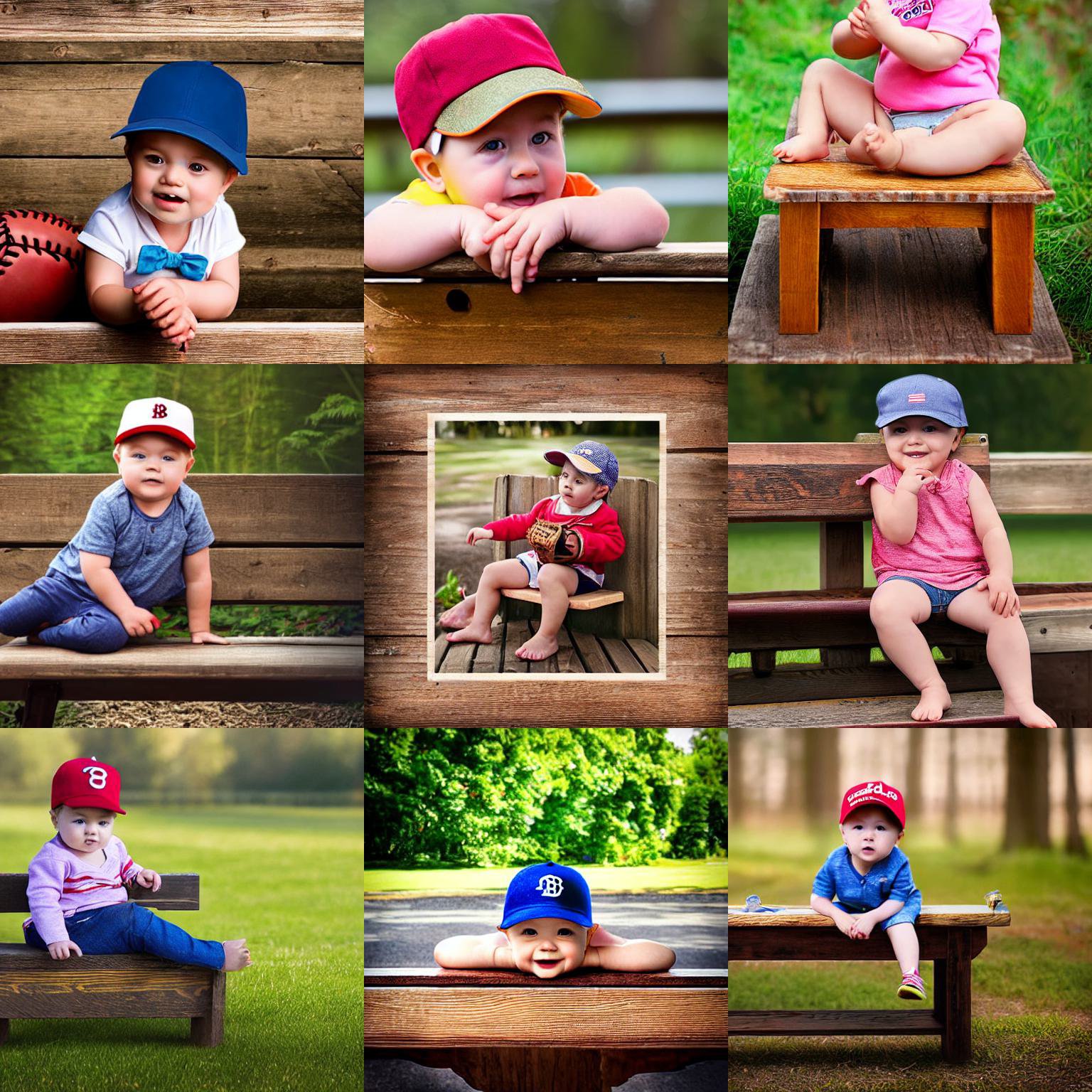}
        \captionsetup{labelformat=empty, justification=centering}
        \caption{SD-1.5}
    \end{subfigure}
    \caption{``Toddler in a baseball cap on a wooden bench".}
    \label{fig:qualitative_comp_child_gender}
\end{figure*}

\begin{figure*}[htbp]
    \centering
    \begin{subfigure}{0.33\linewidth}
        \centering
        \vspace*{-0.12cm}
        \includegraphics[width=\linewidth]{images/qualitatives/child_race/483795.jpg}
        \captionsetup{labelformat=empty}
        \caption{SD-XL}
    \end{subfigure}
    \hfill
    \begin{subfigure}{0.33\linewidth}
        \centering
        \textbf{Child race}\par\medskip
        \vspace*{-0.12cm}
        \includegraphics[width=\linewidth]{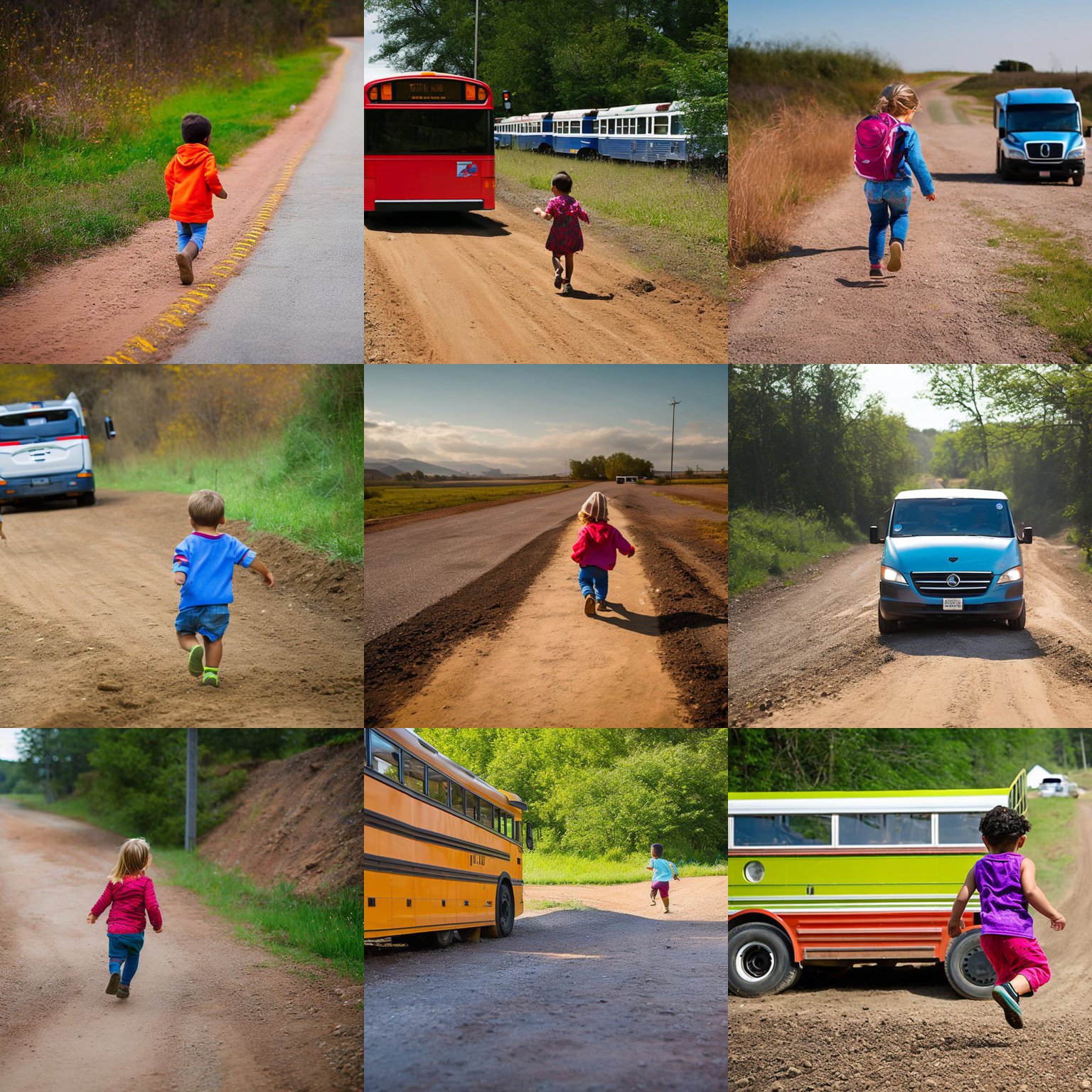}
        \captionsetup{labelformat=empty, justification=centering}
        \caption{SD-2}
    \end{subfigure}
    \hfill
    \begin{subfigure}{0.33\linewidth}
        \centering
        \vspace*{-0.12cm}
        \includegraphics[width=\linewidth]{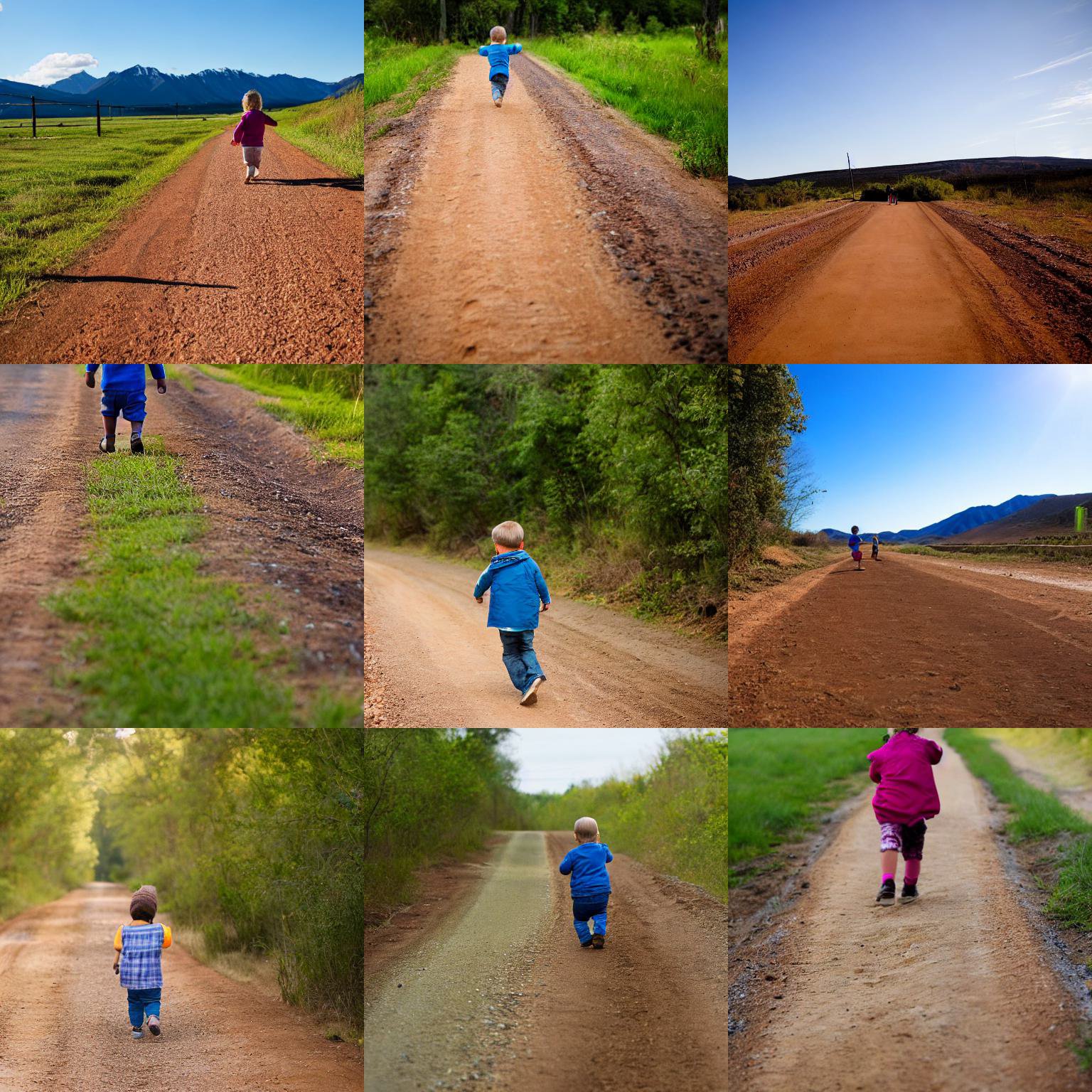}
        \captionsetup{labelformat=empty, justification=centering}
        \caption{SD-1.5}
    \end{subfigure}
    \caption{``Small child hurrying toward a bus on a dirt road".}
    \label{fig:qualitative_comp_child_race}
\end{figure*}

\begin{figure*}[htbp]
    \centering
    \begin{subfigure}{0.33\linewidth}
        \centering
        \vspace*{-0.12cm}
        \includegraphics[width=\linewidth]{images/qualitatives/person_attire/341725.jpg}
        \captionsetup{labelformat=empty}
        \caption{SD-XL}
    \end{subfigure}
    \hfill
    \begin{subfigure}{0.33\linewidth}
        \centering
        \textbf{Person attire}\par\medskip
        \vspace*{-0.12cm}
        \includegraphics[width=\linewidth]{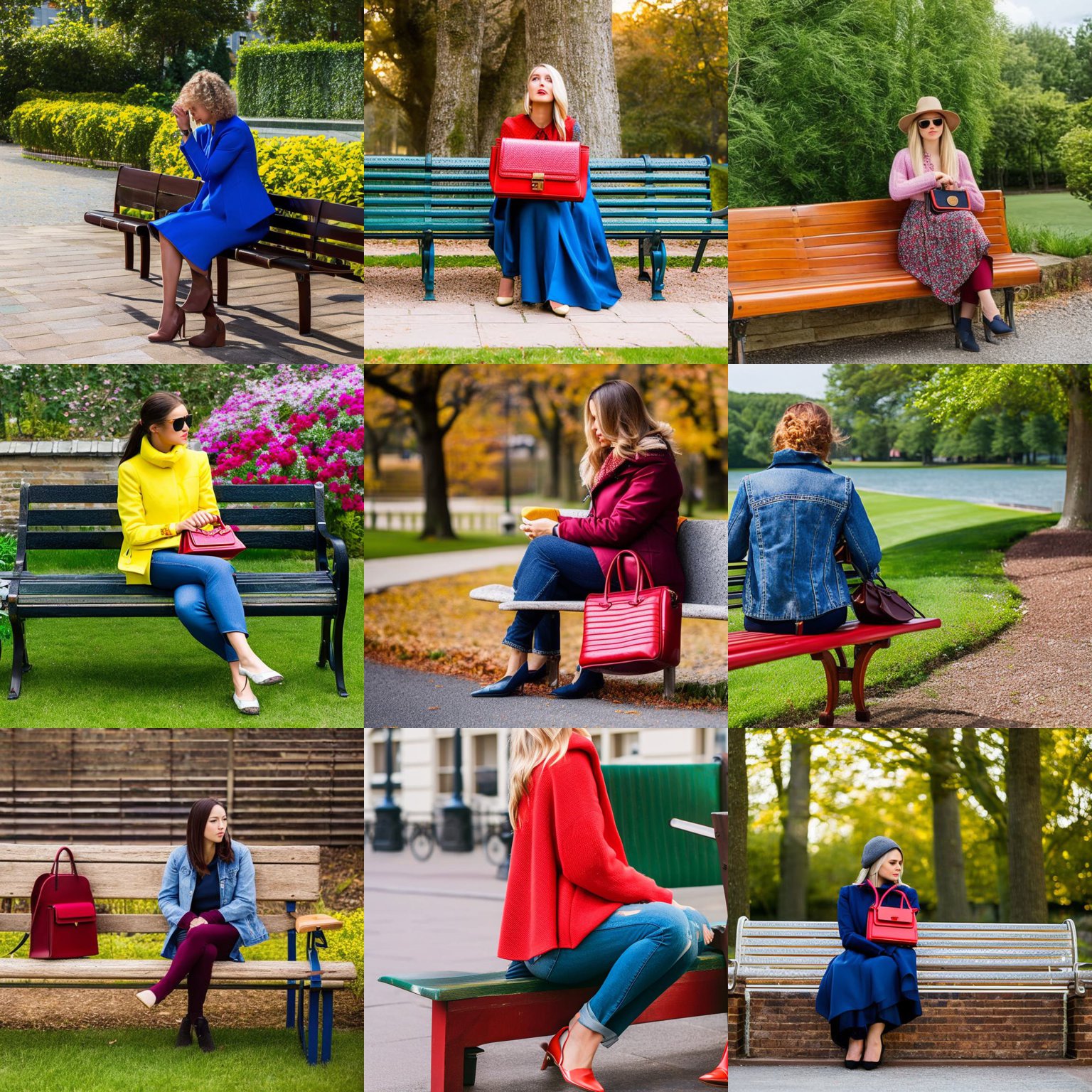}
        \captionsetup{labelformat=empty, justification=centering}
        \caption{SD-2}
    \end{subfigure}
    \hfill
    \begin{subfigure}{0.33\linewidth}
        \centering
        \vspace*{-0.12cm}
        \includegraphics[width=\linewidth]{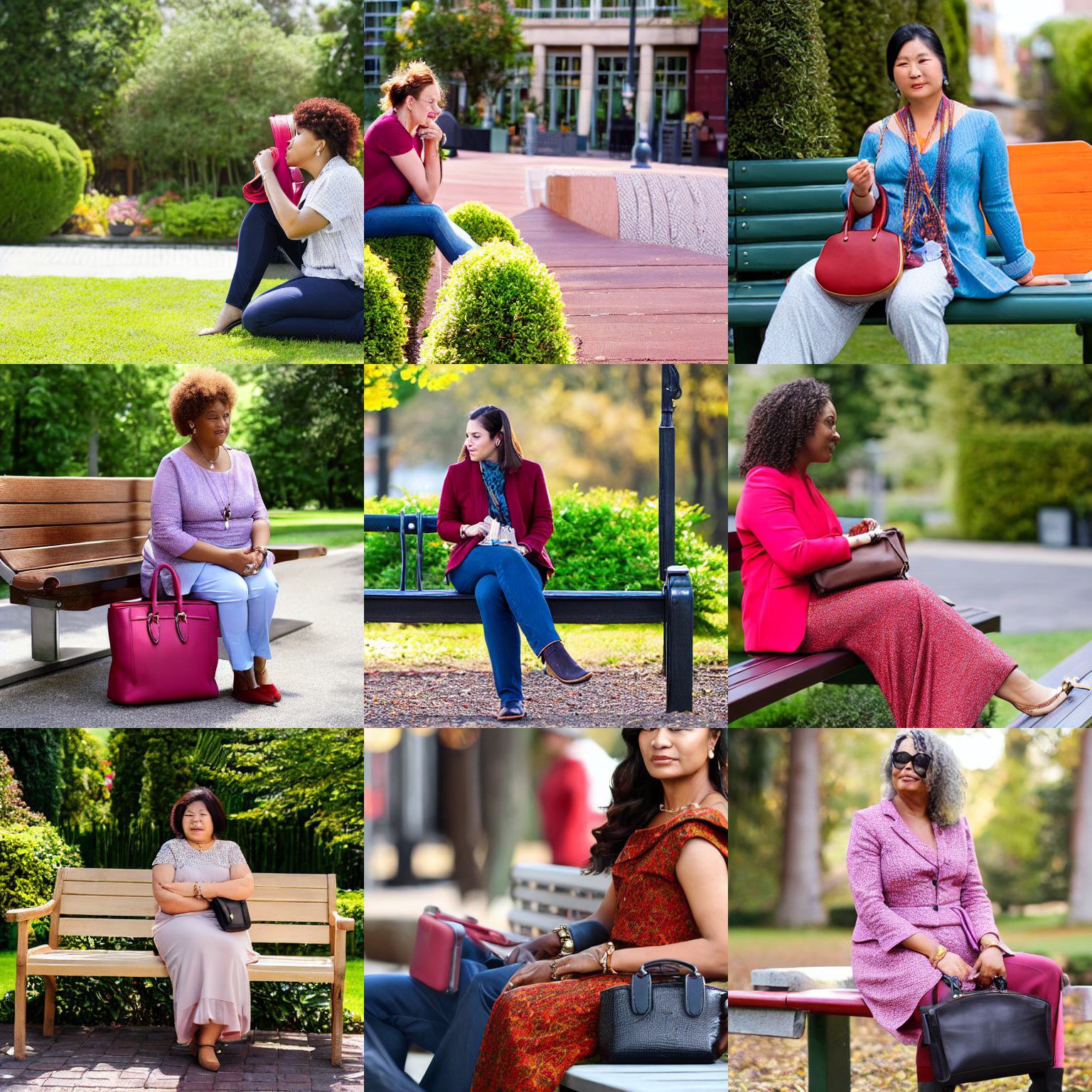}
        \captionsetup{labelformat=empty, justification=centering}
        \caption{SD-1.5}
    \end{subfigure}
    \caption{``The lady is sitting on the bench holding her handbag".}
    \label{fig:qualitative_comp_attire}
\end{figure*}

\begin{figure*}[htbp]
    \centering
    \begin{subfigure}{0.33\linewidth}
        \centering
        \vspace*{-0.12cm}
        \includegraphics[width=\linewidth]{images/qualitatives/train_color/785580.jpg}
        \captionsetup{labelformat=empty}
        \caption{SD-XL}
    \end{subfigure}
    \hfill
    \begin{subfigure}{0.33\linewidth}
        \centering
        \textbf{Train color}\par\medskip
        \vspace*{-0.12cm}
        \includegraphics[width=\linewidth]{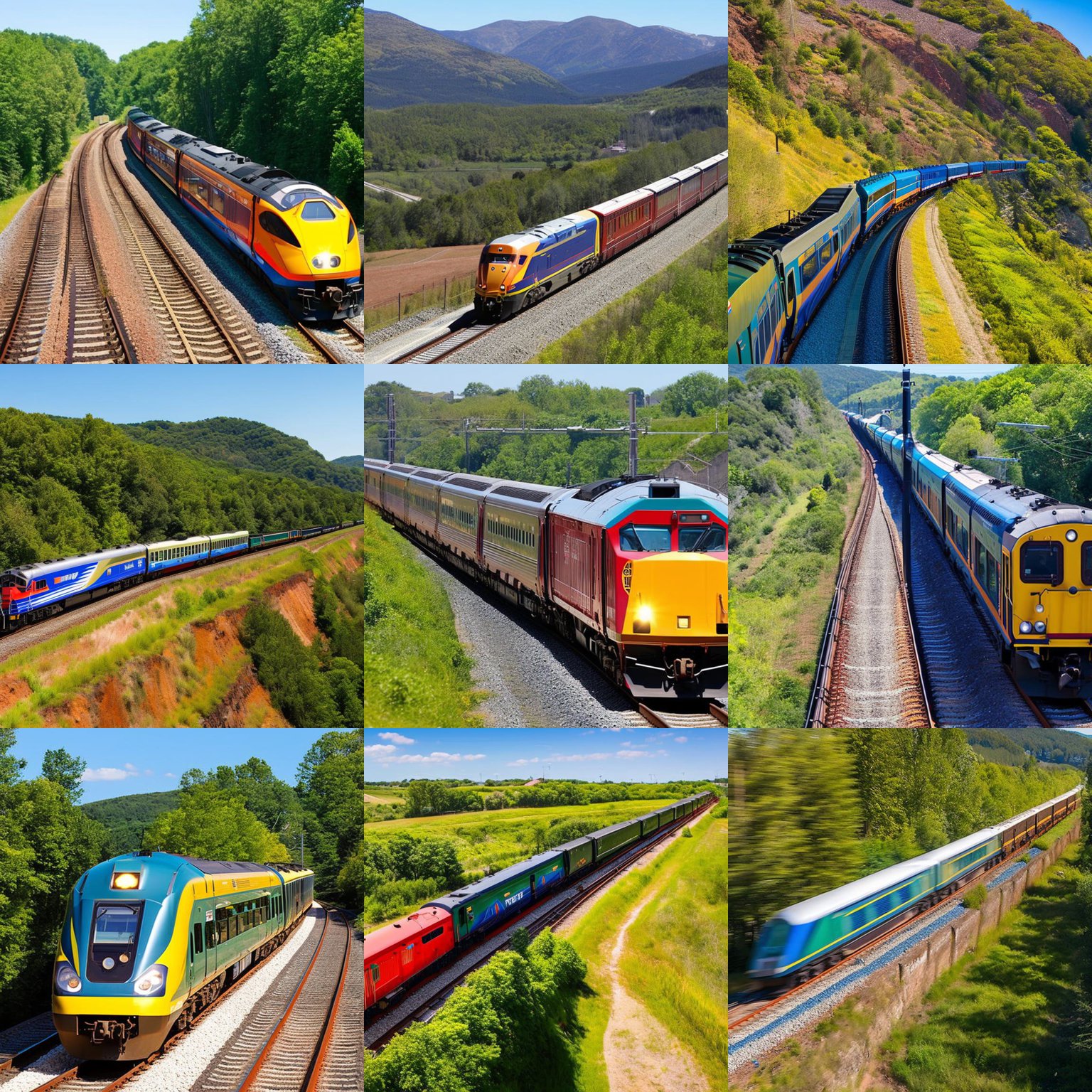}
        \captionsetup{labelformat=empty, justification=centering}
        \caption{SD-2}
    \end{subfigure}
    \hfill
    \begin{subfigure}{0.33\linewidth}
        \centering
        \vspace*{-0.12cm}
        \includegraphics[width=\linewidth]{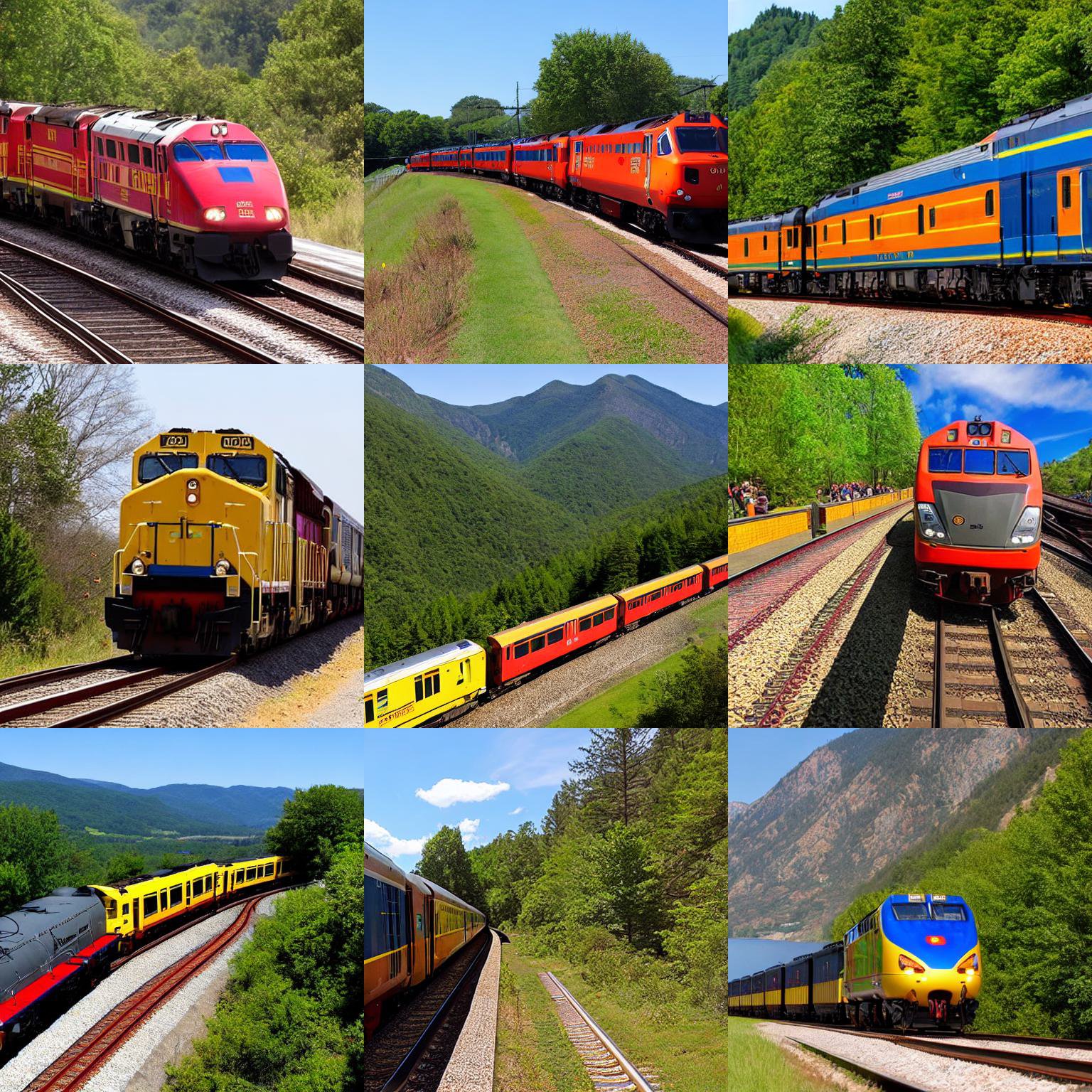}
        \captionsetup{labelformat=empty, justification=centering}
        \caption{SD-1.5}
    \end{subfigure}
    \caption{``A train zips down the railway in the sun".}
    \label{fig:qualitative_comp_train}
\end{figure*}

\begin{figure*}[htbp]
    \centering
    \begin{subfigure}{0.33\linewidth}
        \centering
        \vspace*{-0.12cm}
        \includegraphics[width=\linewidth]{images/qualitatives/laptop_brand/194273.jpg}
        \captionsetup{labelformat=empty}
        \caption{SD-XL}
    \end{subfigure}
    \hfill
    \begin{subfigure}{0.33\linewidth}
        \centering
        \textbf{Laptop brand}\par\medskip
        \vspace*{-0.12cm}
        \includegraphics[width=\linewidth]{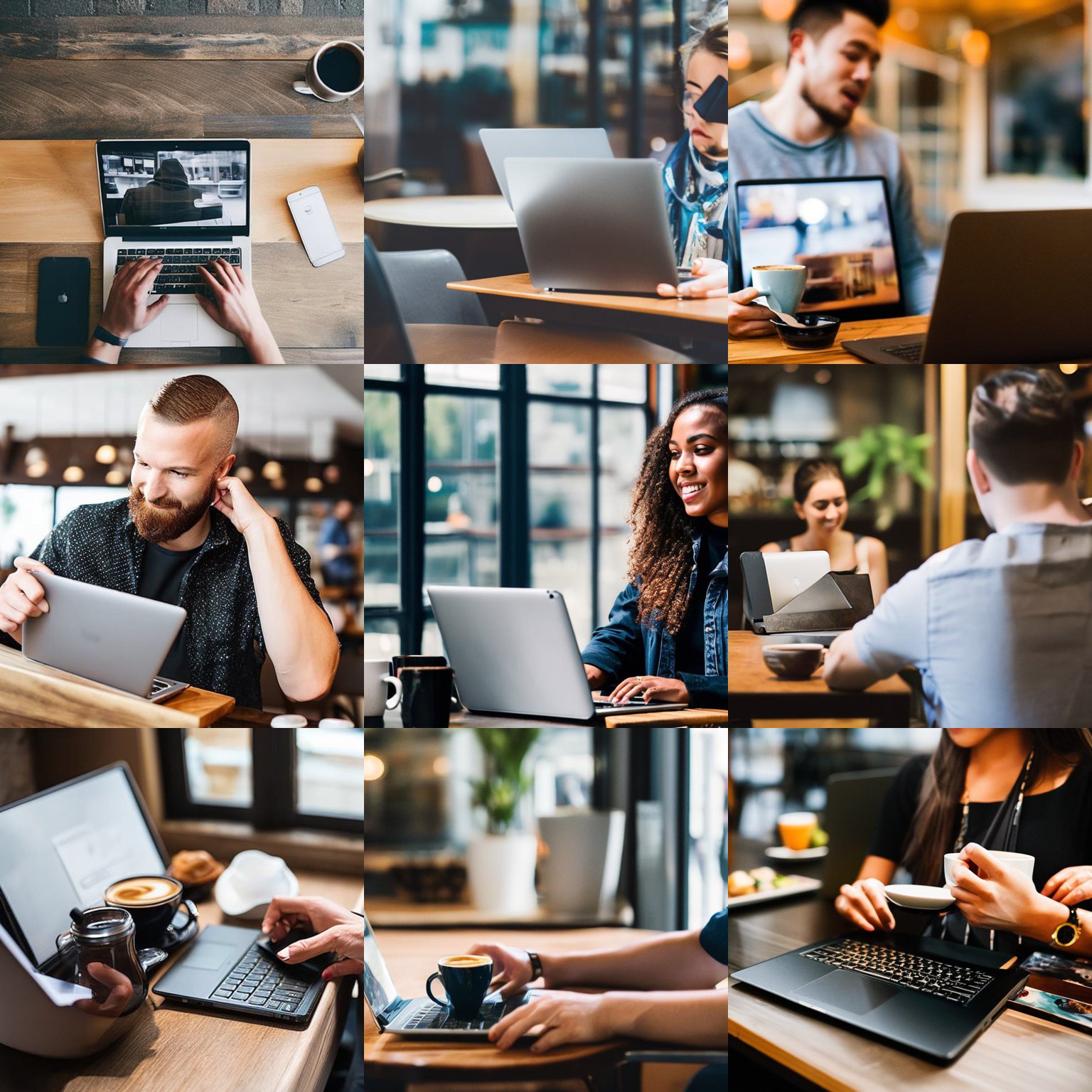}
        \captionsetup{labelformat=empty, justification=centering}
        \caption{SD-2}
    \end{subfigure}
    \hfill
    \begin{subfigure}{0.33\linewidth}
        \centering
        \vspace*{-0.12cm}
        \includegraphics[width=\linewidth]{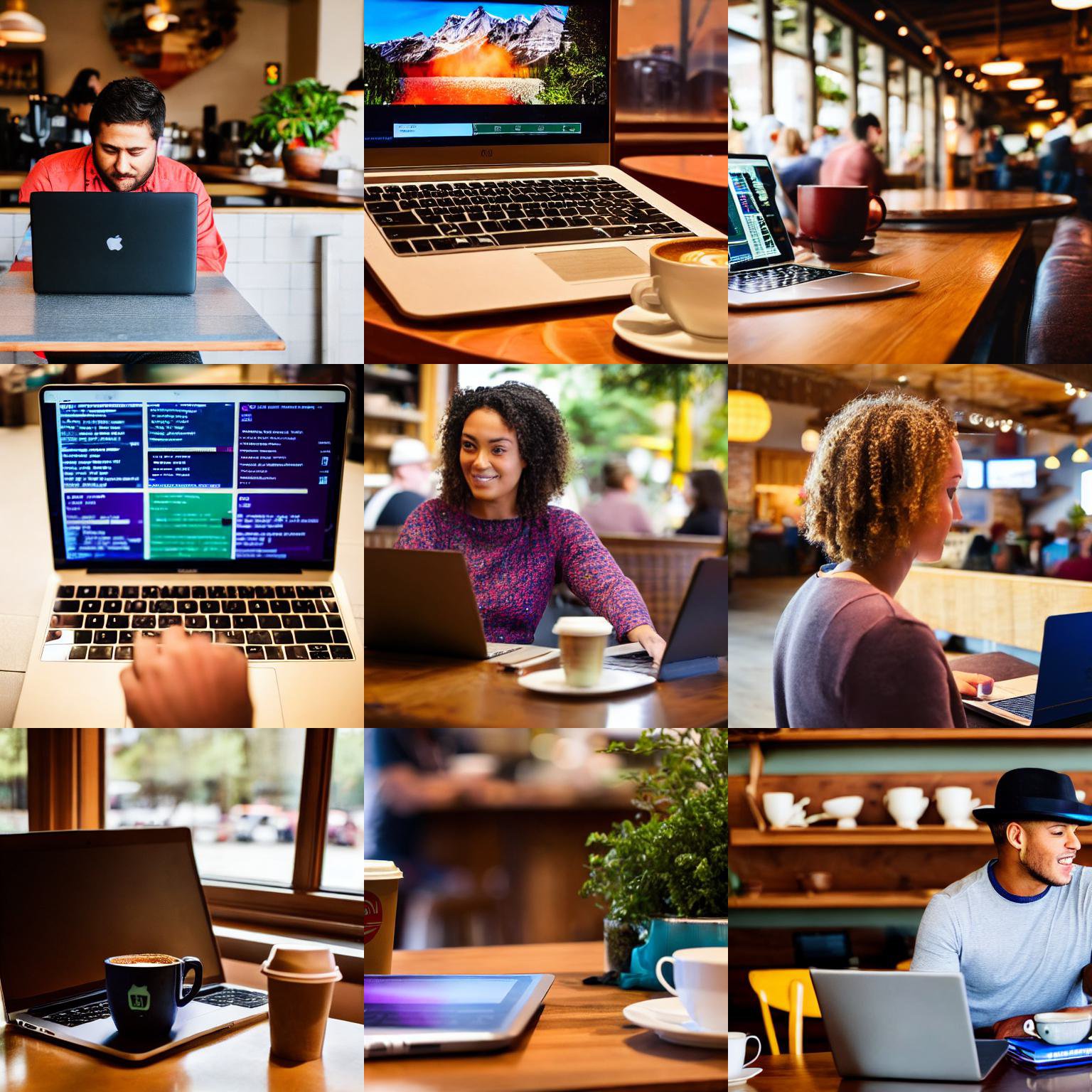}
        \captionsetup{labelformat=empty, justification=centering}
        \caption{SD-1.5}
    \end{subfigure}
    \caption{``A photo of a person on a laptop in a coffee shop".}
    \label{fig:qualitative_comp_laptop}
\end{figure*}

\begin{figure*}[htbp]
    \centering
    \begin{subfigure}{0.33\linewidth}
        \centering
        \vspace*{-0.12cm}
        \includegraphics[width=\linewidth]{images/qualitatives/horse_breed/619417.jpg}
        \captionsetup{labelformat=empty}
        \caption{SD-XL}
    \end{subfigure}
    \hfill
    \begin{subfigure}{0.33\linewidth}
        \centering
        \textbf{Horse breed}\par\medskip
        \vspace*{-0.12cm}
        \includegraphics[width=\linewidth]{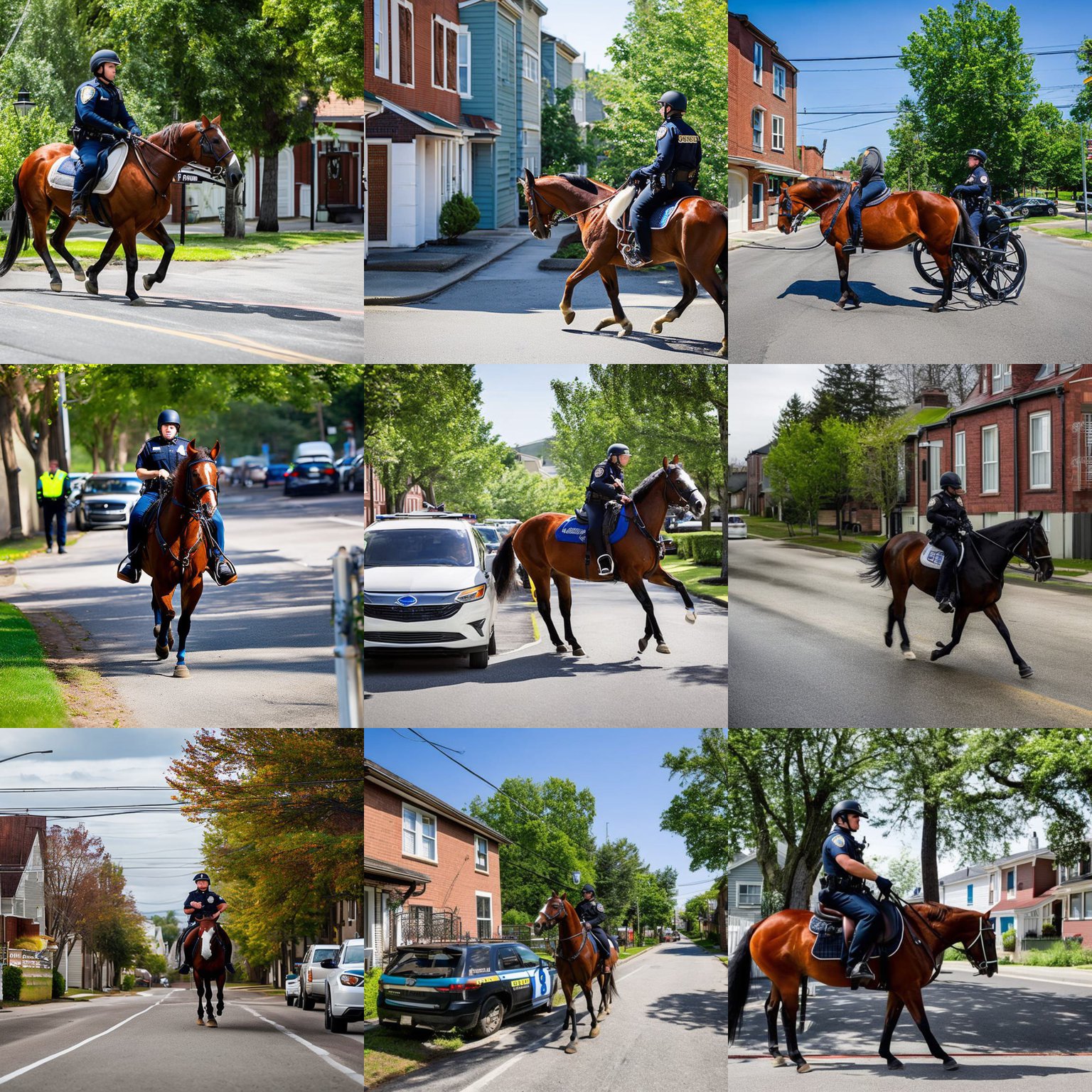}
        \captionsetup{labelformat=empty, justification=centering}
        \caption{SD-2}
    \end{subfigure}
    \hfill
    \begin{subfigure}{0.33\linewidth}
        \centering
        \vspace*{-0.12cm}
        \includegraphics[width=\linewidth]{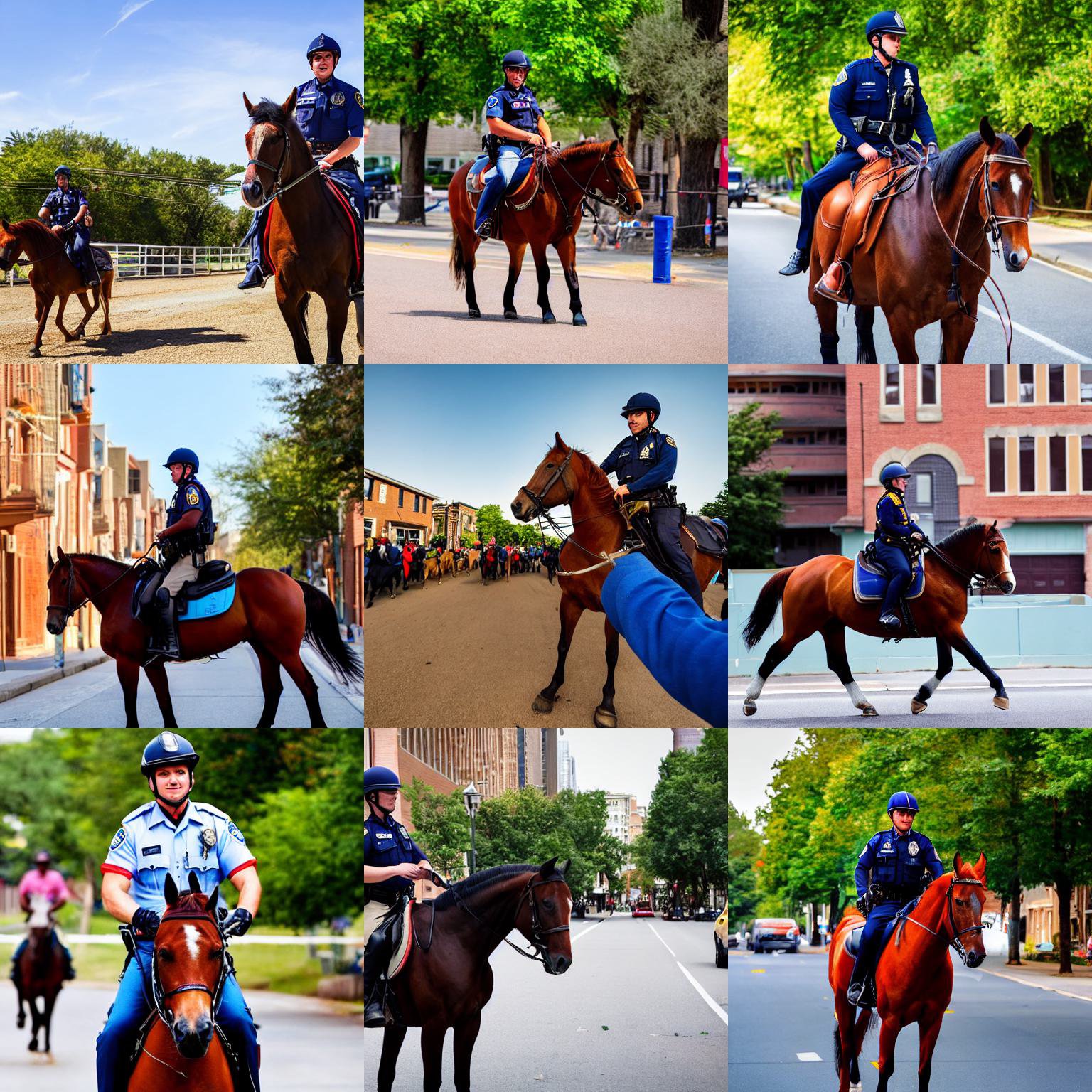}
        \captionsetup{labelformat=empty, justification=centering}
        \caption{SD-1.5}
    \end{subfigure}
    \caption{``A woman riding a horse in front of a car next to a fence".}
    \label{fig:qualitative_comp_horse}
\end{figure*}

\begin{figure*}[htbp]
    \centering
    \begin{subfigure}{0.33\linewidth}
        \centering
        \vspace*{-0.12cm}
        \includegraphics[width=\linewidth]{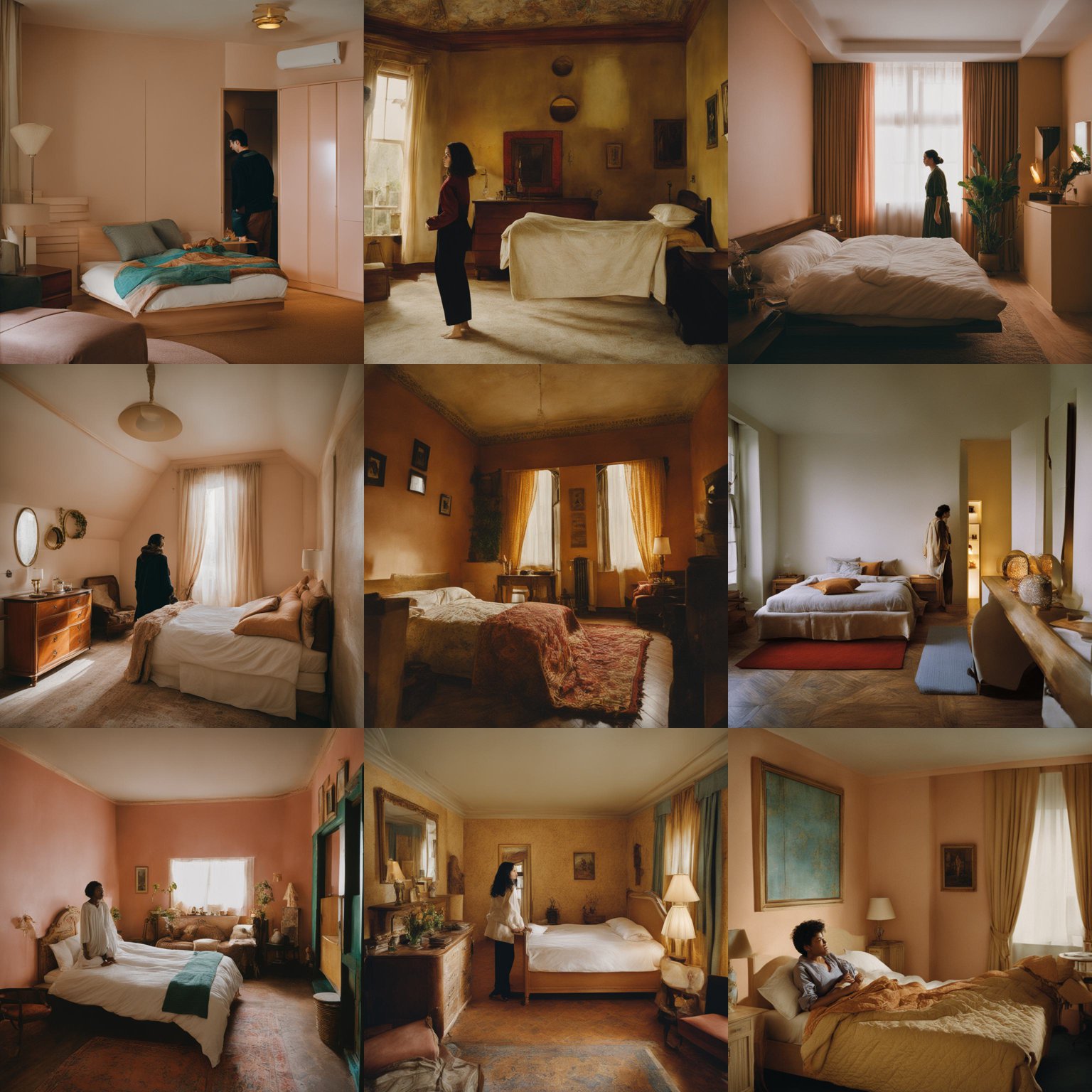}
        \captionsetup{labelformat=empty}
        \caption{SD-XL}
    \end{subfigure}
    \hfill
    \begin{subfigure}{0.33\linewidth}
        \centering
        \textbf{Bed type}\par\medskip
        \vspace*{-0.12cm}
        \includegraphics[width=\linewidth]{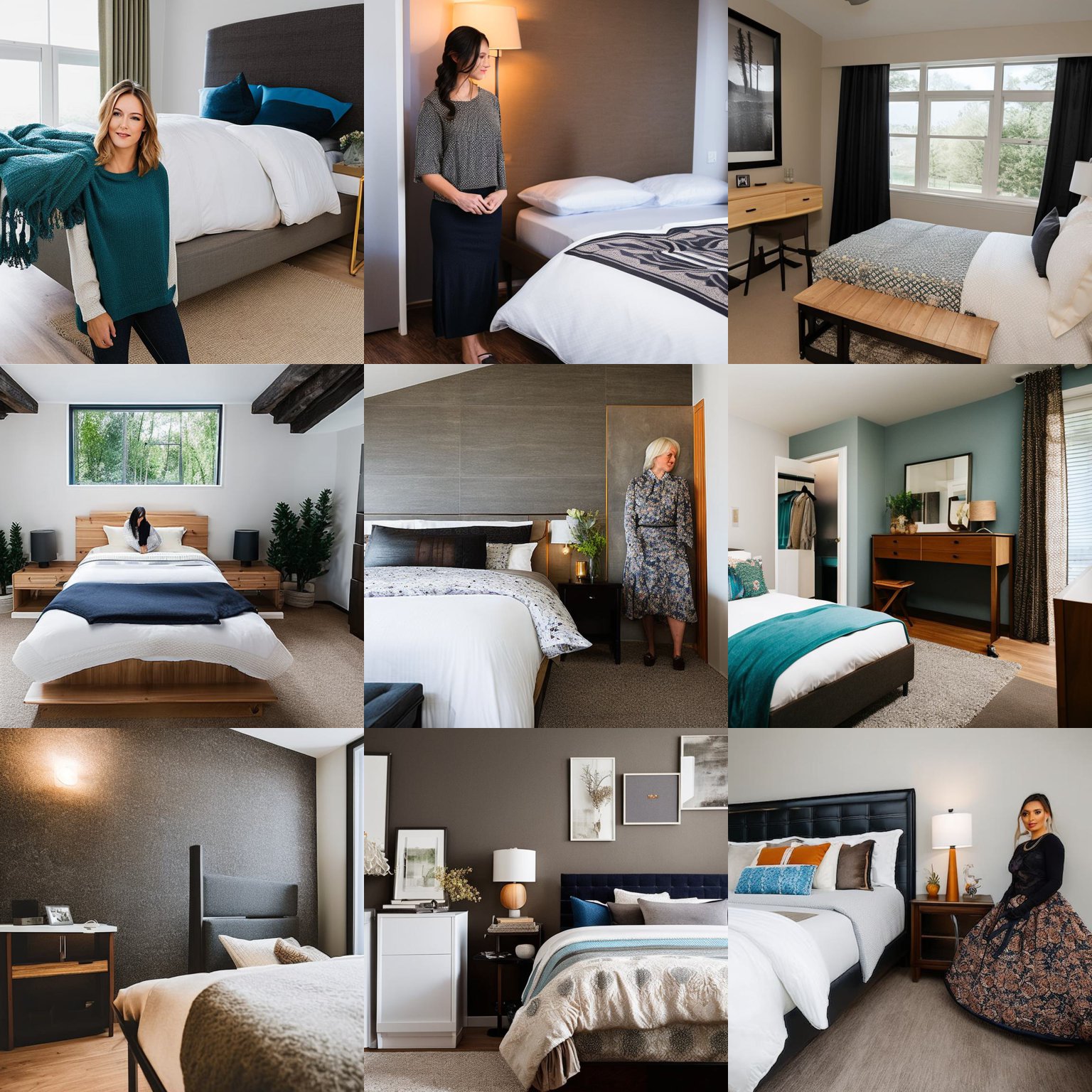}
        \captionsetup{labelformat=empty, justification=centering}
        \caption{SD-2}
    \end{subfigure}
    \hfill
    \begin{subfigure}{0.33\linewidth}
        \centering
        \vspace*{-0.12cm}
        \includegraphics[width=\linewidth]{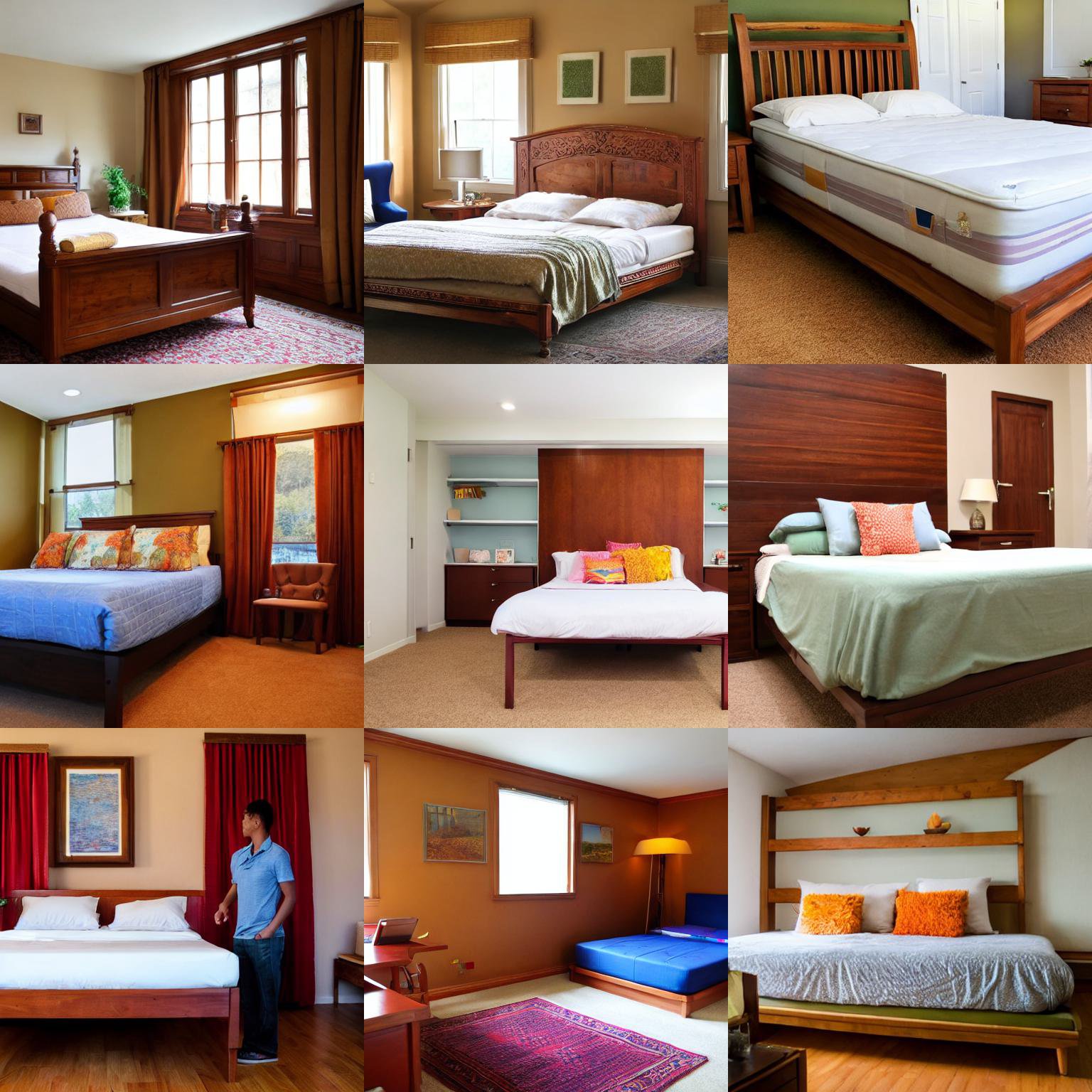}
        \captionsetup{labelformat=empty, justification=centering}
        \caption{SD-1.5}
    \end{subfigure}
    \caption{``A person standing in a bedroom with a bed and a table".}
    \label{fig:qualitative_comp_bed}
\end{figure*}

\begin{figure*}[htbp]
    \centering
    \begin{subfigure}{0.33\linewidth}
        \centering
        \vspace*{-0.12cm}
        \includegraphics[width=\linewidth]{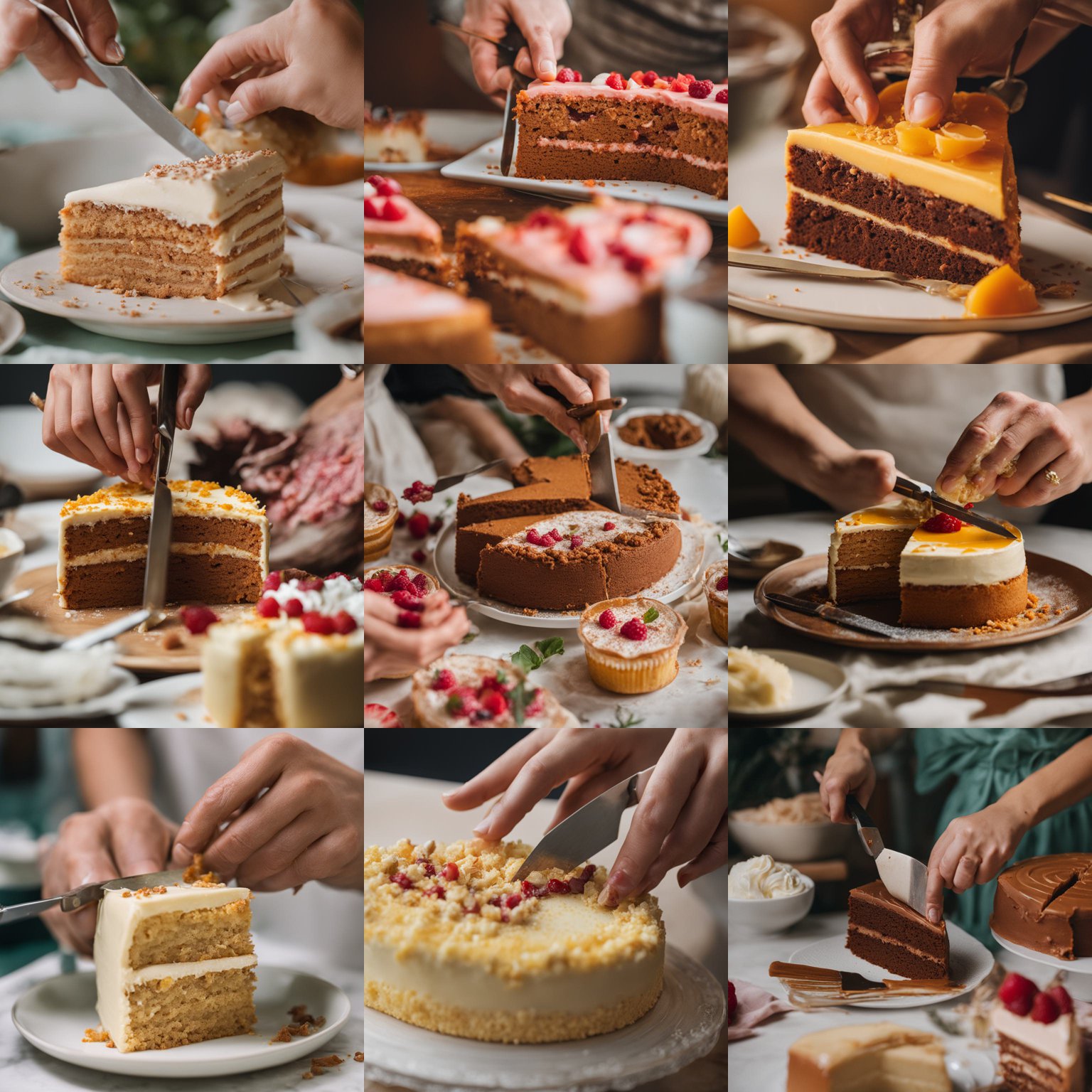}
        \captionsetup{labelformat=empty}
        \caption{SD-XL}
    \end{subfigure}
    \hfill
    \begin{subfigure}{0.33\linewidth}
        \centering
        \textbf{Cake type}\par\medskip
        \vspace*{-0.12cm}
        \includegraphics[width=\linewidth]{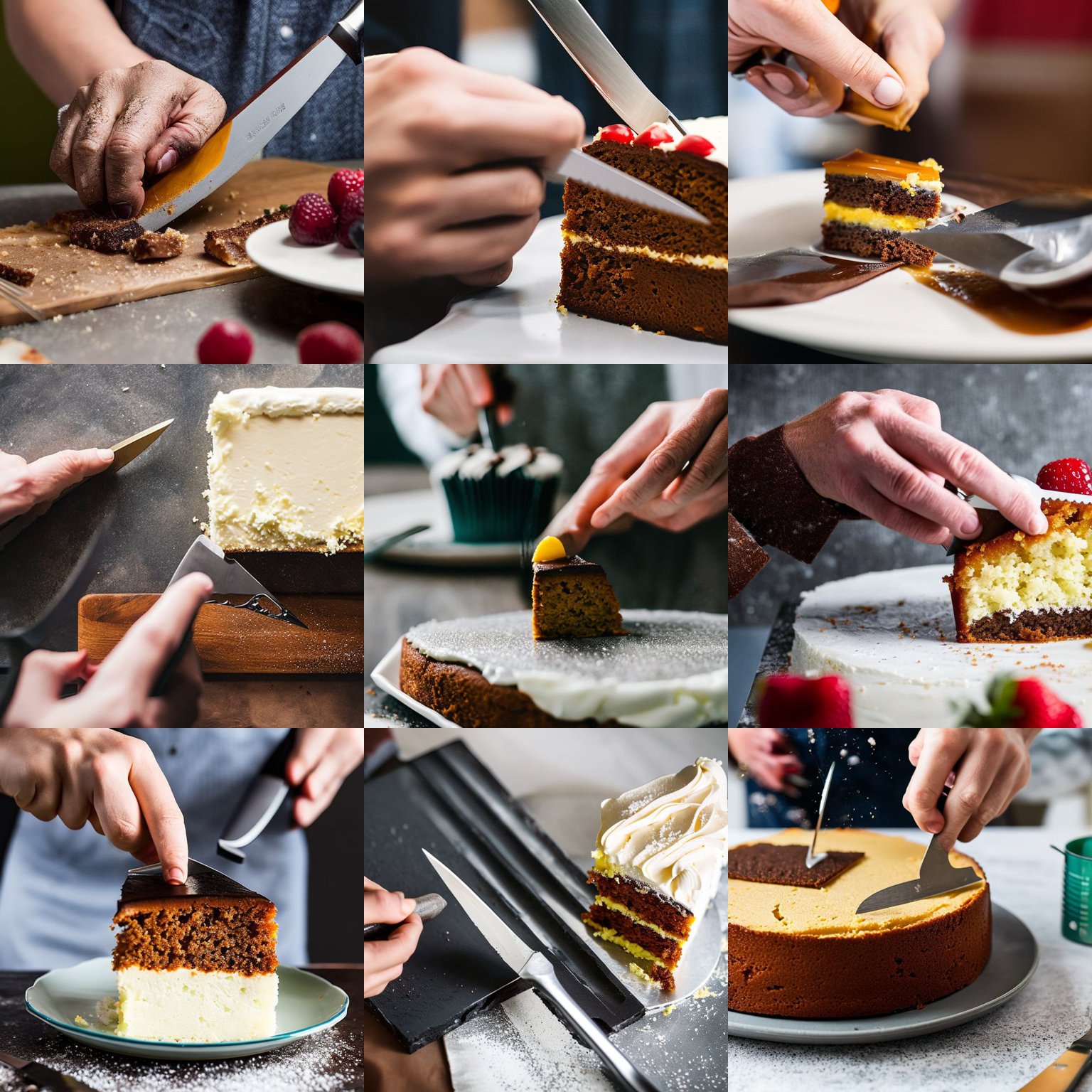}
        \captionsetup{labelformat=empty, justification=centering}
        \caption{SD-2}
    \end{subfigure}
    \hfill
    \begin{subfigure}{0.33\linewidth}
        \centering
        \vspace*{-0.12cm}
        \includegraphics[width=\linewidth]{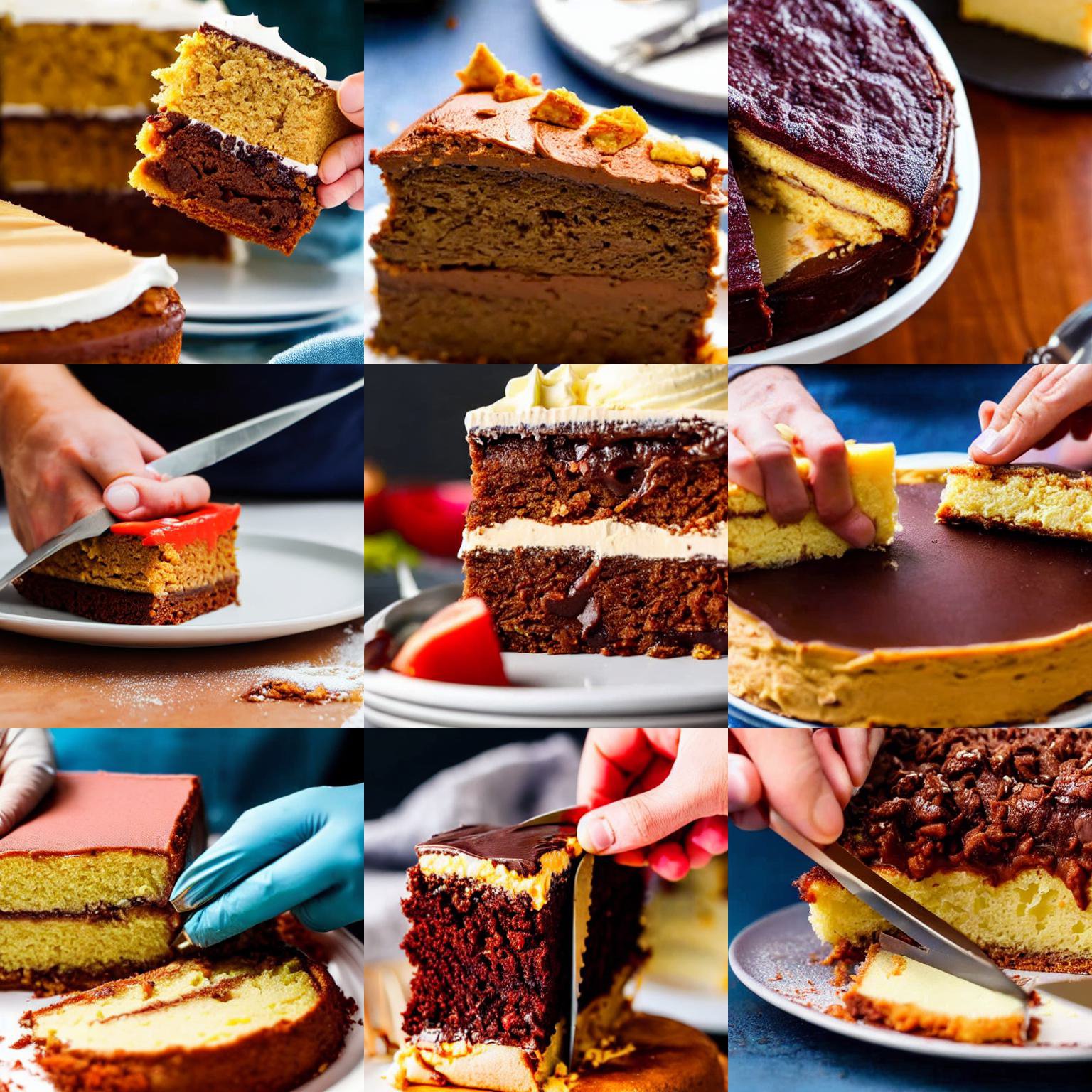}
        \captionsetup{labelformat=empty, justification=centering}
        \caption{SD-1.5}
    \end{subfigure}
    \caption{``A close-up of a person cutting a piece of cake".}
    \label{fig:qualitative_comp_cake}
\end{figure*}

\begin{figure*}[htbp]
    \centering
    \begin{subfigure}{0.33\linewidth}
        \centering
        \vspace*{-0.12cm}
        \includegraphics[width=\linewidth]{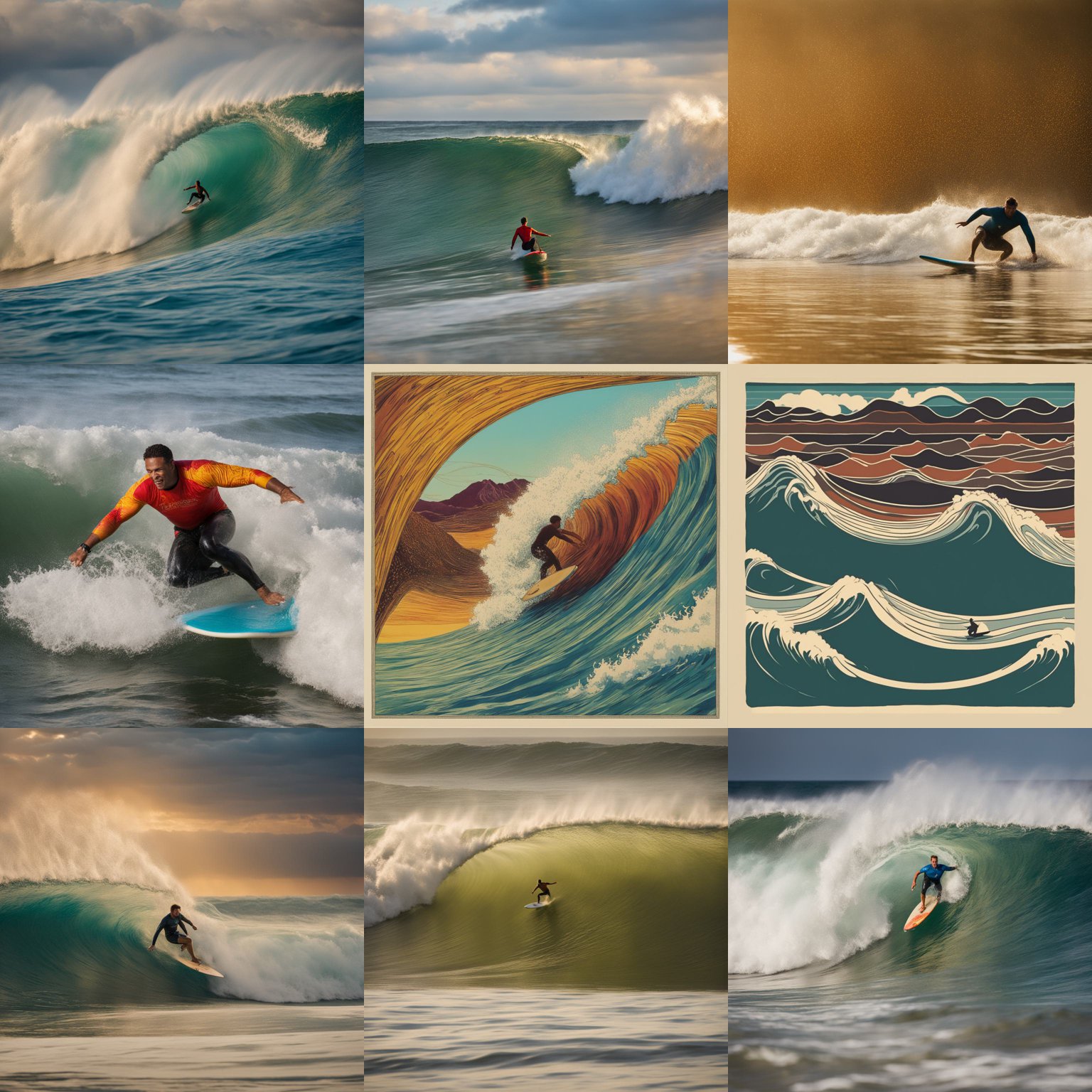}
        \captionsetup{labelformat=empty}
        \caption{SD-XL}
    \end{subfigure}
    \hfill
    \begin{subfigure}{0.33\linewidth}
        \centering
        \textbf{Wave size}\par\medskip
        \vspace*{-0.12cm}
        \includegraphics[width=\linewidth]{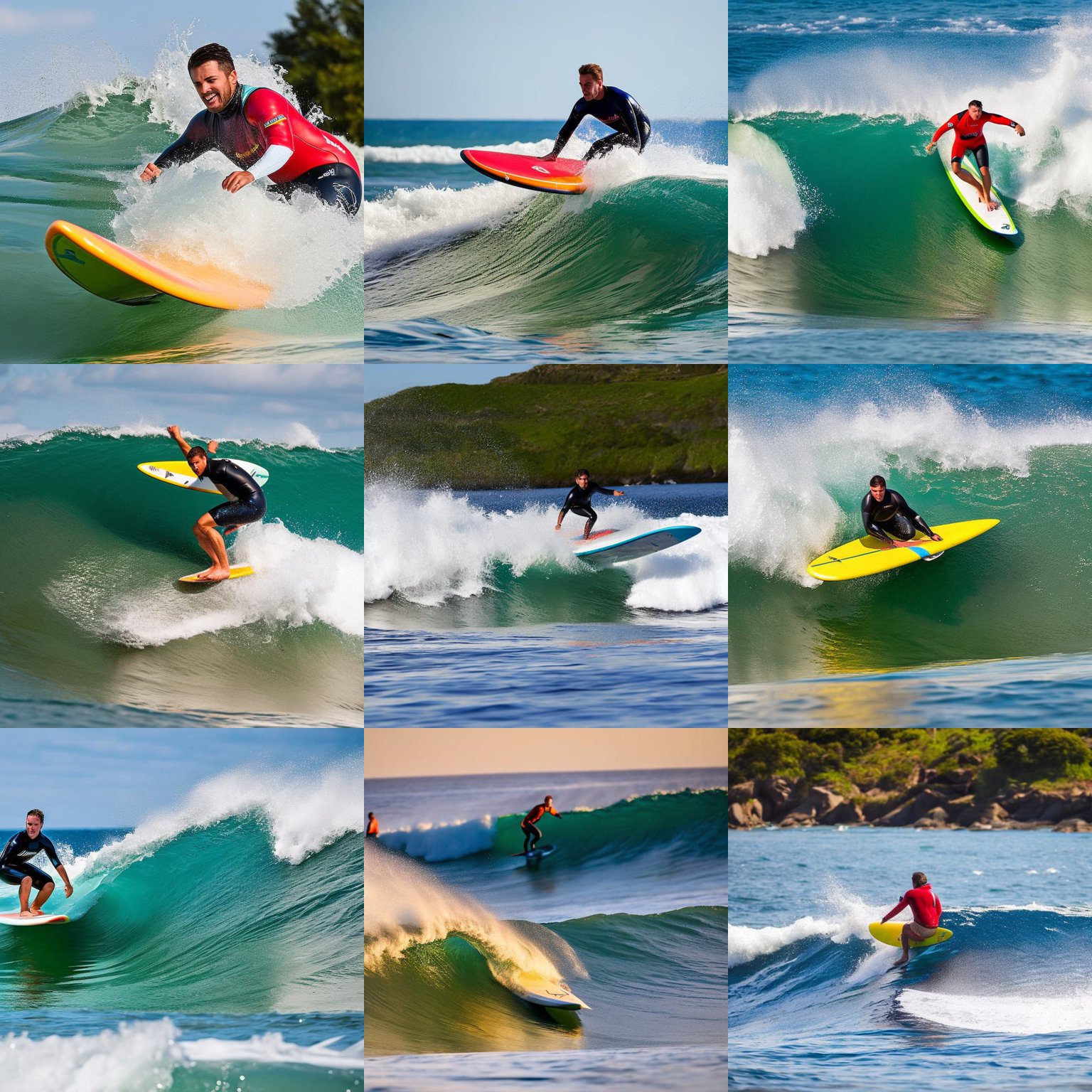}
        \captionsetup{labelformat=empty, justification=centering}
        \caption{SD-2}
    \end{subfigure}
    \hfill
    \begin{subfigure}{0.33\linewidth}
        \centering
        \vspace*{-0.12cm}
        \includegraphics[width=\linewidth]{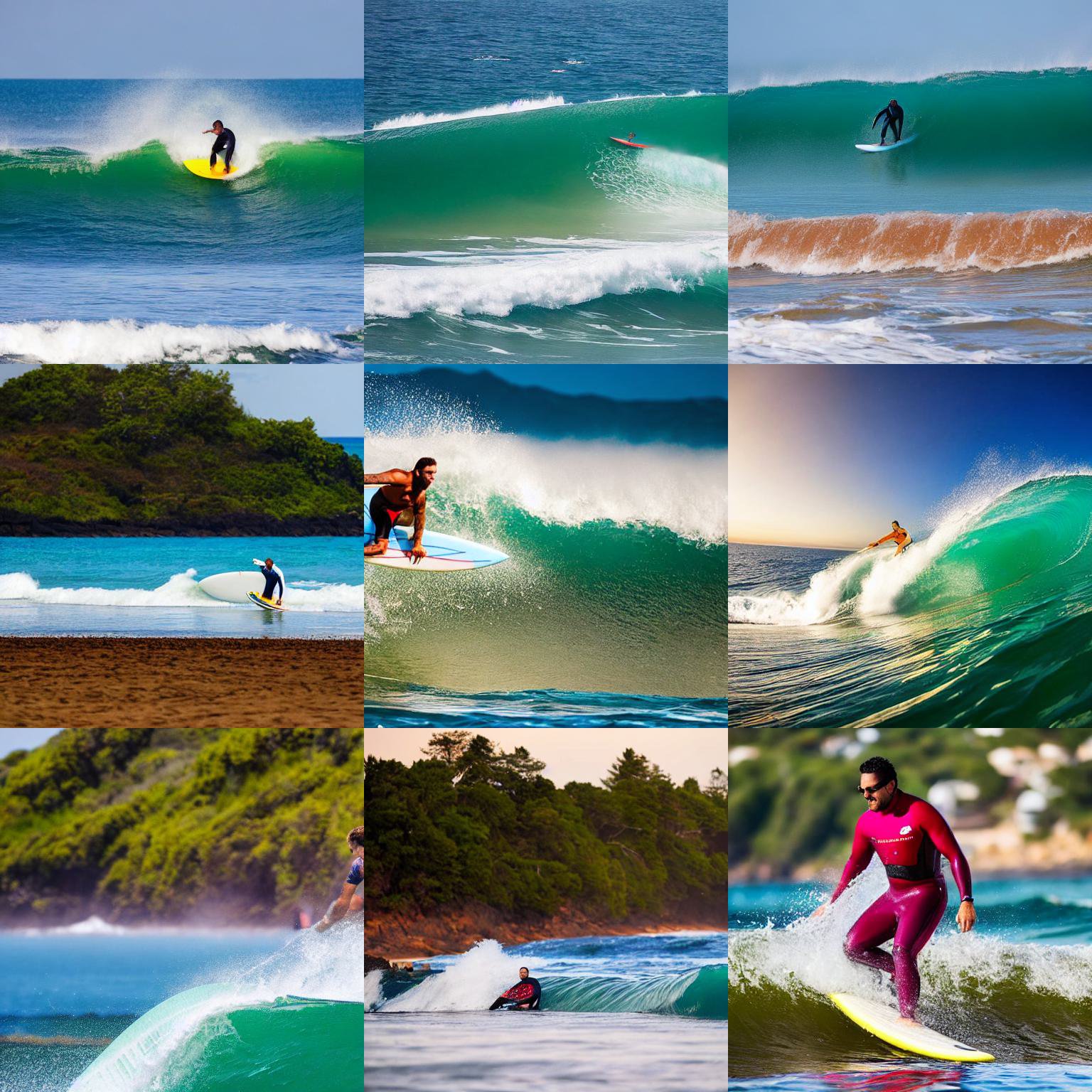}
        \captionsetup{labelformat=empty, justification=centering}
        \caption{SD-1.5}
    \end{subfigure}
    \caption{``A man rides a wave on a surfboard".}
    \label{fig:qualitative_comp_wave}
\end{figure*}

\cref{fig:user_study_screenshot} provides a screenshot of the conducted user study described in Sec. 4.2 of the main paper. The user study provides, for each bias, the generated images at the context level (\ie, generated with the same caption). The user has to choose the majority class and the magnitude of each bias.



\end{document}